\documentclass[preprint,authoryear,10pt,a4paper]{elsarticle}
\usepackage[top=0.75in, bottom=0.75in, left=0.75in, right=0.75in]{geometry}
\usepackage{palatino}
\usepackage{amsmath,amsthm}
\usepackage{amssymb}
\usepackage{bbm}
\usepackage{amsfonts}
\usepackage{graphicx}
\usepackage{subfigure}
\usepackage{grffile}
\graphicspath{{graphics/}}
\usepackage[ruled,vlined,linesnumbered]{algorithm2e}
\usepackage{tabularx,booktabs}
\usepackage{inputenc} 
\usepackage[T1]{fontenc}   
\usepackage{url}            % simple URL 
\usepackage{booktabs}       % 
\usepackage{amsmath,amsthm}
\usepackage{amssymb}
\usepackage{bbm}
\usepackage{amsfonts}       % blackboard math 
\usepackage{nicefrac}       % compact symbols 
\usepackage{microtype}      % microtypography
\usepackage{tikz}
\usepackage{rotfloat}
\usepackage{subfigure}
\usepackage{graphicx}
\usepackage{caption}
\usepackage{appendix}
\usepackage{tablefootnote}
\usepackage{bm}
\usepackage{mathrsfs}
\usepackage{xcolor}
\usepackage{colortbl}

\usepackage{xcolor}
\definecolor{darkblue}{rgb}{0.0,0.5,0.5}
\definecolor{blue}{rgb}{0.0,0.59,0.84}
\definecolor{myblue}{RGB}{0,0,255}

\usepackage[colorlinks]{hyperref}
\hypersetup{colorlinks,breaklinks,linkcolor=darkblue,urlcolor=darkblue,anchorcolor=darkblue,citecolor=darkblue}
\usepackage{booktabs,caption}
\usepackage{multirow,bigdelim}
\usepackage{lscape}
\usepackage{lineno}
\usepackage{microtype}
\newtheorem{lemma}{Lemma}

\newtheorem{remark}{Remark}

\newtheorem{definition}{Definition}

\DeclareMathOperator{\Tr}{Tr}

\usepackage{soul}

\journal{Transportation Research Part C: Emerging Technologies}

\begin{document}

\begin{frontmatter}



% \title{{\fontfamily{lmss}\selectfont Spatiotemporal graph embedded low-rank tensor learning for network-wide traffic speed kriging with incomplete observations}}
% \title{{\fontfamily{lmss}\selectfont  Spatiotemporal traffic speed kriging for enhancing network-wide sensor perception with low coverage: an integrated graph tensor learning method}}

% \title{{\fontfamily{lmss}\selectfont  Correlating sparse sensing for network-wide traffic speed estimation: An integrated graph tensor-based kriging approach}}

\title{{\fontfamily{lmss}\selectfont  Correlating sparse sensing for large-scale traffic speed estimation: A Laplacian-enhanced low-rank tensor kriging approach}}

\author[label]{Tong Nie}
\author[label]{Guoyang Qin}
\author[label1]{Yunpeng Wang}
\author[label]{Jian Sun\corref{cor1}}
\ead{sunjian@tongji.edu.cn}

\address[label]{Department of Traffic Engineering \& Key Laboratory of Road and Traffic Engineering, Ministry of Education, Tongji University, Shanghai 201804, China}
\address[label1]{
Beijing Key Laboratory for Cooperative Vehicle Infrastructure Systems and Safety Control, School of Transportation Science and Engineering, Beihang University, Beijing 100191, China}

\cortext[cor1]{Corresponding author. Address: Cao’an Road 
4800, Shanghai 201804, China}

\begin{abstract}
Traffic speed is central to characterizing the fluidity of the road network. Many transportation applications rely on it, such as real-time navigation, dynamic route planning, and congestion management. Rapid advances in sensing and communication techniques make traffic speed detection easier than ever. However, due to sparse deployment of static sensors or low penetration of mobile sensors, speeds detected are incomplete and far from network-wide use. In addition, sensors are prone to error or missing data due to various kinds of reasons, speeds from these sensors can become highly noisy. These drawbacks call for effective techniques to recover credible estimates from the incomplete data. In this work, we first identify the issue as a spatiotemporal kriging problem and propose a Laplacian enhanced low-rank tensor completion (LETC) framework featuring both low-rankness and multi-dimensional correlations for large-scale traffic speed kriging under limited observations. To be specific, three types of speed correlation including temporal continuity, temporal periodicity, and spatial proximity are carefully chosen and simultaneously modeled by three different forms of graph Laplacian, named temporal graph Fourier transform, generalized temporal consistency regularization, and diffusion graph regularization. We then design an efficient solution algorithm via several effective numeric techniques to scale up the proposed model to network-wide kriging. By performing experiments on two public million-level traffic speed datasets, we finally draw the conclusion and find our proposed LETC achieves the state-of-the-art kriging performance even under low observation rates, while at the same time saving more than half computing time compared with baseline methods. Some insights into spatiotemporal traffic data modeling and kriging at the network level are provided as well.

\end{abstract}


\end{frontmatter}

\section{Introduction}\label{Introduction}
% \textcolor{red}{
% \begin{itemize}
%   \item Traffic speed is the macroscopic quantity used to characterize the fluidity of the road network. Various aspects transportation domain are reliant on accurate traffic speed estimates, such as real-time navigation, dynamic route planning, congestion management, pollution evaluation, and infrastructure optimization.
%   \item Detectors are constantly sensing the speed of vehicles traveling on the road network. Due to prevalent data missing defects from detection failures and absence of detectors, extracting credible and complete traffic speed information from detection for the entire road network becomes a complicated problem.
%   \item Fortunately, traffic speeds are highly interdependent in space and time, posing some degree of redundancy even some values are missing or erroneous. Therefore, detectors can achieve collaborative perception of the traffic speed of the road network if the redundant correlation is effectively utilized (Fig. illustration). 
%   \item Summarize challenges in capturing the spatiotemporal correlation over the road network.
%   \begin{itemize}
%         \item value ambiguity, priors are needed (why low-rank).
%         \item spatial-temporal distance metric as a proxy to speed correlations (why those regularizers and transformations)
%         \item large-scale estimation (why algorithmic approximation)
%     \end{itemize}
%   \item Contribution.
% \end{itemize}
% }

%[Speed matters] 
Traffic speed is the central macroscopic quantity used to characterize the fluidity of the road network. Many applications in the transportation domain rely on the knowledge of traffic speed over the road network to support various dimensions of decision-making, such as real-time navigation, dynamic route planning, congestion management, pollution evaluation, and infrastructure optimization \citep{boriboonsomsin2012eco,liebig2017dynamic,liu2019spatial,han2020congestion}. %To this end, obtaining fine-grained traffic state information in the city wide to enhance the perception is indispensable for constructing more intelligent transportation systems and smarter cities. 
With rapid advances in sensing and communication techniques, collecting traffic speed measurements becomes easier than ever: %have made available an unprecedented amount of traffic measurement data of the transportation systems. %Among various types of traffic data, spatiotemporal traffic data that chronicles the time-varying traffic states over locations is ubiquitous in many downstream traffic applications \citep{zhang2011data}. 
static sensors such as loop detectors are continuously logging vehicular speeds past a cross-section, while mobile sensors such as floating cars are chronicling speeds of their background traffic flow. In fluid mechanics' terminology, static sensors generate Eulerian measurements at fixed locations, while mobile sensors produce Lagrangian measurements following the flow as they move. To obtain a full picture of the network-wide traffic speed, either Eulerian measurements with dense sensor deployments or Lagrangian measurements with high vehicular penetration rates is desired, whereas in fact the reality is far from desirable.%Despite huge potential value of large-scale traffic sensor data, there exists some obstacles weaken its effect in real-world applications. 

\begin{figure}[!ht]
  \centering
  \includegraphics[scale=0.318]{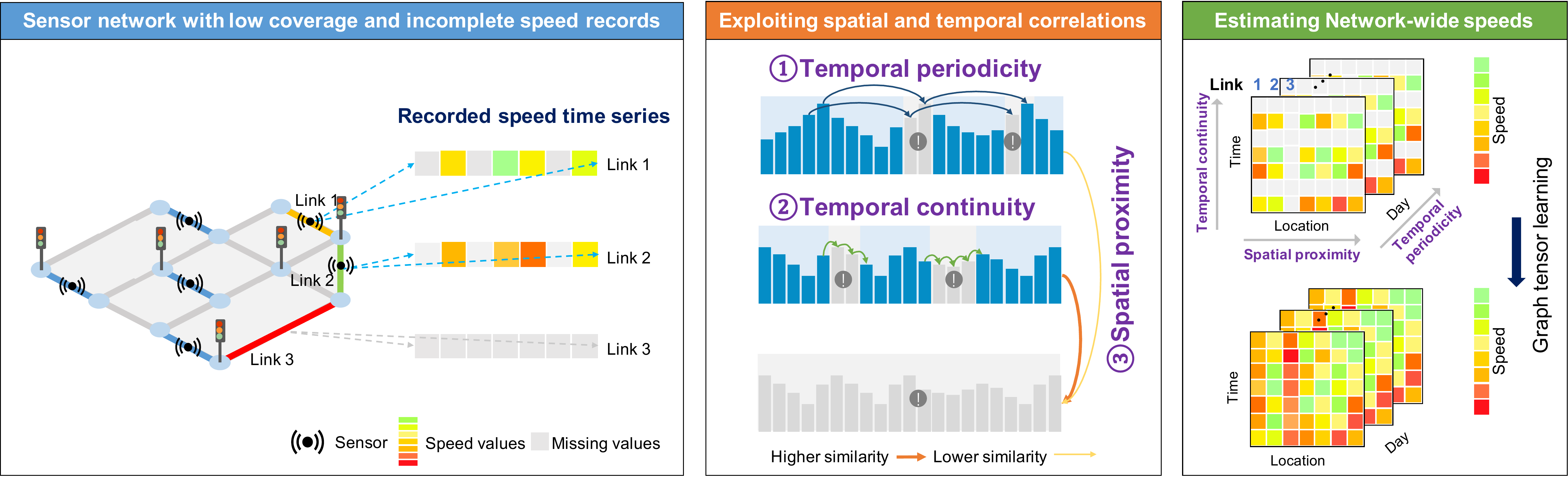}
  \caption{Illustration of spatiotemporal kriging problem. Take link 1, 2 and 3 as an example: link 1 and 2 record incomplete traffic speeds with missing time intervals, and link 3 are not equipped with a sensor so there is no historical record at all. The aim of spatiotemporal kriging is using the multi-dimensional spatial and temporal correlations (temporal continuity, temporal periodicity and spatial proximity) to infer the unmeasured locations (link 3) from the measured locations (link 1 and 2). In this figure, link 2 is physically closer to link 3 than link 1, so it is rational to assume link 2 and link 3 have stronger similarity.}
  \label{intro_fig}
\end{figure}

%[Sparse sensing issues] 
One may see in reality that the static sensors often have a sparse coverage subject to high installation budget and maintenance costs and leave many road links undetected, thereby rendering low spatial resolution data that is hard to extend to the entire network. %For these locations without sensors, there is no historical data to exploit at all.
In the meanwhile, mobile sensors, which are mostly installed on commercial vehicle fleets like taxis, often result in a low penetration rate, which may render dubious speed estimation due to its biased representation of the population. In addition, sensors are prone to error or missing data due to various kinds of reasons such as equipment failure, communication outage, and out-of-service, speeds detected from these sensors can become highly noisy \citep{asif2016matrix}. These drawbacks combined imply that the sensing data is by no means ideal, and call for effective techniques to recover credible estimates from the incomplete data. % Another prevailing issue is the missing data problem of the locations equipped with sensing equipment \citep{asif2016matrix} and the existence of missing data directly affect the usability of traffic measurements to varying degrees. Missing data in traffic system may arise from temporary equipment failure or communication outage, which leads to a complete randomly missing pattern \citep{asif2016matrix}. A more general case is that in non-operational time or energy non-supply time all equipped sensors stop to work, where no information is available at this interval. Therefore, due to prevalent data missing defects from detection failures and absence of detectors, extracting credible and complete traffic speed information from detection for the entire road network becomes a complicated problem.
Fortunately, traffic speeds bounded in the road network are highly correlated in both space and time, and therefore pose a rich degree of redundancy (or low-rankness in a matrix or tensor theory's term \citep{tan2013tensor}) that undetected or erroneous speeds of some road links are recoverable by virtue of the correlation. In other words, sparse sensors are able to achieve a "collaborative perception" of traffic speeds of the undetected road links, when their intrinsic multi-dimensional correlations are effectively exploited (see Fig. \ref{intro_fig}). 

%[Identified as a kriging problem for correlation modeling] 
Recent years have witnessed a vast array of studies investigating the problem of estimating traffic state on undetected road links \citep{meng2017city,zhang2020network,wu2021inductive,lei2022bayesian}. This problem is often cast into spatiotemporal traffic kriging \citep{wu2021inductive} in the literature. Unlike the traffic forecasting problem that aims at predicting future state of a road link given its historical state, spatiotemporal traffic kriging attempts to estimate the traffic state of an undetected road link without any historical state of that link; it instead needs to utilize measurements of other detected road links to complete the "spatial prediction." %unmeasured locations given the measurements from sampled locations during the same period. 
Kriging is therefore regarded as a special missing data imputation problem. \textit{How to effectively utilize the correlation between undetected road links and other sparsely detected roads} is the core research question of this paper. Upon close inspection, we identify three-fold specific challenges: 

% \begin{itemize}
% \item How to characterize correlation: Low-rankness is a widely applied priors used to model the correlation in data imputation problems. But it is become ill-posed when all data of a road link is missing.
% \item Spatial-temporal correlation: why hard? (speed manifests microscopic traffic that is time-varying, dynamic, periodic, and spatially interdependent, spatial correlation is bounded by a graph instead of on a 2-D plane, spatial correlation should consider the graph features as well as traffic flow loaded on that graph)
% \item Computational efficiency: a larger scale network offers more data to leverage correlation for added accuracy, however, it comes at the cost of more time consumption if the computation is costly. Specific theoretical difficulties...
% \end{itemize}

% Contribution (corres. to challenges): 
% \begin{itemize}
% \item How do we adapt low-rankness?
% \item We consider temporal continuity/periodicity, and spatial proximity: graph-based.
% \item Algorithm to speed up (scalable) and keep accuracy.
% \end{itemize}
% --------

First, correlation between speeds of the detected road links and speeds of the undetected ones can be arbitrary unless reasonable prior assumptions are made. % to value ambiguity, and remarkable changes of traffic state over space and time.
Low-rankness is a widely-adopted general prior in literature to reflect linear correlation of traffic data over space and time \citep{tan2013tensor,asif2016matrix,chen2021scalable}. However, low-rankness is not directly applicable to our kriging problem in which data are fully missing in some locations, as in this case, we can fill different sets of values to the missing data without changing the rank and the missing values therefore remain undetermined under the low-rankness assumption. %as low-rank assumption is not sensitive to the permutation of data index (without positional encoding) and apt to capture global similarity of temporal dimension, when all data of a road link or time point is missing, the imputed results could be erroneous. In this work, we aim at tackling an intricate scenario: kriging with incomplete observations, even with whole missing time intervals or days, as shown in Fig. \ref{intro_fig}. 
Conceptually, low-rankness is not sufficient when imputing a space-time matrix with entire row or column missing. Additional regularization terms should be carefully considered to complement the low-rankness assumption  \citep{bahadori2014fast,yokota2018missing,yamamoto2022fast}. 

Second, in addition to general correlation assumptions, exploiting the semantic spatiotemporal correlation of speeds is an necessary add-on. Since speed manifests microscopic traffic that is time-varying, dynamic, periodic, and spatially interdependent, in the meanwhile, spatial correlation is bounded on a graph instead of on a 2D plane, spatial correlation should consider the graph features (e.g., network topology) as well as traffic flow loaded on that graph (e.g., moving directions of traffic flow) \citep{li2017diffusion}, exploiting spatio-temporal correlations is by no means trivial. %[Correlation modeling is not easy] Unlike RGB images or videos, speed manifests microscopic traffic that is time-varying, periodic, and spatially interdependent. Temporal dimension shows multi-resolution traits and usually can not be captured by a single-scale model. Spatial correlation is bounded by a graph instead of on a 2D plane and graph-distance based metric can be used as a proxy to represent correlation of the traffic flow loaded on that sensor graph \citep{li2017diffusion}. 
Moreover, as a graph signal, traffic data features inherent time series characteristics such as continuity, periodicity, and stationary, providing another prior for ones to achieve time-varying recovery.
Tailoring an effective kriging model to capture the multi-modal correlations of spatiotemporal traffic data is still an open question yet to be resolved.

Third, scalability of the kriging approach onto the network level is desired as it allows exploiting correlation of a wider scope of speeds over space with added accuracy. Moreover, %as matrix and tensor completion based models are transductive, 
when new measurements are collected, a kriging model with fast parametric update is preferable for an online estimation.%, as the data storage and transmission is expensive of the whole city network.
To this end, computational efficiency is the key.
%On the scale of city network, more measurements are available and can be leveraged to mine more comprehensive correlations as the sensor graph becomes larger. And the spatial correlations among different links get more distinct both globally and locally. To fully utilize these spatial correlations, which is at the core of spatiotemporal kriging, we are supposed to integrate the traffic measurements at the whole network. While, handling network-wide of data is nontrivial. 
However, classical Gaussian process (GP) based kriging doesn't show scalability due to its computational complexity, while existing scalable tensor based imputation methods \citep{lu2019low,chen2021scalable} are mainly designed for general data imputation problem, and are not applicable to the kriging task. Although Bayesian Gaussian kriging methods %are incorporated into the matrix factorization framework to 
achieve flexible hyper-parameter tuning for spatiotemporal kriging \citep{lei2022bayesian,chen2022bayesian}, its high consumption of time and memory in parametric learning of the Gaussian kernel inhibits itself from being scaled to the network level. %Spatiotemporal kriging at network level requires models not only can capture diverse spatial correlations but also save time and space complexity. 
Custom scalable solution algorithms need to be designed to maximize the kriging performance.

%While prior works initially introduce the usage of matrix/tensor based methods to spatiotemporal traffic kriging problem, there is still a large gap towards the large-scale applicability of such approaches. To bridge the gap, 
To tackle the challenges and realize a network-wide kriging of missing speed data for the undetected road links, we propose 
a Laplacian enhanced tensor completion (LETC) based kriging model to exploit multi-modal spatiotemporal correlations of traffic speeds, where three kinds of carefully chosen speed correlation -- \textit{temporal continuity}, \textit{temporal periodicity} and \textit{spatial proximity} -- are simultaneously captured by assigning three forms of graph Laplacian. By integrating graph Laplacian into tensor completion model either explicitly or implicitly, LETC can leverage both global low-rankness property and local dependency informed by physical constraints at the same time.
 %a scalable and unified framework, to lightweight the real-world application of tensor based method on . 
The framework contributes to the current literature in four aspects as follows: 

\begin{enumerate}
%(How do we adapt low-rankness?) 
\item {We customize a new temporal graph Fourier transform (TGFT) based on temporal periodic graph Laplacian to encode the temporal periodicity of speeds. The TGFT complements the low-rankness assumption and enables parallelization of the tensor completion by a series of daily matrix in graph spectral domain.}
% (We consider temporal continuity/periodicity, and spatial proximity: graph-based.)
% \item  We embed spatial proximity and temporal continuity of speeds in space-time domain as an add-on to low-rankness and integrate them in a holistic graph tensor learning framework through a matrix-tensor conversion. Both global and local correlations can therefore be modeled simultaneously. 
% (Algorithm to speed up (scalable) and keep accuracy.)
\item {We propose a novel diffusion graph regularization (DGR) to exploit spatial dependency on directed graphs. By modeling as random walkers on graphs, DGR considers the directed nature of traffic flows and closely connects with the graph diffusion process. Moreover, DGR features explicit physical explanations and encourages a low-pass propagation which is beneficial for reconstructing spatiotemporal traffic data.
% significantly improves the kriging accuracy by avoiding errors arise from the symmetrization of graphs. 
}
\item {We develop a general formulation of temporal consistency regularization (GTCR) to impose smoothness constraints on time slots within a kernel size. GTCR considers the causality of time series in a model-free manner and can directly convert into some previously developed temporal regularization terms.} 
\item {To scale up the kriging on large-scale network, we speed up our model by randomized tensor singular value decomposition and conjugate gradient method. Experiments on million-level traffic datasets not only reveal the superiority of LETC on both accuracy and computational efficiency, but also provide insights into handling large-scale kriging problem.}
% \item  \textcolor{myblue}{To scale up the kriging on large-scale network, we propose a randomized singular value decomposition-based tensor singular value thresholding algorithm, and adopt conjugate gradient method to update matrix variables in an efficient manner. Both of the numeric techniques significantly reduce the computational burden while maintaining high accuracy.}
\end{enumerate}
% \begin{itemize}
% \item We model the spatiotemporal traffic speed kriging with incomplete observations problem as a graph embedded tensor learning problem and a new temporal graph Fourier transform (TGFT) is developed to encode the 'day' mode correlations. Induced by TGFT, the transformed tensor nuclear (t-TNN) norm minimization can be solved efficiently by a series of matrix subproblems in graph spectral domain.
% \item To perform kriging at network wide, we develop a randomized singular value decomposition based tensor singular value thresholding algorithm to solve the t-TNN minimization, and adopt conjugate gradient to update matrix variables in an efficient manner. Both of the numeric techniques significantly reduce the computational burden while maintain high accuracy.
% \item Comprehensive experiments are conducted on two large-scale public traffic speed datasets and results show that the proposed LETC is computationally efficient and can achieve high accuracy. This paper also reveals that the kriging precision, time cost, and hyper-parameter sensitivity all benefit from large data scale.
% \end{itemize} 

The rest of this article is organized as follows. Section \ref{Literature review} reviews related works about spatiotemporal kriging. Section \ref{Notations and Problem Definitions} introduces basic concepts about tensor operations in this work and defines the tensor-based kriging problem. In Section \ref{methodology}, we propose the graph tensor-based kriging model and a scalable solution algorithm. In Section \ref{experiments}, we perform numerical experiments of the proposed approach on two public data sets, which is followed by model comparison, ablation studies, and sensitivity analysis. Section \ref{conclusions} concludes this work and provides future directions.

\section{Related work}\label{Literature review}
Kriging is typically a regression problem for spatial modeling and interpolation of random fields based on covariance functions, which is a ubiquitous topic in geostatistics and usually associated with GP regression. 
Recently, kriging tasks have aroused great interests from different disciplines in machine learning community and many of these works can be viewed as special variants of kriging. In this section, we discuss kriging problems and summarize related solutions from a general perspective: spatial prediction with no historic data.

\subsection{Graph regularized low-rank matrix and tensor completion based methods}
In recommendation system community, graph regularized matrix completion/factorization models are commonly used for collaborative filtering tasks \citep{kalofolias2014matrix,rao2015collaborative,strahl2020scalable}. The main purpose is to predict the unknown node values using side information about edges, which resembles the motivation of kriging. \cite{kalofolias2014matrix} first proposed a matrix completion model with graph Laplacian regularization. \cite{rao2015collaborative} solved this problem in a scalable manner called graph regularized alternating least squares method, where conjugate gradient method was applied to update parameters.
In the study of transportation, graph regularization is introduced either in space-time domain \citep{bahadori2014fast,wang2018traffic,zhang2020network,yang2021real}, or graph spectral domain \citep{deng2021graph} to model spatial dependency. Among these works, \cite{bahadori2014fast} proposed a tensor learning model for spatiotemporal co-kriging. The optimization problem was solved by greedy algorithm with orthogonal projections. Most of these works are developed for other data source and can not be directly transferred to our problem, as the unique characteristics of traffic speed should be taken into consideration.

\subsection{Bayesian/probabilistic matrix factorization based method}
Since kriging is a basic problem in GP modeling, some works combine GP priors with matrix factorization in a whole probabilistic framework to inference unknown data \citep{luttinen2009variational,zhou2012kernelized,yang2018fast,strahl2020scalable}. When graph Laplacian is specified as the kernel (covariance) function, this kind of model is equivalent to the graph regularized matrix factorization model. Most recently, \cite{lei2022bayesian} proposed a Bayesian kernelized matrix factorization model for spatiotemporal traffic data imputation and kriging. Using a fully Bayesian treatment, this method can learn the hyper-parameters of graph kernels through resampling. One downside can not be overlooked is that the hyper-parameter sampling procedure is quite time-consuming. Therefore, there seems to be a trade-off between model scalability and flexibility.

\subsection{Graph neural network based method}
Graph convolution neural networks (GCN) has shown great potential for inductive modeling and achieved promising performances on kriging tasks \citep{appleby2020kriging,wu2021inductive,wu2021spatial,liang2022spatial}. \cite{appleby2020kriging} first developed a kriging convolutional networks using K-nearest neighbor to reconstruct all the node values on a defined graph. \cite{wu2021inductive} proposed an inductive GCN to model the spatial dependency and a random mask sample generation method was used for model training. However, both of these methods consider the sample over a specific observation time as individual or additional features, thus ignoring the temporal consistency.  \cite{liang2022spatial} further improved the kriging performance of GCN by integrating an external attention  mechanism and a temporal convolution network. Despite high accuracy has been achieved, training these elaborate GCN models is computationally very expensive and may require GPU hardware acceleration, which is prohibitive for large-scale problems.

\subsection{Multi-way delay embedding (Hankelization) based methods}
In the field of computer vision, kriging can be regarded as recovery of missing pixels along entire rows or columns of RGB images, i.e., a special image inpainting task \citep{yokota2018missing}. Hankel matrix that bases on shift-invariant features of image is supposed to be a powerful tool for this issue. \cite{yokota2018missing} proposed a Tucker tensor decomposition in multi-way delay-embedded (MDT) space method (aka., Hankelization) and extend the Hankel matrix to tensor. \cite{yamamoto2022fast} further improved the scalability of Hankel tensor completion by using circulant MDT approximation. Although these pioneering works do not aim to solve spatiotemporal traffic kriging, the MDT based method has been applied to various transportation problems, such as traffic speed estimation from trajectories \citep{wang2021low}, traffic data anomaly detection using RPCA \citep{wang2021hankel}, and short-term traffic forecasting \citep{shi2020block}. In the context of spatiotemporal traffic data, there exists shift-invariant patterns in the time dimension, while this phenomenon is not distinctive in the space dimension. Without external spatial information, the MDT may produce erroneous kriging results.

\section{Preliminaries and problem definitions}
\label{Notations and Problem Definitions}

In this section, we first introduce some basic notation and preliminaries about tensor algebra. After that, we describe the spatiotemporal traffic speed kriging problem in the perspective of graph tensor learning framework. 

\subsection{Notation and basic concepts}
Throughout this paper, we use the same notations as in \citep{kolda2009tensor}. Specifically, we adopt boldface capital letters to denote matrices, e.g., $\mathbf{M}\in\mathbb{R}^{M\times N}$, boldface lowercase letters, e.g., $\mathbf{a}\in\mathbb{R}^{M}$ to denote vectors, and scalars are denoted by lowercase letters, e.g., $a$. For a matrix $\mathbf{M}$, the $i$-th row and $j$-th column are abbreviated as $\mathbf{M}^{i}$, $\mathbf{M}_{j}$. A third-order tensor is denoted by calligraphic letter, e.g., $\mathcal{X}\in\mathbb{R}^{n_1\times n_2\times n_3}$, whose $\left(i,j,k\right)$-th entry is $x_{ijk}$, and we use the MATLAB notation $\mathcal{X}(i,:,:)$, $\mathcal{X}(:,i,:)$ and $\mathcal{X}(:,:,i)$ to denote the $i$-th horizontal, lateral and frontal slice, respectively. Specially, the frontal slice $\mathcal{X}(:,:,i)$ is denoted compactly as $\mathbf{X}^{(i)}$. A tubal fiber (mode-3 fiber) $\mathcal{X}(i,j,:)$ is obtained by fixing the first two indices and varying the third index.
The inner product of two matrix is given by $\left\langle\mathbf{A},\mathbf{B}\right\rangle=\Tr(\mathbf{A}^\mathsf{T}\mathbf{B})$, where $\Tr(\cdot)$ signifies the matrix trace.
The inner product of two third-order tensors is defined by $\left\langle\mathcal{A},\mathcal{B}\right\rangle=\sum_{i,j,k}a_{i,j,k}b_{i,j,k}$, and the corresponding tensor Frobenius norm is $\parallel\mathcal{A}\parallel_F=\sqrt{\left\langle\mathcal{A},\mathcal{A}\right\rangle}$.

For matrices, we introduce the reshaping and vectorization operation for efficient matrix-vector product. If $\mathbf{X}\in\mathbb{R}^{M\times N}$, then the vectorization of $\mathbf{X}$ denoted by $\operatorname{vec}(\mathbf{X})$ is an $NM$-by-$1$ vector obtained by stacking the columns of $\mathbf{X}$:
\begin{equation}
\operatorname{vec}(\mathbf{X})=
\begin{bmatrix}
\mathbf{X}(:,1) \\
\vdots \\
\mathbf{X}(:,N)
\end{bmatrix}.
\end{equation}

Let $\mathbf{A}\in\mathbb{R}^{m\times n}$ and $m_1n_1=mn$, then the outcome of reshaping $\mathbf{A}$: 
\begin{equation}
    \mathbf{B}=\operatorname{reshape}(\mathbf{A},m_1,n_1)
\end{equation}
is a $m_1$-by-$n_1$ matrix defined by $\operatorname{vec}(\mathbf{B})=\operatorname{vec}(\mathbf{A})$.

Graph Laplacian serves as a regularization term and has ubiquitous applications in geometric matrix factorization models \citep{kalofolias2014matrix,rao2015collaborative}. Given an undirected and possibly weighted graph $\mathscr{G}=(\mathscr{N},\mathscr{E},\mathbf{W})$ with nodes $\mathscr{N}$ and edges $\mathscr{E}\subseteq \mathscr{N}\times\mathscr{N}$ weighted with a non-negative weight matrix (adjacent matrix) $\mathbf{W}$, the graph Laplacian is calculated from the weight matrix:
\begin{equation}
    \mathbf{Lap}=\mathbf{D}-\mathbf{W}, \mathbf{D}=\operatorname{Diag}(\sum_{i'}w_{ii'}).
\label{Laplacian}
\end{equation}

Invertible linear transform, is proved to be a flexible tool for different real data formatted as tensors \citep{lu2019low}. Given an invertible linear transform, one can establish a transform induced tensor product (t-product) and tensor singular value decomposition (t-SVD) accordingly. Let $\mathbf{L}\in\mathbb{C}^{n_3\times n_3}$ be an arbitrary invertible transform matrix (e.g., discrete Fourier transform (DFT), discrete cosine/sine transform (DCT/DST), unitary transform (UT) and graph Fourier transform (GFT)) that satisfies:
\begin{equation}
    \mathbf{L}\mathbf{L}^\mathsf{H} = \mathbf{L}^\mathsf{H}\mathbf{L} = l\mathbf{I}_{n_3},
\label{transform}
\end{equation}
where $l>0$ is a constant and $\mathsf{H}$ is the conjugate transposition, then given a third-order tensor $\mathcal{X}\in\mathbb{R}^{n_1\times n_2\times n_3}$, the transform matrix is applied on each tube fiber $\mathcal{X}(i,j,:)$ along the \textbf{third-dimension} of $\mathcal{X}$, i.e., 
\begin{equation}
    \Bar{\mathcal{X}}(i,j,:)=\mathbf{L}\mathcal{X}(i,j,:),
\end{equation}
which is equivalent to:
\begin{equation}
    \Bar{\mathcal{X}}=\mathscr{L}(\mathcal{X})=\mathcal{X}\times_3\mathbf{L},
\label{forward}
\end{equation}
where we use $\mathscr{L}(\cdot)$ to denote the linear transform operator, $\Bar{\mathcal{X}}$ is the transformed tensor, and $\times_3$ denotes the mode-3 product \citep{kolda2009tensor}. Similarly, the inverse transform is given by:
\begin{equation}
    \mathcal{X}=\mathscr{L}^{-1}(\Bar{\mathcal{X}})=\Bar{\mathcal{X}}\times_3\mathbf{L}^\mathsf{H}.
\label{inverse}
\end{equation}

To give an intuitive example, when we adopt the DFT, the transformed tensor can be obtained by performing DFT of $\mathcal{X}$ along the third-dimension, i.e., $\Bar{\mathcal{X}}=\operatorname{fft}(\mathcal{X},[~],3)$ by using the MATLAB command $\operatorname{fft}$.

The mechanism of transformed tensor completion is that by imposing transform along certain dimension, the whole tensor completion problem can be divided into solving a series of small-scale matrix completion problems for each frontal slice $\Bar{\mathcal{X}}^{(i)}$, and their solutions are then concatenated through inverse transform. This treatment reduces the computational cost of tensor completion and enable us to perform it on large-scale data set by computing in parallel.

Like the product of two matrices, the product of two tensors induced by linear transform can be defined analogously.
\begin{definition}(t-product \citep{lu2019low,song2020robust})\label{tprod}
Given any invertible transform $\mathscr{L}$, and $\mathcal{A}\in\mathbb{R}^{n_1\times n_2\times n_3}$ and $\mathcal{B}\in\mathbb{R}^{n_1\times n_2\times n_3}$, then the transform based t-product denoted as $\mathcal{C}=\mathcal{A}*_{\mathscr{L}}\mathcal{B}$, is defined such that $\Bar{\mathbf{C}}^{(i)}=\Bar{\mathbf{A}}^{(i)}\Bar{\mathbf{B}}^{(i)}$, for $i=1,\dots,n_3$.
\end{definition}
Note that $\Bar{\mathbf{C}}^{(i)}=\Bar{\mathbf{A}}^{(i)}\Bar{\mathbf{B}}^{(i)}$ can be denoted compactly using the block diagonal matrix.
% , i.e., equivalent to $\widetilde{\mathbf{C}}=\widetilde{\mathbf{A}}\widetilde{\mathbf{B}}$. 
 Def. \ref{tprod} implies that the transform induced t-product can be performed by the matrix-matrix product in the transform domain. In the following sections, we will adopt this t-product to achieve some important tensor operations in the graph spectral domain.

\subsection{Graph-regularized low-rank tensor learning for traffic speed kriging}

Spatiotemporal traffic speed kriging can be treated as a special data imputation problem on a pre-specified graph so that we can model it through a tensor completion/learning method. Specifically, given an observation speed matrix $\mathbf{Z}\in\mathbb{R}^{(IK)\times J}$ whose rows correspond to $IK$ uniformly distributed time points with $I$ time intervals per day within $K$ days, and columns correspond to sensors at $J$ locations, the objective of spatiotemporal kriging in this paper can be described as learning the graph signals (node values) at unmeasured locations (entire columns) and missing time intervals (entire rows), given the observations and graph structure. This can be formulated as maximum of a posterior probability (MAP): 
\begin{equation}
\hat{\mathbf{Z}}=\arg\max_\mathbf{Z}\operatorname{P}(\mathbf{Z}|\mathbf{Z}_\Omega,\mathscr{G}_r,\mathscr{G}_c),
\label{MAP}
\end{equation}
where $\operatorname{P}(\cdot|\cdot)$ is the conditional probability function, $\Omega$ is the index where rows and columns are measured. $\mathscr{G}_r$ and $\mathscr{G}_c$ are prior information about rows and columns, i.e., temporal and spatial correlations in our case.
From the perspective of tensor completion model, MAP is equivalent to the following form:
\begin{equation} 
        \begin{aligned}
        \min_{\mathcal{X}}~&\Vert \mathcal{X}\Vert_{\text{norm}}+\mathscr{R}_s(\mathcal{X})+\mathscr{R}_t(\mathcal{X}),  \\ 
        \text {s.t.}~&\left\{\begin{array}{l}
        \mathbf{P}\odot\mathbf{Z}=\mathbf{P}\odot\mathbf{T}, \\
        \mathcal{X}=\mathscr{T}(\mathbf{Z}),\\
        \end{array}\right. \\
        \end{aligned}
    \label{model}
\end{equation}
where $\Vert \mathcal{X}\Vert_{\text{norm}}$ is the tensor norm of $\mathcal{X}$ to approximate the tensor rank, $\mathscr{R}_s(\cdot),\mathscr{R}_t(\cdot)$ are spatial and temporal consistency regularization, respectively. $\mathbf{P}$ is an indicating matrix whose entry is 1 when this data point is observed and 0 otherwise. $\mathbf{T}$ is the partially observed matrix, $\odot$ is the Hadamard product. Note that $\mathscr{T}(\cdot):\mathbb{R}^{(IK)\times J}\rightarrow\mathbb{R}^{I\times J\times K}$ denotes a forward tensorization operator that converts the spatiotemporal matrix into a third-order tensor of shape $(\text{time of day}\times\text{locations}\times\text{days})$ by stacking the 'day' dimension \citep{chen2021Autoregressive}. Conversely, the inverse matricization operator is denoted by $\mathscr{T}^{-1}(\cdot)$.

The first term in Eq. \eqref{model} encourages the tensor norm \citep{kolda2009tensor} to be small (or rank to be low) which serves as a global soft constraints. Low-rank (small-norm) pattern can capture inherent redundancy and linearity of traffic data. In this work, we establish a new temporal graph Fourier transform based tensor nuclear norm (t-TNN). Moreover, it is not hard to find that the unmeasured rows and columns of a time-space matrix is manifested as some missing 'slices' of a tensor. Purely low-rank model is incapable of dealing with such 'missing slices' of tensor, because it is agnostic to the positional information \citep{nie2022truncated}. Therefore, the other two terms are designed for encoding side information about rows and columns, which are in fact local hard constraints.
Overall, Eq. \eqref{model} is a simultaneous structured model that encourages sparsity of $\mathcal{X}$ in the tensor singular value vector outer product space, and column-row wise sparsity of $\mathbf{Z}$ in the spatial and temporal Laplacian eigenspaces. 
The proposed method mainly features tensor norm minimization and spatial-temporal regularization, which favors a globally low-rank and locally smooth solution.

\begin{figure}[!htb]
  \centering
  \includegraphics[scale=0.365]{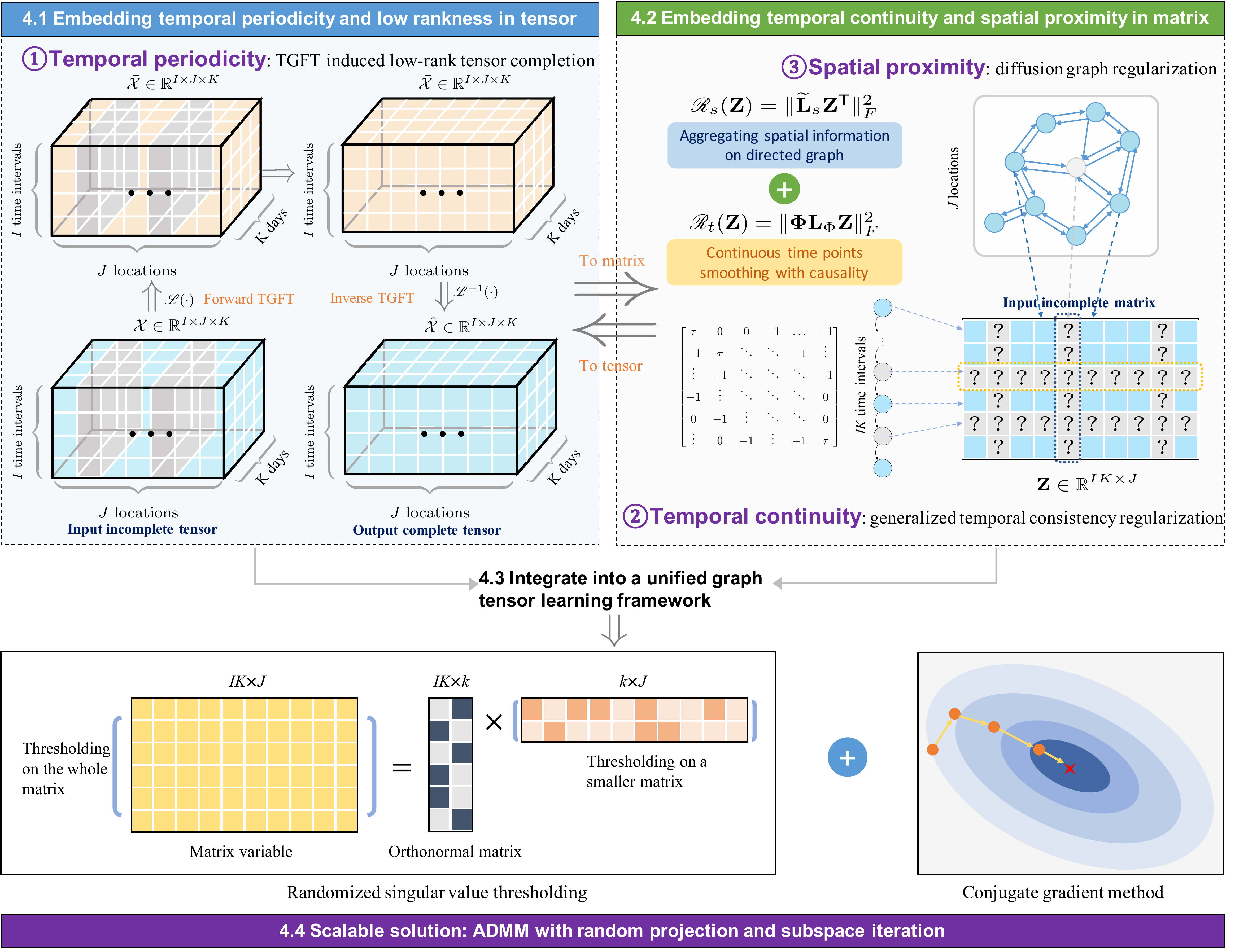}
  \caption{Illustration of proposed LETC framework. The whole method features three parts: embedding temporal periodicity and low rankness with temporal graph Fourier transform (in \ref{TGFT_subsec}), modeling temporal continuity with a generalized temporal consistency regularization, and encoding spatial proximity with diffusion graph regularization (in \ref{TC_subsec} and \ref{SP_subsec}). All three correlations are modeled by three types of graph Laplacian and integrated into a unified tensor learning model (in \ref{SGET_subsec}) and we obtain an efficient solution in \ref{admm_solve} for large-scale problems.}
  \label{method_fig}
\end{figure}

\section{Methodology}\label{methodology}

In this section, we first introduce a novel temporal graph Fourier transform (TGFT) on the basis on periodic graph Laplacian and the corresponding linear transform-induced tensor nuclear norm. Since then, we establish the temporal and spatial regularization with another two forms of graph Laplacian by generalized temporal consistency regularization (GTCR) and diffusion graph regularization (DGR), respectively. Finally, we formulate the integrated tensor completion model for traffic speed kriging and solve it in a highly scalable manner using the alternating direction method of multipliers (ADMM) with some valuable numeric techniques, i.e., randomized tensor singular value  thresholding (r-tSVT) and conjugate gradient (CG). The proposed model is referred to as 
Laplacian enhanced tensor completion (LETC) model. The overall framework is shown in Fig. \ref{method_fig}.
% Spatiotemporal Graph Embedded Tensor (LETC) learning. The proposed framework is shown in Fig. \ref{method_fig}.

\subsection{Embedding temporal periodicity and low rankness with temporal graph Fourier transform}
\label{TGFT_subsec}
As described in Section \ref{Notations and Problem Definitions}, invertible linear transform serves as an important role in scalable tensor completion methods. The most common transforms such as FFT, DCT, are adopted for image or video data \citep{lu2016tensor,lu2019low}. Nevertheless, how to determine a knowledge-based transform for traffic data is still worthy to be considered. In the existing literature of transportation, \cite{chen2021scalable} used a data-driven unitary transform which is obtained by the SVD of day-mode unfolding; \cite{deng2021graph} performed GFT along the 'location' dimension to extract the graph information. Recall that we arrange the data in a tensor of $(\text{time of day}\times\text{locations}\times\text{days})$ format, if the linear transform is performed on the 'days' mode, dealing with each daily subproblem of transformed tensor separately may ignore the temporal periodicity of traffic data.

To take advantage of prior knowledge about spatiotemporal traffic data, we establish a novel temporal graph Fourier transform (TGFT) based on classical GFT. GFT has been widely applied into the graph spectral analysis and the emerging graph neural networks to capture the non-Euclidean relationships on graphs \citep{kipf2016semi}. The basic GFT is defined as:
\begin{definition}(GFT \citep{shuman2013emerging}) Given a graph signal $\mathbf{f}\in\mathbb{R}^{n_3}$ and the graph Laplacian $\mathbf{Lap}$, the GFT of $\mathbf{f}$ is 
\begin{equation}
    \Bar{\mathbf{f}} = \operatorname{GFT}(\mathbf{f})=\mathbf{U}\mathbf{f},
\label{gft_cal}
\end{equation}
and the inverse operator is 
\begin{equation}
    f = \operatorname{iGFT}(\Bar{\mathbf{f}})=\mathbf{U}^{\mathsf{H}}\Bar{\mathbf{f}},
\end{equation}
where the transform matrix operator $\mathbf{U}$ is obtained by the eigenvalue decomposition $\mathbf{Lap}=\mathbf{U}\Lambda\mathbf{U}^{\mathsf{H}}$.
\end{definition}

Since the Laplacian matrix can be viewed as the Laplace operator on graphs, GFT is in fact a projection using the eigenvectors of the Laplacian matrix as the basis vectors \citep{sandryhaila2013discrete}. Similar to DFT, GFT is decomposing a graph signal into a series of graph signals with different degrees of smoothness. Therefore, graph signals in spectral domain also reflect different frequency components, and the eigenvalue can be analogous to frequency. The larger the eigenvalue, the greater the graph signal fluctuation.

To embed the multi-scale relationships of days and weeks, we establish the TGFT based on a temporal adjacent matrix that describes the day-to-day and day-of-week periodicity of 'day'-mode slices. Since the eigenvalue decomposition of a symmetric matrix always leads to a set of orthogonal basis \citep{golub2013matrix}, we preserve the unitary property of GFT by establishing a symmetric adjacent matrix. Given a multivariate time series within $D$ days, the temporal adjacent matrix $\mathbf{A_t}\in\mathbb{R}^{D\times D}$ is given as:
% \begin{equation}\textcolor{red}{
%      a^t_{ij}=\left\{\begin{array}{l}
%         1 \text{ if } \vert i-j\vert=1 \text{ or } \vert i-j\vert=7,\\
%         0 \text{ otherwise}.\\
%         \end{array}\right. \\}
% \end{equation}
\begin{equation}
\begin{aligned}
\mathbf{A}_t&=\mathbf{I}_D+\mathbf{P}_1+\mathbf{P}_1^{\mathsf{T}}+\sum_{n=1}^{m}\left(\mathbf{P}_T(n)+\mathbf{P}_T(n)^{\mathsf{T}}\right),\\
\mathbf{P}_1&= 
    \left[
    \begin{array}{c|c} 
    \mathbf{0}_{1\times(D-1)} & 0 \\
    \hline
    \omega_1\mathbf{I}_{D-1} & \mathbf{0}_{(D-1)\times1} \\ 
    \end{array}
    \right],\\
\mathbf{P}_T(n)&= 
    \left[
    \begin{array}{c|c} 
    \mathbf{0}_{nT\times(D-nT)} & \mathbf{0}_{nT\times nT} \\
    \hline
    \omega_T\mathbf{I}_{D-nT} & \mathbf{0}_{(D-nT)\times nT} \\ 
    \end{array}
    \right],\\
\end{aligned}   
\label{A_t}
\end{equation}
where $T$ is the length of a period and $m$ is the total number of periods within $D$ days, $\omega_1$ and $\omega_T$ denote the similarity weights of day-to-day and day-of-week patterns respectively, which can be chosen according to the daily variations of traffic states. In practice, ones can possibly consider setting a decaying factor $\beta\in(0,1)$ for $\omega_T$ to take into account the effect of long time sequence, e.g., $\omega_T(n)=\beta(1-\beta)^n\omega_T(1)$. Moreover, consider the fact that traffic patterns could be different among weekends and weekdays, we can incorporate this heterogeneity by simply modifying $\mathbf{P}_1$ as follows:
\begin{equation*}
\begin{aligned}
    \mathbf{P}_1'&= 
    \left[
    \begin{array}{c|c} 
    \mathbf{0} & 0 \\
    \hline
    \operatorname{BDiag}(\mathbf{H}_{T\times T}) & \mathbf{0} \\ 
    \end{array}
    \right],\\
    \mathbf{H} &= \operatorname{Diag}(\underbrace{\omega_1,\dots,\omega_1}_{T-2T_w+1}\underbrace{\omega_2,\dots,\omega_2}_{2T_w-1}).
\end{aligned}  
\end{equation*}
where $\operatorname{BDiag}(\cdot)$ constructs a block diagonal matrix, $T_w$ is the number of different daily patterns, e.g., $T_w=2$ for weekends, and $\omega_2$ is the corresponding weights.

\begin{figure}[htbp!]
    \centering
    \subfigure[Graphical model for periodic time series (self-loops are omitted for clarity).]{
    \centering
    \includegraphics[scale=1.2]{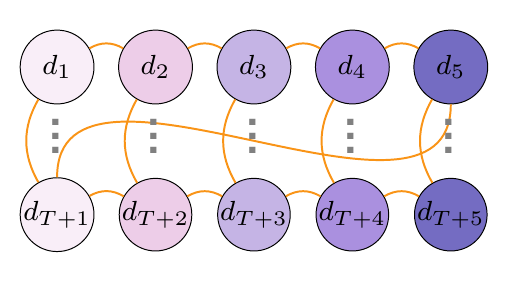}
    }
    \subfigure[Adjacent matrix with $D=6$ and $T=3$. The colors denote different weights for day-to-day and day-of-week patterns.]{
    \centering
    \includegraphics[scale=1.2]{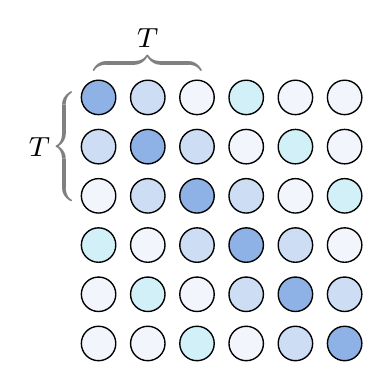}
    }
    \caption{Temporal undirected graphical models with  period $T$. Each node represents the data of a day.}
    \label{periodic_graph}
\end{figure}

Eq. \eqref{A_t} depicts the graphical model in Fig. \ref{periodic_graph} (a), and it has intuitive explanation: the first term $\mathbf{I}_D$ represents the self-loop of each node itself, $\mathbf{P}_1$ denotes the first-order neighborhood, i.e., the adjacent two days, and $\mathbf{P}_T(n)$ signifies the neighbors in $n$-th period, which forms the $nT$-th lower main diagonal. The transpose terms are added to make $\mathbf{A}_t$ undirected and symmetric. Fig. \ref{periodic_graph} (b) gives an example with $D=6$ and $T=3$.

$\mathbf{A}_t$ directly connects two 'day'-mode slices with weight being one if two days are consecutive or the same day of a week. After computing $\mathbf{A}_t$, we can obtain a temporal periodic Laplacian $\mathbf{L}_t$ using Eq. \eqref{Laplacian}. Since $\mathbf{A_t}$ is nonnegative and symmetric, then the Laplacian $\mathbf{L_t}$ is thus symmetric and semi-positive definite. Moreover, as $\mathbf{L_t}$ is real-valued, the resulted eigendecomposition $\mathbf{L_t}=\mathbf{U_t}\Lambda\mathbf{U_t}^{\mathsf{T}}$ is also real-valued. Different from FFT that operates in the complex domain, TGFT enables the tensor operations in real-valued graph spectral domain.
It is obvious that the TGFT operator $\mathbf{U_t}$ holds the property in Eq. \eqref{transform} thus the forward and backward transform can be implemented for tensors using Eq. \eqref{forward} and Eq. \eqref{inverse}.

Each tubal fiber $\mathcal{X}(i,j,:)$ of the transformed tensor describes the graph signal at the $i$-th time interval in the $j$-th location within $K$ days, and all fibers are controlled by a single temporal Laplacian. After performing TGFT, we are allowed to tackle the transformed matrix optimization subproblem for each day (which will be discussed in Section \ref{X_solve}) and the 'day-to-day' and 'day-of-week' correlations (i.e., temporal periodicity) are encoded in the whole transformed tensor (see Fig. \ref{method_fig}).

Compared with GFT performed on the 'location' mode in \citep{deng2021graph}, when there exists a large number of sensors at network level $(J\gg I,K)$, the proposed TGFT can split the original problem into much fewer ($K$) subproblems with well scalability, rather than $J$ in original GFT.

Once we assign TGFT as the linear transform in Eq. \eqref{transform}, the corresponding transform based tensor nuclear norm (t-TNN) and tensor-tensor product (t-product) can be given as follows.
\begin{definition}(t-TNN \citep{lu2019low,song2020robust}) The t-TNN of a tensor $\mathcal{A}\in\mathbb{R}^{n_1\times n_2\times n_3}$ is the sum of the nuclear norms of all frontal slices of $\Bar{\mathcal{A}}$, i.e., 
    \begin{equation}\label{tTNN_def}
    \Vert\mathcal{A}\Vert_{t*}=\sum_{i=1}^{n_3}\Vert\Bar{\mathbf{A}}^{(i)}\Vert_{*}=\Vert\widetilde{\mathbf{A}}\Vert_{*},
    \end{equation}
where $\Vert\cdot\Vert_{*}$ denotes the matrix nuclear norm, $\widetilde{\mathbf{A}}\in\mathbb{R}^{n_1n_3\times n_2n_3}$ is the block diagonal matrix of $\Bar{\mathcal{A}}$ with its $i$-th block on the diagonal as the $i$-th frontal slice of $\Bar{\mathcal{A}}$:
\begin{equation*}\widetilde{\mathbf{A}}=
\begin{bmatrix}
\Bar{\mathbf{A}}^{(1)} &  &  &  \\
 & \Bar{\mathbf{A}}^{(2)} &  &  \\
 &  & \ddots &  \\
 &  &     & \Bar{\mathbf{A}}^{(n_3)}
\end{bmatrix}.
\end{equation*}
\end{definition}
It has been proved that the above t-TNN is the convex envelope of the tensor tubal rank \citep{lu2019low,song2020robust}.
The motivation behind using t-TNN in this work is to preserve the low-rank property of traffic data tensor as well as to reduce the computational cost as much as possible. Traditionally, the tensor rank is approximated by the sum of nuclear norm or Schatten $p$-norm of all unfolding matrices \citep{chen2021Autoregressive,nie2022truncated}, and the SVD needs to be conducted on all unfolding matrices. Consider that we assume $n_2\gg n_1>n_3$ for large-scale problems, this yields time complexity of $\mathcal{O}(n_1^2n_2n_3+n_1n_2n_3\min\{n_2,n_1n_3\}+n_1n_2n_3^2)$ in total. While under the definition of t-TNN, SVD only needs to be computed on $n_3$ frontal slices of size $n_1\times n_2$ and the time complexity reduces to $\mathcal{O}(n_1^2n_2n_3)$.

Induced by TGFT, we are allowed to perform t-TNN optimization in the domain of graph spectral, and obtain the output tensor in original space-time domain through inverse TGFT. This can be viewed as a feature augmentation by coordinate transformation, where a new subspace benefiting low-rankness is chosen. Additionally, by embedding the relationships between data of days
, 'day'-mode slices with similar components on graph Fourier basis are encouraged to share close patterns. 

Performing tensor nuclear norm minimization (i.e., Eq. \eqref{tTNN_def}) can solve the missing data imputation problem with element-like or fiber-like missing patterns \citep{nie2022truncated}. However, when there exists entire missing slices of a tensor, using pure low-rank assumption could be undesirable as the original tensor structure is not sensitive to the permutation of tensor fibers. Therefore, the key to model kriging as a tensor completion problem is to exploit side information as physic guidance. In the following sections, we construct temporal continuity and spatial proximity regularization by developing a generalized temporal consistency regularization and a novel diffusion graph regularization respectively.

\subsection{Modeling time series continuity with generalized temporal consistency regularization}\label{TC_subsec}

In time dimension, we establish the 'time-of-day' relationship (i.e., temporal continuity) explicitly by extra regularization, because the 'day-to-day' and 'day-of-week' correlations are already embedded in the transformed tensor by TGFT. For traffic data imputation and forecasting problems, some elaborate time series models such as vector autoregressive \citep{yu2016temporal,chen2021Autoregressive}, long-short term memory neural network \citep{yang2021real}, quadratic variation \citep{chen2021scalable}, are adopted to capture complex and long-term temporal dynamics. While at the same time, more computational burdens are brought to train such temporal models.
% Nevertheless, \cite{wu2021spatial} found that the temporal receptive field (temporal relationship) only has marginal impacts on the kriging performance. 
 Traditional Toeplitz temporal regularization fails to capture multi-hop dependencies. Besides, recent study \citep{chen2022laplacian} has demonstrated that temporal relationships can be represented by graph of time series, and the Laplacian kernel can be used as a local smoothness regularization, which features both accuracy and efficiency. In their work, undirected and circulant graphs are adopted to characterize the time series dependencies.

% Therefore, we utilize a simple Toeplitz matrix to consider the temporal continuity and keep the scalability of model at the same time. The first-order and $n$-th ($n>1$) order Toeplitz matrices are defined as: 

% \begin{equation}
% \begin{array}{cc}
% \mathbf{E_1}=
% \begin{bmatrix}
% 1 & -1 & 0 & \hdots & 0 &0&0\\
% 0 & 1 & -1 & \hdots & 0&0&0\\
% \vdots & \vdots & \ddots & \ddots &\vdots&\vdots&\vdots\\
% \vdots & \vdots & \vdots & \ddots & \ddots & \vdots &\vdots\\
% 0 & 0 & 0 & 0 & 1 & -1 & 0\\
% 0 & 0 & 0 & 0 &0& 1 &-1\\
% 0 & 0 &  0  & 0 & 0 & 0 &1
% \end{bmatrix}_{I\times I},
% \mathbf{E_n}=
% \begin{bmatrix}
% 1 &\hdots & -n & 1 & 1 & \hdots & 0 \\
% 0 & 1 & \hdots & -n & 1 & \hdots & 0\\
% \vdots & \vdots & \ddots & \ddots &\ddots &\ddots& \vdots \\
% 0 & 0 &  0  & 1 & 1 & -n & 1\\
% % \vdots & \vdots & \ddots & \ddots &\ddots &\ddots& \vdots \\
% 0 & 0 &  0  & 0 & 1 & 1 & -n \\
% \vdots & \vdots & \vdots & \vdots & \vdots & \ddots & \vdots\\
% 0 & 0 &  0  & 0 & 0 & 0 & 1
% \end{bmatrix}_{I\times I}.
% \end{array}
% \end{equation}

% Note that the order $n$ should be even. It can be found that $\mathbf{E_1}$ is an upper triangular matrix with the main diagonal given by ones and the first upper diagonal given by negative ones. Similarly, $\mathbf{E_n}$ is also an upper triangular matrix with the first $[1,\dots, n/2]$ and $[n/2+1,\dots,n+1]$ upper diagonals given by ones, and the $(n/2+1)$-th upper diagonal being $-n$. This matrix can be calculated and stored in advance according to the kriging time span. 

\begin{figure}[htbp!]
    \centering
    \subfigure[Directed and acyclic graph with $\tau=1$]{
    \centering
    \includegraphics[scale=1.5]{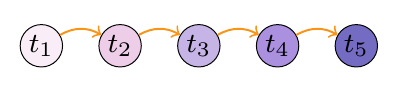}
    }
    \subfigure[Directed and acyclic graph with $\tau=2$]{
    \centering
    \includegraphics[scale=1.5]{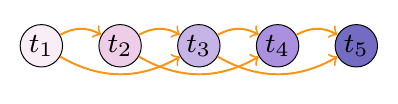}
    }
    \caption{Temporal graphical models with directed and acyclic edges. Each node represents a time point of a series. In these examples, $\mathcal{C}(\mathbf{k}_{1})$ and $\mathcal{C}(\mathbf{k}_{2})$ are their respective directed Laplacian matrices.}
    \label{directed_graph_tau}
\end{figure}

Inspired by \citep{chen2022laplacian,singh2016graph}, we define a general form of temporal consistency regularization in the context of directed graph Laplacian, which is a general case of some temporal smoothness penalty, e.g., Toeplitz matrix. To reflects the causality of time series, directed temporal graph representing the dependent relationships of time series is constructed by first defining a temporal kernel $\mathbf{k}_{\tau}$ parameterized by kernel size $\tau$:
\begin{equation}
\mathbf{k}_{\tau}=(\tau,\underbrace{-1,\dots,-1}_{\tau},0,\dots,0)^{\mathsf{T}}\in\mathbb{R}^{T},
\end{equation}
where $\tau$ determines the window size of temporal smoothness. Then its circulant matrix $\mathcal{C}(\mathbf{k}_{\tau})$ is in fact the Laplacian \citep{singh2016graph} of a directed and unweighted temporal graph (see Fig. \ref{directed_graph_tau}):
\begin{equation}
    \mathcal{C}(\mathbf{k}_{\tau})=\mathbf{L}_{\Phi}=
    \begin{bmatrix}
    \tau & 0 & 0 & -1 & \hdots & -1\\
    -1 & \tau & \ddots & \ddots & -1 & \vdots\\
    \vdots & -1 & \ddots & \ddots &\ddots&-1\\
    -1 & \vdots & \ddots & \ddots & \ddots & 0 \\
    0 & -1 & \vdots & \ddots & \ddots & 0 \\
    \vdots & 0 & -1 & \vdots & -1 & \tau \\
    \end{bmatrix}_{T\times T}.
\end{equation}

Finally, given an input graph signal $\mathbf{Z}\in\mathbb{R}^{T\times N}$, GTCR is formulated as:
\begin{equation}
{\mathscr{R}_t}(\mathbf{Z}) = \Vert\mathbf{\Phi}\mathbf{L}_{\Phi}\mathbf{Z}\Vert_F^2,
\label{temporal regularization}
\end{equation}
where $\mathbf{\Phi}$ is a truncation operator defined as:
\begin{equation}
    \mathbf{\Phi} = 
    \begin{bmatrix}
        \mathbf{0}_{(T-\tau)\times\tau} & \mathbf{I}_{T-\tau}
    \end{bmatrix}.
\end{equation}
This operator is used to remove the first $\tau$ rows in $\mathcal{C}(\mathbf{k}_t)$, thereby avoiding the loops in the temporal graph.

\begin{remark}
Eq. \eqref{temporal regularization} is a general form of temporal regularization and can degrade into many widely used regularization in the literature. For example, 
\begin{enumerate}
    \item when $\tau=1$ and $\mathbf{\Phi}=\mathbf{I}_T$, it yields the traditional one-order Toeplitz temporal regularization like in \citep{wang2018traffic};
    \item when $\mathbf{L}_{\Phi}'=\mathbf{L}_{\Phi}\mathbf{L}_{\Phi}^{\mathsf{T}}$ and $\mathbf{\Phi}=\mathbf{I}_T$, it leads to the circulant and symmetric Laplacian in \citep{chen2022laplacian};
    \item when $\tau=1$ and $\mathbf{\Phi}=[\mathbf{0}_{(T-1)\times1}~\mathbf{I}_{T-1}]$, it becomes the quadratic variation norm in \citep{chen2021scalable}.
\end{enumerate}
\end{remark}
The function of GTCR is to make consecutive time points within the kernel size (time window) to be smooth, and Eq. \eqref{temporal regularization} only imposes this smoothness penalty along the direction of time lapse, without assumption of cyclicity.
The larger the kernel size, the longer the temporal receptive field. This model-free technique could provide a simple but efficient tool for modeling varying degrees of temporal dependencies along 'time interval' dimension.

\subsection{Modeling spatial proximity with diffusion graph regularization}\label{SP_subsec}
\begin{figure}[htbp!]
    \centering
    \includegraphics[scale=1.5]{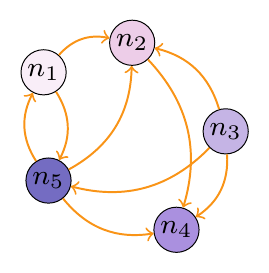}
    \caption{Spatial graphical models with directed edges. Each node represents a sensor on the network and edge direction reflects the moving direction of traffic flow.}
    \label{directed_graph_space}
\end{figure}

As for spatial dimension, symmetric graph Laplacian derived from symmetric adjacent matrix is the most commonly used spatial regularization in the literature to perform kriging \citep{bahadori2014fast,kalofolias2014matrix,rao2015collaborative,wang2018traffic,zhang2020network,yang2021real,lei2022bayesian}. Most of these approaches are limited to the analysis of traffic data lying on undirected graphs. However, traffic network is in essence a directed graph whose edge reflects the directional movement of traffic flow. To operate on such directed graphs (see Fig. \ref{directed_graph_space}), existing works usually construct a symmetric adjacent matrix by bool sum or simple average of the asymmetric adjacent and its transpose.

Despite of its simplicity, recent graph neural network studies \citep{li2017diffusion,wu2019graph} reveal that such symmetrization could bring in extra errors because of the ignore of traffic operation directions.
To tackle this issue, we propose a new diffusion graph regularization that can be applied for directed graph structures.

Since Eq. \eqref{Laplacian} is defined for symmetric adjacent matrix and Laplacian, the first modification is to use the in-degree or out-degree matrix \citep{chung2005laplacians} of asymmetric adjacent matrix to compute Laplacian. Therefore, the resultant graph Laplacian is also asymmetric. Then, based on physical sensor graphs, we assume that the spot speed recorded by a static detector (e.g., loop sensor) tends to be similar with the values of its upstream or downstream neighbor sensors that determined by the graphical structure of traffic network. Thus we treat all sensors as nodes $\mathscr{V}=\{v_1,\dots,v_n\}$ and the relational connectivity of them as directed edges with asymmetric weight matrix $\mathbf{A_s}$ representing similarity intensities.

In this work, we calculate the weights matrix by Gaussian kernel function:
\begin{equation}\label{Gaussian}
a^s_{ij}=\left\{\begin{array}{l}
\operatorname{exp}\left(-\left(\frac{\operatorname{dist}(v_i,v_j)}{\sigma}\right)^2\right)~\text{if a directed edge exists from}~ v_i~\text{to}~v_j,\\
0 ~\text{otherwise},
\end{array}\right.
\end{equation}
where $\operatorname{dist}(v_i,v_j)$ is the physical distance between sensor $v_i$ and $v_j$, which can obtained from the travel distance or the coordinate position; $\sigma$ is a kernel hyper-parameter which can be chose as the standard deviation of the distance matrix.

By performing Eq. \eqref{Laplacian}, the in-degree or out-degree graph Laplacian of sensor graph is derived by $\mathbf{L_s}=\operatorname{Diag}(\sum_{i~\text{or}~j}a^s_{ij})-\mathbf{A_s}$, and finally the proposed DGR for spatial modeling can be formulated as following:
\begin{equation}
    \mathscr{R}_s(\mathbf{Z}) = \sum_{j}^J\Vert \frac{1}{d_j}\sum_{i}a^s_{ij}\mathbf{z}_i-\mathbf{z}_{j}\Vert_2^2=\Vert\widetilde{\mathbf{A}}_s\mathbf{Z}^{\mathsf{T}}-\mathbf{Z}^{\mathsf{T}}\Vert_F^2=\Vert\widetilde{\mathbf{L}}_s\mathbf{Z}^{\mathsf{T}}\Vert_F^2,
\label{spatial regularization}
\end{equation}
where $d_j$ is the $j$-th diagonal element of in-degree (or out-degree) matrix $\mathbf{D}$, $\widetilde{\mathbf{A}}_s=\mathbf{D}^{-1}\mathbf{A}_s$ is the normalized adjacent matrix, and $\widetilde{\mathbf{L}}_s=\mathbf{I}-\mathbf{D}^{-1}\mathbf{A}_s$ is the random-walk normalized Laplacian.

For a graph $\mathcal{G}$ with non-negative weights and no isolated nodes, $\mathbf{D}^{-1}\mathbf{A}_s$ becomes the transition matrix of a random walker \citep{chung2005laplacians} on $\mathcal{G}$, thereby producing a diffusion process.
Note that Eq. \eqref{spatial regularization} is a natural generalization of graph Laplacian regularization on directed graphs. By letting $\widetilde{\mathbf{L}}=\mathbf{L}\mathbf{L}^{\mathsf{T}}$, DGR yields exactly the original form for undirected graphs, e.g.,  $\operatorname{Tr}(\mathbf{Z}\mathbf{L}\mathbf{Z}^{\mathsf{T}})$.

By extending the spatial regularization to directed graphs, we can shed light on the natural connections between the proposed DGR and other diffusion kernels that are successfully applied in graph convolution based models \citep{kondor2002diffusion,xu2020graph}. Note that the core of Eq. \eqref{spatial regularization} is the \textit{forward propagation} operation, i.e., $\widetilde{\mathbf{A}}_s\mathbf{Z}^{\mathsf{T}}$.
Formally, we can re-write it in a more general view:
\begin{equation}
    f(\mathbf{Z}^{\mathsf{T}}|h(\widetilde{\mathbf{A}}_s))=h(\widetilde{\mathbf{A}}_s)\mathbf{Z}^{\mathsf{T}},
\end{equation}
where $h(\cdot):\mathbb{R}^{N\times N}\rightarrow\mathbb{R}^{N\times N}$ is a diffusion function that determines the behaviors of random walker on graphs. Three specific formulations of diffusion functions are discussed in this paper, they are:
\begin{itemize}
    \item High-order diffusion kernel: 
    \begin{equation}
        h(\widetilde{\mathbf{A}}_s) = \widetilde{\mathbf{A}}_s^K,
    \end{equation}
    where $K$ is the diffusion steps;
    \item Personalized Page Rank (PPR) kernel \citep{page1998pagerank}:
    \begin{equation}
        h(\widetilde{\mathbf{A}}_s) = \sum_{K=0}^{\infty}\alpha(1-\alpha)^K\widetilde{\mathbf{A}}_s^K,
    \end{equation}
    where $\alpha\in(0,1)$ is the teleport probability;
    \item Heat diffusion kernel \citep{kondor2002diffusion}:
    \begin{equation}
        h(\widetilde{\mathbf{A}}_s) = \sum_{K=0}^{\infty}e^{-t}\frac{t^k}{k!}\widetilde{\mathbf{A}}_s^K,
    \end{equation}
    where $t$ is the diffusion time.
\end{itemize}

It should be noted that in existing studies diffusion kernels are mostly used as an aggregation operation combined with learnable parameters to serve as underlying functions of graph machine learning models \citep{li2017diffusion,wu2019graph}. Few of them have paid attention to its role in acting as a graph constraint. 

By introducing the above diffusion kernels in Eq. \eqref{spatial regularization}, DGR benefits from the exploration of high-order and hierarchical structures, which can explain most of the success of an effective kriging model.

Furthermore, having carefully inspected the roles of DGR for modeling spatial-temporal traffic data on graphs, we find that there are two physics-informed ways to understand the behaviors of it from the perspective of: 1) manifold learning, and 2) graph diffusion equation, which are less discussed in previous studies based on typical graph Laplacian constraints.

First, let us start with the well-known Laplacian regularized least-square problems (LapRLS) \citep{belkin2004semi}:
\begin{equation}
    f^*=\arg \min_{f\in\mathcal{H}}\ell(f(\mathbf{x})-\mathbf{y}) + \gamma \Vert f\Vert_{\mathcal{H}}^2,
\label{LapRLS}
\end{equation}
where $\mathbf{x}$ and $\mathbf{y}$ are training data and labels, $\Vert f\Vert_{\mathcal{H}}^2$ is a Laplacian kernel induced norm in $\mathcal{H}$, and $\ell$ is a loss function. LapRLS attempts to learn a function $f$ from among a hypothesis space of functions $\mathcal{H}$, with the aim to predict the true labels as well as minimizing the Laplacian manifold regularization. Consider the fact that spatiotemporal traffic data tends to be low-rank in space-time domain and \textit{low-frequency} in spectral domain \citep{liu2022recovery,chen2022laplacian}, we can instantiate Eq. \eqref{LapRLS} from a matrix view with the following objective:
\begin{equation}
    \min_{\Bar{\mathbf{X}}} \Vert\mathbf{P}\odot(\Bar{\mathbf{X}}-\mathbf{X})\Vert_{\mathbf{D}}^2 + \operatorname{Tr}(\Bar{\mathbf{X}}^{\mathsf{T}}\mathbf{L}\Bar{\mathbf{X}}),
\label{dirichlet}
\end{equation}
where $\Vert\cdot\Vert_{\tilde{\mathbf{D}}}$ means the augmented degree matrix induced norm denoted by $\Vert\mathbf{x}\Vert_{\tilde{\mathbf{D}}}=\sqrt{\mathbf{x}^{\mathsf{T}}\tilde{\mathbf{D}}\mathbf{x}}$ \citep{nt2019revisiting}. The first term measures the reconstruction loss on the observed data weighted by the importance of different nodes. The second term is the graph Dirichlet energy which represents the smoothness of graph signals. By optimizing Eq. \eqref{dirichlet} we can obtain an approximated graph signal with low-frequency components. The closed form solution is given by (detailed derivation is given in \nameref{Appendix D.}):
\begin{equation}
    \Bar{\mathbf{X}}=(\mathbf{I}+\widetilde{\mathbf{L}}_s)^{-1}\mathbf{X}.
\end{equation}

If we consider the first-order Taylor approximation of it, we can obtain:
\begin{equation}
    \Bar{\mathbf{X}}:=(\mathbf{I}-\widetilde{\mathbf{L}}_s)\mathbf{X}=\widetilde{\mathbf{A}}_s\mathbf{X},
\end{equation}
which leads to the propagation term in Eq. \eqref{spatial regularization} exactly. In this sense, we can find that DGR explicitly encourages a low-frequency solution in graph spectral domain which is desirable for spatiotemporal traffic data and can be treated as a 'low-pass' filter \citep{nt2019revisiting}.

Second, we can present an exposition from another perspective by revisiting the (heat) diffusion equation on graphs \citep{chamberlain2021grand}. Consider a diffusion process that depicts the evolution of graph signal matrix $\mathbf{X}$ over time:
\begin{equation}
    \frac{\partial \mathbf{X}(t)}{\partial t}=\operatorname{div}(\mathbf{G}(\mathbf{X}(t),t)\odot\nabla\mathbf{X}(t)),
\label{diff_eq}
\end{equation}
where $\mathbf{G}$ is chosen to ensure $\mathbf{A}\odot\mathbf{G}$ to be a right random walk matrix, and $\operatorname{div}(\cdot),\nabla$ are the divergence and gradient operators respectively. We can discrete Eq. \eqref{diff_eq} with a relatively small step $h$ using explicit Euler scheme \citep{chamberlain2021grand} and plugging in the expressions of $\operatorname{div}(\cdot),\nabla$ on graphs:
\begin{equation}
    \mathbf{X}(kh)=\mathbf{X}((k-1)h)-h\widetilde{\mathbf{L}}_s\mathbf{X}((k-1)h):=\widetilde{\mathbf{L}}_s\mathbf{X}((k-1)h).
\label{euler}
\end{equation}
Eq. \eqref{euler} means that if we treat the data flows from one node to another one with a step-by-step propagation, multiplying $\widetilde{\mathbf{L}}_s$ with $\mathbf{X}$ denotes a forward diffusion of all data points on the graph at the current state. More importantly, 
by performing Eq. \eqref{euler} successively and filling the observed values to $\mathbf{X}$ at each step, which are already embedded in our ADMM algorithm in section \ref{admm_solve}, we allow the data propagates from observed ones to unobserved ones as well as among the unknown nodes, which forms a bounded (heat) diffusion process with finite steps on graph.

\begin{remark}
DGR can also be implemented on both directions of diffusion. For example, if we consider inflow and outflow at the same time, Eq. \eqref{spatial regularization} can be formulated as:
\begin{equation*}
\mathscr{R}_s(\mathbf{Z})=\Vert\widetilde{\mathbf{L}}_f\mathbf{Z}^{\mathsf{T}}+\widetilde{\mathbf{L}}_b\mathbf{Z}^{\mathsf{T}}\Vert_F^2.
\end{equation*}
In this case, $\widetilde{\mathbf{L}}_f=\mathbf{I}-\mathbf{A}_s/\operatorname{rowsum}(\mathbf{A}_s),\widetilde{\mathbf{L}}_b=\mathbf{I}-\mathbf{A}_s^{\mathsf{T}}/\operatorname{columnsum}(\mathbf{A}_s)$ are forward and backward transition Laplacian respectively. Additionally, DGR can be applied to many other graph-based tasks, e.g., network-wide traffic forecasting.
\end{remark}

The intuition behind Eq. \eqref{spatial regularization} is to first aggregate neighborhood features according to the information passing mechanism on graph, then impose a column-wise smoothness to make locally connected nodes have similar node values. This process shares similar ideas with the graph diffusion convolution network \citep{li2017diffusion,wu2019graph}, while we only consider one-step random walk. This operation can encode the local spatial dependency and impute the unknown node values by borrowing referable information from measured ones.

Moreover, one can find that both $\mathscr{R}_t(\mathbf{Z})$ and $\mathscr{R}_s(\mathbf{Z})$ has similar form of Laplacian representation. Both of them encourage column or row wise sparsity of $\mathbf{Z}$ in the spatial and temporal Laplacian eigenspaces, representing the relationships of sensors on two individual dimensions.

\subsection{Low-rank tensor learning with Laplacian enhancement}\label{SGET_subsec}
Having introduced all the components of model above, we now integrate the low-rank optimization part with spatial and temporal regularization in a unified tensor learning framework, making Eq. \eqref{model} the following specific form:

% \begin{equation} 
%     \begin{aligned}
%     \min_{\mathcal{X}}~&\Vert \mathcal{X}\Vert_{t*}+\frac{\lambda_1}{2}\operatorname{Tr}(\mathbf{Z}\mathbf{L_s}\mathbf{Z}^\mathsf{T})+\frac{\lambda_2}{2}\Vert\mathbf{E_nZ}\Vert_F^2  \\ 
%     &\text {s.t.}~ \mathbf{P}\odot\mathbf{Z}=\mathbf{P}\odot\mathbf{T}, \mathcal{X}=\mathscr{T}(\mathbf{Z}), \\
%     \end{aligned}
% \label{final model}
% \end{equation}
\begin{equation}
    \begin{aligned}
    \min_{\mathcal{X}}~&\Vert \mathcal{X}\Vert_{t*}+\frac{\lambda_1}{2}\Vert\widetilde{\mathbf{L}}_s\mathbf{Z}^{\mathsf{T}}\Vert_F^2+\frac{\lambda_2}{2}\Vert\mathbf{\Phi}\mathbf{L}_{\Phi}\mathbf{Z}\Vert_F^2  \\ 
    &\text {s.t.}~ \mathbf{P}\odot\mathbf{Z}=\mathbf{P}\odot\mathbf{T}, \mathcal{X}=\mathscr{T}(\mathbf{Z}), \\
    \end{aligned}
\label{final model}
\end{equation}
where $\lambda_1$ and $\lambda_2$ are model hyper-parameters to balance the scale of t-TNN and extra regularization, $\mathcal{X}\in\mathbb{R}^{I\times J\times K}$ or $\mathbf{Z}\in\mathbb{R}^{(IK)\times J}$ are the target variables.

Eq. \eqref{final model} is a transductive semi-supervised learning model, which means that the model training and inference process are in the same stage. 
The motivation of using the tensorization operator $\mathscr{T}$ to connect tensor with matrix is that we can impose regularization on columns and rows of the spatiotemporal matrix directly, and utilize the t-TNN of tensor to achieve 'low-rank' in the graph spectral domain simultaneously, without pre-specified rank parameter as in \citep{kolda2009tensor}. Another important reason is that the optimization problem of typical tensor completion is usually coupled \citep{liu2013tensor}, and we can decouple it using the connection between tensor and matrix with no need for introducing extra auxiliary variables.

\begin{remark}
    It can be seen that Eq. \eqref{final model} contains the 'location' and 'time-of-day' mode Laplacian matrices explicitly, and keeps the 'day' mode Laplacian implicitly in the t-TNN. This consideration is twofold: (1) We treat 'day-to-day' and 'day-of-week' as global features, which are complementary to low-rankness; (2) As TGFT is performed on the 'day' dimension for all locations, we can preserve the periodic structure of traffic data by 'day' mode Laplacian, which is a natural and straightforward choice. Ones can also impose Laplacian regularization on three unfolding matrices in turn, as in \citep{sofuoglu2022gloss}. However, this treatment could bring extra computational complexity and may cause over-smoothing problem.
\end{remark}

\subsection{Scalable solution with ADMM}
\label{admm_solve}
It can be seen that after incorporating the spatial and temporal regularizers the model updating process confronts with more computational overhead, because both the Toeplitz and graph Laplacian are item-by-item smoothing operators and more computations are introduced.
A traditional choice for solving tensor completion problem is the alternating direction method of multipliers (ADMM) method \citep{liu2013tensor}. ADMM is an iterative algorithm very suitable for numerous convex problems and the basic idea is to update each independent variable in a proper order for each epoch until the convergence. 

We first give the basic formulation of ADMM. The corresponding augmented Lagrange function of Eq. \eqref{final model} is defined as:
% \begin{equation}
% \begin{aligned}
% \mathscr{F}_\mu(\mathcal{X},\mathbf{Z},\mathcal{Y})=&
% \Vert \mathcal{X}\Vert_{t*}+\frac{\lambda_1}{2}\operatorname{Tr}(\mathbf{Z}\mathbf{L_s}\mathbf{Z}^\mathsf{T})+\frac{\lambda_2}{2}\Vert\mathbf{E_nZ}\Vert_F^2 \\
% +&\big\langle\mathcal{Y},\mathcal{X}-\mathscr{T}(\mathbf{Z})\big\rangle+\frac{\mu}{2}\Vert\mathcal{X}-\mathscr{T}(\mathbf{Z})\Vert_F^2,
% \label{ALM}
% \end{aligned}
% \end{equation}
\begin{equation}
\begin{aligned}
\mathscr{F}_\mu(\mathcal{X},\mathbf{Z},\mathcal{Y})=&
\Vert \mathcal{X}\Vert_{t*}+\frac{\lambda_1}{2}\Vert\widetilde{\mathbf{L}}_s\mathbf{Z}^{\mathsf{T}}\Vert_F^2+\frac{\lambda_2}{2}\Vert\mathbf{\Phi}\mathbf{L}_{\Phi}\mathbf{Z}\Vert_F^2 \\
+&\big\langle\mathcal{Y},\mathcal{X}-\mathscr{T}(\mathbf{Z})\big\rangle+\frac{\mu}{2}\Vert\mathcal{X}-\mathscr{T}(\mathbf{Z})\Vert_F^2,
\label{ALM}
\end{aligned}
\end{equation}
where $\mathcal{Y}$ is the Lagrange dual variable tensor, $\mu$ is the penalty parameter of equality constraint as well as the learning rate. Note that we do not incorporate the constraint $\mathbf{P}\odot\mathbf{Z}=\mathbf{P}\odot\mathbf{T}$ into $\mathscr{F}_\mu$ as we directly use this equality to transfer observations. Then, the ADMM alternately minimizes each of the target variable and dual variable until convergence. The iterative scheme of ADMM leads to the following updating rules:
\begin{equation}
    \begin{aligned} 
    \left\{\begin{array}{l}
    \mathcal{X}^{j+1}=\operatorname{arg}\min_{\mathcal{X}}\mathscr{F}_\mu(\mathcal{X},\mathbf{Z}^j,\mathcal{Y}^j), \\   
    \mathbf{Z}^{j+1}=\operatorname{arg}\min_{\mathbf{Z}}\mathscr{F}_\mu(\mathcal{X}^{j+1},\mathbf{Z},\mathcal{Y}^j), \\ 
    \mathcal{Y}^{j+1}=\mathcal{Y}^{j}+\mu\left(\mathcal{X}^{j+1}-\mathscr{T}(\mathbf{Z}^{j+1})\right).
    \end{array}\right. \\
    \end{aligned}
\label{updating rule}
\end{equation}
    
For $j$-th iteration, we need to compute the optimal solution to each subproblem so that the ADMM guarantees the ideal solution. However, the overall computational efficiency is determined by the complexity of each subproblem and declines dramatically with the increase of problem scale.

As discussed in section \ref{Introduction}, there exists a wealth of spatial correlations to be mined on the entire road network and integrating network-wide of data could contain more comprehensive relationships. In order to make the proposed model in Eq. \eqref{final model} scalable to large network, we reformulate the ADMM scheme by improving the computation of each subproblem. Technically, we find the bottleneck (the highest time complexity) of each updating rule and accelerate it with numeric approximation methods: refining the solutions to $\mathcal{X}$ and $\mathbf{Z}$ subproblems using random projection and subspace iterative techniques separately.
As can be found below, these useful designs make the model scalable to large-scale problems while keep high accuracy.

\subsubsection{Solving X subproblem with random projection}\label{X_solve}
To update $\mathcal{X}$ variable, we first arrange the first line in Eq. \eqref{updating rule} in the standard form of t-TNN minimization:

\begin{equation}
\begin{aligned} 
\mathcal{X}^{j+1}=&\operatorname{arg}\min_{\mathcal{X}}\mathscr{F}_\mu(\mathcal{X},\mathbf{Z}^j,\mathcal{Y}^j), \\   
=&\operatorname{arg}\min_{\mathcal{X}}\Vert\mathcal{X}\Vert_{t*}+\big\langle\mathcal{Y}^j,\mathcal{X}-\mathscr{T}(\mathbf{Z}^j)\big\rangle+\frac{\mu}{2}\Vert\mathcal{X}-\mathscr{T}(\mathbf{Z}^j)\Vert_F^2, \\
=&\operatorname{arg}\min_{\mathcal{X}}\Vert\mathcal{X}\Vert_{t*}+\frac{\mu}{2}\Vert\mathcal{X}-\left(\mathscr{T}(\mathbf{Z}^j)-\mathcal{Y}^j/\mu\right)\Vert_F^2.
\end{aligned}
\label{xsub}
\end{equation}

Under the definition of t-TNN, a general choice for solving Eq. \eqref{xsub} is to resort to a proximity operator called tensor singular value thresholding (t-SVT). We give the result in Lemma \ref{lemma1} as follows.

\begin{lemma}{(t-SVT \citep{lu2019low})}
\label{lemma1}
For any tensor $\mathcal{M}\in\mathbb{R}^{n_1\times n_2\times n_3}$, let $\mathscr{L}$ be any invertible linear transform and the tensor singular value decomposition (t-SVD) is given by $\mathcal{M}=\mathcal{U}*_{\mathscr{L}}\mathcal{S}*_{\mathscr{L}}\mathcal{V}^\mathsf{T}$, then the optimal solution to the following problem
    \begin{equation*}
    \min_{\mathcal{X}}\Vert\mathcal{X}\Vert_{t*}+\frac{\mu}{2}\Vert\mathcal{X}-\mathcal{M}\Vert_F^2,
    \end{equation*}
is given by the t-SVT:
    $\mathscr{D}_{1/\mu}(\mathcal{M})=\mathcal{U}*_{\mathscr{L}}\mathcal{S}_{1/\mu}*_{\mathscr{L}}\mathcal{V}^{\mathsf{T}}, 
    \mathcal{S}_{1/\mu}=\mathscr{L}^{-1}([\mathscr{L}(\mathcal{S})-\frac{1}{\mu}]_+),$
where $[t]_+=\operatorname{max}(t,0)$.
\end{lemma}

The variables in Lemma \ref{lemma1} are all in the form of tensor and this operator simply performs the soft-thresholding rule to $\mathscr{L}(\mathcal{S})$. Practically, since the t-SVD can be computed by performing matrix SVD in the transformed domain (graph spectral domain in this work), the t-SVT described above can also be calculated in an easy-to-understand fashion based on the definition of t-product (see Def. \ref{tprod}). Algorithm \ref{t-svt} summarizes the procedure to obtain $\mathscr{D}_{1/\mu}(\mathcal{M})$.

\begin{algorithm}[H]
\caption{computation of t-SVT based on t-product $*_{\mathscr{L}}$}\label{t-svt}
\KwIn{$\mathcal{M}\in\mathbb{R}^{n_1\times n_2\times n_3},\text{invertible transform }\mathscr{L},\text{thresholding parameter }1/\mu$.}
\KwOut{$\mathscr{D}_{1/\mu}(\mathcal{M})$.}

$\Bar{\mathcal{M}}\leftarrow\mathscr{L}(\mathcal{M})$\;
compute each frontal slice of $\Bar{\mathcal{U}},\Bar{\mathcal{S}},\Bar{\mathcal{V}}$ from the slices of $\Bar{\mathcal{M}}$ by:\\
\For{$i=1$ \KwTo $n_3$}{$[\Bar{\mathbf{U}}^{(i)},\Bar{\mathbf{S}}^{(i)},\Bar{\mathbf{V}}^{(i)}]\leftarrow\operatorname{svd}(\Bar{\mathbf{M}}^{(i)})$\;
$\Bar{\mathbf{S}}^{(i)}_{1/\mu}\leftarrow[\Bar{\mathbf{S}}^{(i)}-1/\mu]_+$\;
$\Bar{\mathbf{M}}^{(i)}_{1/\mu}\leftarrow\Bar{\mathbf{U}}^{(i)}\Bar{\mathbf{S}}^{(i)}_{1/\mu}\Bar{\mathbf{V}}^{(i)\mathsf{T}}$\;
}
concatenate $\Bar{\mathbf{M}}^{(i)}_{1/\mu}$ along the third dimension to form $\Bar{\mathcal{M}}_{1/\mu}$\;
$\mathscr{D}_{1/\mu}(\mathcal{M})\leftarrow\mathscr{L}^{-1}(\Bar{\mathcal{M}}_{1/\mu})$.
\end{algorithm}

From Lemma \ref{lemma1} one can see that the $\mathcal{X}$ subproblem boils down to solve a series of matrix SVT problems on daily slices along the third axis in the transformed domain and the temporal periodicity are preserved in the TGFT tensor. This is a significant advantage of t-TNN that makes it possible to be extended to large data sets. Moreover, as daily matrices are independent on each other, the computation can be conducted in a highly-parallel way.

Despite the t-TNN itself has reduced the computation cost to some extent, there is still space for improvement. The main bottleneck of Algorithm \ref{t-svt} is to compute the SVD of certain $n_1\times n_2$ matrices and the per-iteration cost is $\mathcal{O}(n_1^2n_2n_3)$ as discussed in section \ref{TGFT_subsec}. However, tensors with sensor numbers $n_2$ much larger than time intervals $n_1$ could suffer from high computational cost at each iteration. In fact, from the calculation procedure of SVT we can observe that exact and complete SVD calculation is unnecessary for thresholding operation, because only a few singular values larger than the threshold value can have an effect. To this end, we expect to reduce the computation burden by finding an approximation of partial SVD.

A desired choice is using the random projection method to accelerate the SVD, i.e., randomized SVD (rSVD) \citep{halko2011finding}. The core idea of rSVD is using a random projection matrix to help identify the subspace that dominates the pattern of matrix. Then, a rank-$k$ approximation of the original matrix can be obtained which further leads to the approximate SVD. In general, a Gaussian i.i.d matrix can be specified as the random projection, and the subspace's orthonormal basis can be computed by the QR decomposition \citep{golub2013matrix}. For more details about rSVD please refer to \citep{halko2011finding}.

In light of this view, we can speed up the matrix SVT calculation by the following lemma, then we adapt it for t-SVT by the property of t-product:

\begin{lemma}\label{svt-approx}
Let $\mathbf{A}=\mathbf{QB}\in\mathbb{R}^{m\times n}$, where $\mathbf{Q}\in\mathbb{R}^{m\times k}$ is partially orthonormal, then we have:
\begin{equation}
    \mathscr{S}_{\tau}(\mathbf{A})=\mathbf{Q}\mathscr{S}_{\tau}(\mathbf{B}),
\end{equation}
where $\mathscr{S}_{\tau}(\cdot)$ denotes the matrix SVT operator, $k$ is the rank of $\mathbf{A}$.
\end{lemma}

Based on Lemma \ref{svt-approx}, we could avoid expensive computational burden by instead performing SVT on a smaller matrix $\mathbf{B}\in\mathbb{R}^{k\times n}$. If the orthonormal matrix $\mathbf{Q}$ is available and suppose $m\geq n$, the complexity for SVD becomes $\mathcal{O}(nk^2)$, which is much cheaper than the original SVD. 

To perform Lemma \ref{svt-approx}, there are two key points to carefully consider. One major challenge is how to get an appropriate $\mathbf{Q}$. Fortunately, the orthonormal basis can be computed from the procedure of rSVD using QR decomposition. Another task is to estimate the target rank $k$ at each iteration of ADMM. An empirical observation is that during the training process of tensor completion, the rank increases over iterations since the missing values are filled gradually. By exploiting this prior, a simple but effective way is to set a small value at first and increase it gradually with iteration until the upper bound. Another insightful consideration of r-tSVT is that setting a small rank $k$ in each epoch encourages the low-rank characteristics of the solution. As the threshold value $1/\mu$ decreases over iterations (to benefit convergence), a lower rank solution could improve the estimation accuracy in practice. 

The whole process to get $\mathscr{D}_{1/\mu}(\mathcal{\widetilde{A}})$ is summarized in Algorithm \ref{r_tsvt}. Since the above lemma is based on matrix SVT, we integrate it with t-product to adapt to tensor situation.
The core idea is that the computation of t-SVT can be parallelized for each slice matrix, thereby facilitating the integration of r-SVD \citep{zhang2018randomized}.

\begin{algorithm}[H]\label{r_tsvt}
\caption{fast randomized t-SVT based on rSVD}
\KwIn{$\mathcal{A}\in\mathbb{R}^{n_1\times n_2\times n_3},\text{rank parameter }k<\min\{n_1,n_2\}, \text{power parameter }p, \text{oversampling parameter }s$, $\text{transform }\mathscr{L}, \text{thresholding parameter }1/\mu$.}
\KwOut{Low rank approximation $\mathscr{D}_{1/\mu}(\mathcal{\widetilde{A}})$ of $\mathscr{D}_{1/\mu}(\mathcal{A})$.}
$\Bar{\mathcal{A}}\leftarrow\mathscr{L}(\mathcal{A})$\;
sample a Gaussian random matrix: $\boldsymbol{\Omega}\leftarrow\operatorname{randn}(n_1,k+s)$\;
\For{$i=1$ \KwTo $n_3$}
{$\Bar{\mathbf{G}}\leftarrow\Bar{\mathbf{A}}^{(i)\mathsf{T}}\boldsymbol{\Omega}$\;
conduct power iteration by:\\
\For{$j=1$ \KwTo $p$}{$\Bar{\mathbf{G}}\leftarrow\Bar{\mathbf{A}}^{(i)\mathsf{T}}\Bar{\mathbf{A}}^{(i)}\Bar{\mathbf{G}}$\;}
compute a reduced QR decomposition: $\Bar{\mathbf{Q}}^{(i)}\leftarrow\operatorname{QR}(\Bar{\mathbf{G}})$\;
$\Bar{\mathbf{B}}^{(i)}\leftarrow\Bar{\mathbf{Q}}^{(i)\mathsf{T}}\Bar{\mathbf{A}}^{(i)\mathsf{T}}$\;
}
concatenate $\Bar{\mathbf{B}}^{(i)}$ and $\Bar{\mathbf{Q}}^{(i)}$ along the third dimension to form $\Bar{\mathcal{B}}$, $\Bar{\mathcal{Q}}$ respectively\;
compute the t-SVT of $\Bar{\mathcal{B}}$ using line 2-7 in Algorithm \ref{t-svt} to obtain $\Bar{\mathcal{B}}_{1/\mu}$\;
$\Bar{\mathcal{A}}_{1/\mu}\leftarrow\Bar{\mathcal{Q}}*_{\mathscr{L}}\Bar{\mathcal{B}}_{1/\mu}$\;
$\mathscr{D}_{1/\mu}(\mathcal{\widetilde{A}})\leftarrow\mathscr{L}^{-1}(\Bar{\mathcal{A}}_{1/\mu})$.
\end{algorithm}

There are some expositions about Algorithm \ref{r_tsvt}. In our case $n_1 (I)<n_2 (J)$, we first transpose each slice matrix $\Bar{\mathbf{A}}^{(i)}$ from $n_1\times n_2$ to $n_2\times n_1$ such that the resulted small matrix $\Bar{\mathbf{B}}^{(i)}\in\mathbb{R}^{k\times n_1}$ requires minimal computation for SVD. Moreover, a useful technique called power iteration is applied in line 5-7 in Algorithm \ref{r_tsvt} to improve the estimation accuracy of $\mathbf{Q}$ and \cite{halko2011finding} also found that in practice $p=1$ or $2$ usually suffices.

Finally, by adopting Algorithm \ref{r_tsvt}, the optimal solution to update $\mathcal{X}$ in Eq. \eqref{xsub} is:
\begin{equation}
\mathcal{X}^{j+1} = \mathscr{D}_{1/\mu}\left(\mathscr{T}(\mathbf{Z}^j)-\mathcal{Y}^j/\mu\right).
\label{x-update}    
\end{equation}

\subsubsection{Solving Z subproblem with conjugate gradient} 
The second term in Eq. \eqref{updating rule} is the optimization of spatial proximity and temporal continuity regularization that closely related to kriging tasks. After introducing the (inverse) tensorization operator $\mathbf{Z}=\mathscr{T}^{-1}(\mathcal{X})$, we can update $\mathbf{Z}$ in the form of matrix easily:

% \begin{equation}
% \begin{aligned}    \mathbf{Z}^{j+1}&=\operatorname{arg}\min_{\mathbf{Z}}\mathscr{F}_\mu(\mathcal{X}^{j+1},\mathbf{Z},\mathcal{Y}^j), \\
%     &=\operatorname{arg}\min_{\mathbf{Z}}\frac{\lambda_1}{2}\operatorname{Tr}(\mathbf{Z}\mathbf{L_s}\mathbf{Z}^\mathsf{T})+\frac{\lambda_2}{2}\Vert\mathbf{E_nZ}\Vert_F^2+\big\langle\mathcal{Y}^{j},\mathcal{X}^{j+1}-\mathscr{T}(\mathbf{Z})\big\rangle+\frac{\mu}{2}\Vert\mathcal{X}^{j+1}-\mathscr{T}(\mathbf{Z})\Vert_F^2, \\
%     &=\operatorname{arg}\min_{\mathbf{Z}}
%     \underbrace{\frac{\lambda_1}{2}\operatorname{Tr}(\mathbf{Z}\mathbf{L_s}\mathbf{Z}^\mathsf{T})+\frac{\lambda_2}{2}\Vert\mathbf{E_nZ}\Vert_F^2
%     +\frac{\mu}{2}\Vert\mathbf{Z}-\mathscr{T}^{-1}(\mathcal{X}^{j+1}+\mathcal{Y}^{j}/\mu)\Vert_F^2}_{\mathscr{H}(\mathbf{Z})}. \\
% \end{aligned}
% \end{equation}
\begin{equation}
\begin{aligned}    \mathbf{Z}^{j+1}&=\operatorname{arg}\min_{\mathbf{Z}}\mathscr{F}_\mu(\mathcal{X}^{j+1},\mathbf{Z},\mathcal{Y}^j), \\
    &=\operatorname{arg}\min_{\mathbf{Z}}\frac{\lambda_1}{2}\Vert\widetilde{\mathbf{L}}_s\mathbf{Z}^{\mathsf{T}}\Vert_F^2+\frac{\lambda_2}{2}\Vert\mathbf{\Phi}\mathbf{L}_{\Phi}\mathbf{Z}\Vert_F^2+\big\langle\mathcal{Y}^{j},\mathcal{X}^{j+1}-\mathscr{T}(\mathbf{Z})\big\rangle+\frac{\mu}{2}\Vert\mathcal{X}^{j+1}-\mathscr{T}(\mathbf{Z})\Vert_F^2, \\
    &=\operatorname{arg}\min_{\mathbf{Z}}
    \underbrace{\frac{\lambda_1}{2}\Vert\widetilde{\mathbf{L}}_s\mathbf{Z}^{\mathsf{T}}\Vert_F^2+\frac{\lambda_2}{2}\Vert\mathbf{\Phi}\mathbf{L}_{\Phi}\mathbf{Z}\Vert_F^2
    +\frac{\mu}{2}\Vert\mathbf{Z}-\mathscr{T}^{-1}(\mathcal{X}^{j+1}+\mathcal{Y}^{j}/\mu)\Vert_F^2}_{\mathscr{H}(\mathbf{Z})}. \\
\end{aligned}
\end{equation}

Set the gradient of $\mathscr{H}(\mathbf{Z})$ to zero, we can have the least square solution:
% \begin{equation}
% \lambda_1\mathbf{Z}\mathbf{L_s}+\lambda_2\mathbf{E_n}^{\mathsf{T}}\mathbf{E_nZ}+\mu\mathbf{Z}=\mu\mathscr{T}^{-1}(\mathcal{X}^{j+1}+\mathcal{Y}^{j}/\mu).
% \label{sylvester}
% \end{equation}
\begin{equation}
\lambda_1\mathbf{Z}\widetilde{\mathbf{L}}_s^{\mathsf{T}}\widetilde{\mathbf{L}}_s+\lambda_2(\mathbf{\Phi}\mathbf{L}_{\Phi})^{\mathsf{T}}\mathbf{\Phi}\mathbf{L}_{\Phi}\mathbf{Z}+\mu\mathbf{Z}=\mu\mathscr{T}^{-1}(\mathcal{X}^{j+1}+\mathcal{Y}^{j}/\mu).
\label{sylvester}
\end{equation}

Eq. \eqref{sylvester} is a standard Sylvester matrix equation in the form of $\mathbf{AX+XB=C}$. Typically, a direct manner to solve this system is vectorizing it into a standard linear equations, as shown in Eq. \eqref{vectorization}. 
% \begin{equation}
% (\lambda_1\mathbf{L_s}\otimes\mathbf{I}_{IK}+\lambda_2\mathbf{I}_{J}\otimes\mathbf{E_n}^{\mathsf{T}}\mathbf{E_n}+\mu\mathbf{I}_{IJK})\operatorname{vec}(\mathbf{Z})=\mu\operatorname{vec}(\mathscr{T}^{-1}(\mathcal{X}^{j+1}+\mathcal{Y}^{j}/\mu)),
% \label{vectorization}
% \end{equation}
\begin{equation}
(\lambda_1\widetilde{\mathbf{L}}_s^{\mathsf{T}}\widetilde{\mathbf{L}}_s\otimes\mathbf{I}_{IK}+\lambda_2\mathbf{I}_{J}\otimes\mathbf{L}_{\Phi}^{\mathsf{T}}\mathbf{\Phi}^{\mathsf{T}}\mathbf{\Phi}\mathbf{L}_{\Phi}+\mu\mathbf{I}_{IJK})\operatorname{vec}(\mathbf{Z})=\mu\operatorname{vec}(\mathscr{T}^{-1}(\mathcal{X}^{j+1}+\mathcal{Y}^{j}/\mu)),
\label{vectorization}
\end{equation}
where $\otimes$ is the Kronecker product, $\operatorname{vec}(\cdot)$ denotes the vectorization operator, and $\mathbf{I}_n$ denotes an identity matrix of size $n\times n$. This conversion holds due to the fact that $\operatorname{vec}(\mathbf{AXB})=(\mathbf{B}^{\mathsf{T}}\otimes\mathbf{A})\operatorname{vec}(\mathbf{X})$.

However, the computational complexity of this vectorized solution is $\mathcal{O}(I^3J^3K^3)$, which is both computationally expensive and memory-consuming for large-scale problems. An alternative solution for Sylvester system is using Bartels-Stewart method \citep{golub2013matrix}. Although the cost can be reduced to $\mathcal{O}(I^3K^3+J^3)$, it is still prohibitive in practice. Note that in our case $\mathbf{A}$ and $\mathbf{B}$ are highly structured, i.e., symmetric and positive definite, so that the matrix vector multiplies involving them can be efficient. Taking this into account, subspace iterative methods could potentially perform better for fixing this problem. 

Conjugate gradient (CG) is a representative Krylov subspace iteration method and has been verified to be an efficient and stable manner in recent works \citep{kalofolias2014matrix,rao2015collaborative}. One distinct advantage of CG is that it is nonparametric and requires only a small number of iterations to solve the structured linear system with desired accuracy. By using the vectorization operation in Eq. \eqref{vectorization}, we apply CG to such a symmetric positive definite system to approximate its numeric solution. Details about CG is omitted for simplicity and we present the procedure for solving $\mathbf{Z}$ variable in Algorithm \ref{CG}.

\begin{algorithm}[H]\label{CG}
\caption{Conjugate gradient for Z-subproblem}
\KwIn{$\mathbf{Z}\in\mathbb{R}^{(IK)\times J}, \mathcal{X,Y},\mu, \mathbf{L_s,E_n},\lambda_1,\lambda_2,\text{maximum iterations }t$.}
\KwOut{$\text{Estimated variable }\hat{\mathbf{Z}}$.}
$\mathscr{V}(\mathbf{Z})\leftarrow(\lambda_1\widetilde{\mathbf{L}}_s^{\mathsf{T}}\widetilde{\mathbf{L}}_s\otimes\mathbf{I}_{IK}+\lambda_2\mathbf{I}_{J}\otimes\mathbf{L}_{\Phi}^{\mathsf{T}}\mathbf{\Phi}^{\mathsf{T}}\mathbf{\Phi}\mathbf{L}_{\Phi}+\mu\mathbf{I}_{IJK})\operatorname{vec}(\mathbf{Z})$\;
$\mathbf{r}_0\leftarrow\mu\operatorname{vec}(\mathscr{T}^{-1}(\mathcal{X}^{j+1}+\mathcal{Y}^{j}/\mu))-\mathscr{V}(\mathbf{Z})$\;
$\mathbf{q}_0\leftarrow\mathbf{r}_0$\;
\For{$i=1$ \KwTo $t$}
{$\mathbf{Q}_i\leftarrow\operatorname{reshape}(\mathbf{q}_i)$\;
$\alpha_i\leftarrow\mathbf{r}_i^{\mathsf{T}}\mathbf{r}_i/\left(\mathbf{q}_i^{\mathsf{T}}\mathscr{V}(\mathbf{Q}_i)\right)$\;
$\mathbf{z}_{i+1}\leftarrow \mathbf{z}_i+\alpha_i\mathbf{q}_i$\;
$\mathbf{r}_{i+1}\leftarrow \mathbf{r}_i-\alpha_i\mathscr{V}(\mathbf{Q}_i)$\;
$\beta_i\leftarrow\frac{\mathbf{r}_{i+1}^{\mathsf{T}}\mathbf{r}_{i+1}}{\mathbf{r}_i^{\mathsf{T}}\mathbf{r}_i}$\;
$\mathbf{q}_{i+1}\leftarrow \mathbf{r}_{i+1}+\beta_i\mathbf{q}_i$\;
}
$\hat{\mathbf{Z}}\leftarrow\operatorname{reshape}(\mathbf{z}_{t})$.
\end{algorithm}

In Algorithm \ref{CG}, each iteration only involves one single matrix-vector product, several inner products and scalar operations. This makes it highly scalable to large problems and we can incorporate it in the ADMM framework.

The main cost of Algorithm \ref{CG} is the matrix-vector product to calculate $\mathscr{V}(\mathbf{Q}_i)$ and the time complexity is $\mathcal{O}(I^2J^2K^2)$ for each inner CG iteration. What is noteworthy is that this cost can be further reduced if we first compute the matrix-matrix products in $\lambda_1\mathbf{Q}_i\widetilde{\mathbf{L}}_s^{\mathsf{T}}\widetilde{\mathbf{L}}_s+(\lambda_2\mathbf{L}_{\Phi}^{\mathsf{T}}\mathbf{\Phi}^{\mathsf{T}}\mathbf{\Phi}\mathbf{L}_{\Phi}+\mu\mathbf{I})\mathbf{Q}_i$
and then obtain $\mathscr{V}(\mathbf{Q}_i)$ using $\operatorname{reshape}(\cdot)$ directly. And the complexity is essentially $\mathcal{O}\left(IJK(IK+J)\right)$ in this case, which is substantially reduced compared to $\mathcal{O}(I^3J^3K^3)$ required by the vectorized solution.

\subsubsection{Solution algorithm}
Having solved each ADMM subproblem, the proposed LETC algorithm is finally summarized in Algorithm \ref{SGET}. In this implementation, the updating order follows $\{\mathcal{X}^j\Rightarrow\mathbf{Z}^{j}\Rightarrow\mathcal{Y}^j\}$ and the estimated data values can be returned either in matrix $\hat{\mathbf{Z}}$ or in tensor $\hat{\mathcal{X}}$. In line 8, we increase the penalty parameter as well as the learning rate $\mu$ to promote the convergence of ADMM, and also increase the target rank $k$ for Algorithm \ref{r_tsvt}. As we solve both of the two subproblems in a highly scalable routine, this algorithm is efficient for large-scale kriging tasks. Moreover, two most important hyper-parameters in Algorithm \ref{SGET} are the regularization parameters $\lambda_1,\lambda_2$. These coefficients function as a balance between t-TNN minimization and local consistency penalty, and we will discuss the impacts of them in Section \ref{hpt}.

\begin{algorithm}[!htb]
% \caption{Spatiotemporal graph embedded low-rank tensor learning for kriging}\label{SGET}
\caption{Spatiotemporal Laplacian enhanced low-rank tensor completion for kriging}\label{SGET}
\KwIn{Measured speed matrix $\mathbf{T}$, indicating matrix $\mathbf{P}$, linear transform $\mathscr{L}$, learning rate $\mu$, initial rank parameter $k$, regularization terms and coefficients $\mathbf{L_s,E_n},\lambda_1,\lambda_2$, convergence condition $\epsilon$.}
\KwOut{Estimated full matrix $\hat{\mathbf{Z}}$.}
Initialize $\mathcal{Y}^0$ as zeros, and set $\mathbf{P}\odot\mathbf{Z}^0\leftarrow\mathbf{P}\odot\mathbf{T}$, $j\leftarrow0$;  \\
\While{not converged}{
Update $\mathcal{X}^{j+1}$ by Eq. \eqref{x-update} and Algorithm \ref{t-svt}, \ref{r_tsvt}; \\
Update $\mathbf{Z}^{j+1}$ by Algorithm \ref{CG}; \\
Transmit the observations by setting 
$\mathbf{P}\odot\mathbf{Z}^{j+1}\leftarrow\mathbf{P}\odot\mathbf{T}$;\\
Update $\mathcal{Y}^{j+1}$ by $\mathcal{Y}^{j+1}\leftarrow\mathcal{Y}^{j}+\mu\left(\mathcal{X}^{j+1}-\mathscr{T}(\mathbf{Z}^{j+1})\right)$;\\
$e^{j+1}\leftarrow\Vert\mathbf{Z}^{j+1}-\mathbf{Z}^{j}\Vert_F/\Vert\mathbf{Z}^{j}\Vert_F$;\\
Increase the learning rate $\mu$ and rank parameter $k$ with a fixed step; \\
\If{$e^{j+1}<\epsilon$}{Converge.}
$j\leftarrow j+1$;
}
\end{algorithm}

\subsection{Overall computational complexity analysis}
\label{complexity analysis}
The computational complexity of each solving step is introduced separately and we give a brief summary and discussion here. Overall, the main cost of Algorithm \ref{SGET} is t-SVT in $\mathcal{X}$-subproblem and CG in $\mathbf{Z}$-subproblem. The step-by-step computational complexity of Algorithm \ref{r_tsvt} and \ref{CG} is summarized in Tab. \ref{complexity analysis 1} and Tab. \ref{complexity analysis 2}, respectively. Since lines 4-11 in Tab. \ref{complexity analysis 1} are conducted for each daily slice, their complexity become $K$ times. As we can see, the main bottleneck of LETC is the matrix-vector product in CG iteration. But this implementation is efficient enough for large-scale applications.

\begin{table}[!htb]
\begin{minipage}{0.48\linewidth}
  \centering
  \caption{Time complexity analysis for Algorithm \ref{r_tsvt}.}
  \footnotesize
    \begin{tabular}{c|cll}
    \toprule
    Line No. & Specific operation & \multicolumn{2}{c}{Complexity} \\
    \midrule
    4 & $\Bar{\mathbf{G}}\leftarrow\Bar{\mathbf{A}}^{(i)\mathsf{T}}\boldsymbol{\Omega}$  & $\mathcal{O}(kIJ)$ & \rdelim\}{5}{5mm}[$\times K$] \\
    7 & $\Bar{\mathbf{G}}\leftarrow\Bar{\mathbf{A}}^{(i)\mathsf{T}}\Bar{\mathbf{A}}^{(i)}\Bar{\mathbf{G}}$  & $\mathcal{O}(kIJ)$   & \\
    8& $\Bar{\mathbf{Q}}^{(i)}\leftarrow\operatorname{QR}(\Bar{\mathbf{G}})$   & $\mathcal{O}(Jk^2)$ & \\
    9& $\Bar{\mathbf{B}}^{(i)}\leftarrow\Bar{\mathbf{Q}}^{(i)\mathsf{T}}\Bar{\mathbf{A}}^{(i)\mathsf{T}}$   & $\mathcal{O}(kIJ)$ & \\
    11& $\operatorname{svd}(\Bar{\mathbf{B}}^{(i)})$   & $\mathcal{O}(Ik^2)$ & \\
    12& $\Bar{\mathcal{A}}_{1/\mu}\leftarrow\Bar{\mathcal{Q}}*_{\mathscr{L}}\Bar{\mathcal{B}}_{1/\mu}$   & $\mathcal{O}(kIJK)$ & \\
    \bottomrule
    \end{tabular}
    \label{complexity analysis 1}
    \end{minipage}
    %\hfill
    \begin{minipage}{0.48\linewidth}  
    \centering
  \caption{Time complexity analysis for Algorithm \ref{CG}.}
  \footnotesize
    \begin{tabular}{c|cc}
    \toprule
    Line No. & Specific operation & Complexity \\
    \midrule
    1 & Compute $\mathscr{V}(\mathbf{Z})$  & $\mathcal{O}\left(IJK(J+IK)\right)$ \\
    2 & Compute $\mathbf{r}_0$  & $\mathcal{O}(IJK)$    \\
    6& $\alpha_i\leftarrow\mathbf{r}_i^{\mathsf{T}}\mathbf{r}_i/\left(\mathbf{q}_i^{\mathsf{T}}\mathscr{V}(\mathbf{Q}_i)\right)$   & $\mathcal{O}\left(IJK(J+IK)\right)$  \\
    7,8,10& Compute $\mathbf{z}_{i+1}$, $\mathbf{r}_{i+1}$, $\mathbf{q}_{i+1}$   & $\mathcal{O}(IJK)$ \\
    11& $\beta_i\leftarrow\mathbf{r}_{i+1}^{\mathsf{T}}\mathbf{r}_{i+1}/(\mathbf{r}_i^{\mathsf{T}}\mathbf{r}_i)$   & $\mathcal{O}(IJK)$  \\
    \bottomrule
    \end{tabular}
    \label{complexity analysis 2}
    \end{minipage}
\end{table}

\section{Case study}\label{experiments}

In this section, we conduct experiments to evaluate the kriging performance of the proposed LETC model on two real-world and network-wide traffic speed datasets under different settings. First we briefly describe the datasets and introduce some state-of-the-art kriging baseline models. Apart from model comparison, we also conduct various ablation studies and sensitivity studies for algorithmic analysis. All the experiments are carried out on a Windows 10 platform with Intel(R) Core(TM) i7-12700KF 3.60GHz CPU (12 cores in total) and 32 GB RAM. \textbf{The NumPy implementation is shared in our GitHub repository:} \url{https://github.com/tongnie/tensor4kriging}.

\subsection{Data preparation and experiment setup}
\subsubsection{Spatiotemporal traffic speed datasets}
As we aim to evaluate kriging performances on large data, two publicly available large-scale traffic speed datasets: PeMS-4W freeway speed data \citep{chen2021scalable} and Portland highway speed data are used for benckmark experiments. These datasets are briefly summarized as follows:
\begin{itemize}
\item \textbf{PeMS-4W freeway loop speed data\footnote{from \url{https://doi.org/10.5281/zenodo.3939793}.}}: Large-scale traffic speed data collected from 11160 loop sensors installed on the freeway in California. The speed records include the first 4 weeks in 2018 with a 5-min resolution and it can be organized in a spatiotemporal matrix of shape $(8064\times11160)$ or a tensor of shape $(288\times11160\times28)$. This data set also provides a travel distance based adjacent matrix. This data contains about 90 million observations.

\item \textbf{Portland highway loop speed data\footnote{from \url{https://portal.its.pdx.edu/home}.}}: The Portland database provides 20-second granularity loop detector records from freeways in the Portland-Vancouver metropolitan region. We collect traffic speed data from 1057 loop sensors with a 15-minute resolution over 3 months (from January 1 to March 31, 2021). The data size is $(8640\times1057)$ in matrix or $(96\times1057\times90)$ in tensor. This data provides the location information of all sensors. This data contains about 10 million observations.
\end{itemize}

The input data is arranged in a third-order tensor of $(\text{time of day}\times\text{locations}\times\text{days})$ or a matrix of $(\text{time points}\times\text{locations})$. One can find that the 'location' dimension is much larger than the 'time of day' dimension for both of the two datasets. Different from other large datasets used for traffic prediction studies, which usually feature a large 'days' dimension, the two data with pretty large 'locations' dimension are more suitable for testing the extendability of kriging models.

\subsubsection{Competing models and experiment settings}
In order to demonstrate the superiority of the proposed model, we compare LETC with several state-of-the-art baseline methods. These baselines not only contain typical tensor/matrix based spatiotemporal kriging models, but also cover advanced methods from machine learning communities. 

\begin{itemize}
\item \textbf{TGMF \citep{zhang2020network}}: Temporal geometric matrix factorization. This model uses graph Laplacian and one-order Toeplitz matrix as spatial and temporal regularization. ADMM is used to update variables.
\item \textbf{LSTM-GRMF \citep{yang2021real}}: This model extends the graph Laplacian regularized matrix factorization framework with LSTM temporal regularization. It can perform online imputation and prediction, while we only test the performance of static training and kriging phase.
\item \textbf{FMDT-Tucker \citep{yamamoto2022fast}}: Fast algorithm for multi-way delay embedded (MDT, aka., Hankelization) Tucker decomposition. This work is an enhanced version of \cite{yokota2018missing} where the MDT is accelerated by FFT. It utilizes the shift-invariant features of rows/columns to perform image-inpainting which can also achieve spatiotemporal kriging.
\item \textbf{GLTL \citep{bahadori2014fast}}: Greedy low-rank tensor learning. This is a classical tensor learning method for spatiotemporal kriging and is solved by greedy algorithm.
% \item \textbf{GRALS \citep{rao2015collaborative}}: Graph regularized alternating least squares method. This is a well scalable graph-based matrix factorization model. Both of the factor matrices are solved by conjugate gradient.
\item \textbf{KPMF \citep{zhou2012kernelized}}: Kernelized probabilistic matrix factorization. Row-wise and column-wise kernel(covariance) functions are used as graph side information and a Gaussian process prior is imposed on the factor matrices. This model is solved by stochastic gradient descend method.
\item \textbf{GLOSS \citep{sofuoglu2022gloss}}: Graph regularized low-rank plus temporally
smooth sparse decomposition model, which uses the unfolding-based nuclear norm \citep{liu2013tensor} to capture low-rankness. One-order Toeplitz and symmetric graph Laplacian are also adopted for local smoothness.
\item \textbf{WDG-CP \citep{li2020tensor}}: Weakly-dependent graph regularized CANDECOMP/PARAFAC (CP) tensor decomposition. A $L_1$ norm penalty is adopted to encourage sparsity of temporal factor.
\item \textbf{LETC-separate}: We also examine a variation of our LETC model that handles the global (low-rank) and local (spatially and temporally regularized) parts separately. In this setting, we first conduct imputation for missing data points by keeping the missing rows/columns excluded , using only the first term in Eq. \eqref{final model}. Then we conduct kriging on the imputed matrix using only the last two terms.
\end{itemize}

In fact, although FMDT-Tucker, GRALS and KPMF are not designed for spatiotemporal kriging in their original works (image inpainting and collaborative filtering), the intrinsic mathematical problems are the same as ours such that they can be transferred to this task. Moreover, BKMF \citep{lei2022bayesian} and GNNs \citep{wu2021inductive,liang2022spatial} are also emerging kriging methods, but we do not include them in the baselines because the Bayesian hyper-parameter sampling procedure in former is time-consuming and the training stage in latter also brings heavy computational burden and memory consumption. Both of them are not well-suited for large-scale problems.

To evaluate the models' performance, a series of observation conditions are tested: we randomly mask the ground truth values of $30\%,50\%,70\%$ locations as unmeasured locations (without sensors), and select $20\%,50\%$ time intervals as unmeasured time (non-operational time). So there are six test scenarios in total, and we name them SM3TM2, SM5TM2, SM7TM2, SM3TM5, SM5TM5, SM7TM5, respectively. Furthermore, we also add $20\%$ randomly element-wise missing data to simulate the reality.
The performances are evaluated by comparing the estimated values with masked ones. Two evaluation metrics are used in this section: mean absolute error (MAE) and root mean square error (RMSE):
\begin{equation}
    \begin{aligned}
    &\text{MAE}= \frac{1}{|\Omega_m|}\sum_{i\in\Omega_m} |x_i-\hat{x}_i|,\\
    &\text{RMSE}= \sqrt{\frac{1}{|\Omega_m|}\sum_{i\in\Omega_m} \left(x_i-\hat{x}_i\right)^2},\\
    % &\text{MAPE}= \frac{1}{|\Omega_m|}\sum_{i\in\Omega_m} \vert\frac{x_i-\hat{x}_i}{x_i}\vert,\\
    \end{aligned}
\end{equation}
where $\Omega_m=\{(i,j,k)|~\text{when} ~x_{ijk} ~\text{is manually masked and observed data in original tensor}\}$. Detailed hyper-parameter settings for LETC and other baselines are given in \nameref{Appendix A.}.

\subsection{Network-wide kriging results analysis and model comparison}
The comparison of kriging performances of LETC and its competing baselines are given in Tab. \ref{kriging_results}. All experiments are repeated 5 times with 5 random seeds to generate missing nodes, and the mean values as well as standard deviations are reported. We also report the CPU running time of all models on PeMS data in Tab. \ref{pems_time}, to show the computation cost. To focus on the basic performance of LETC and reduce the number of hyper-parameters to be discussed, we only display the results of setting $ h(\widetilde{\mathbf{A}}_s) = \widetilde{\mathbf{A}}_s$ in Eq. \eqref{spatial regularization} and $\omega_1=\omega_2=\omega_T=1$ in 
 Eq. \eqref{A_t}. More superior performances of LETC can be achieved by carefully selecting more expressive diffusion kernels in section \ref{SP_subsec} as well as temporal weights in section \ref{TGFT_subsec} and tuning these hyper-parameters.
% More referable results on different diffusion kernels are given in \nameref{Appendix B.}.

\begin{table}[!htb]
 \setlength\tabcolsep{1pt}
  \centering
  \caption{Kriging evaluation results on PeMS-4W and Portland data}
  \scriptsize
    \begin{tabular}{c|llllll}
    \toprule
          & \multicolumn{6}{c}{PeMS-4W (MAE/RMSE)} \\
    \midrule
    Models & SM0.3,TM0.2 & SM0.5,TM0.2 & SM0.7,TM0.2 & SM0.3,TM0.5 & SM0.5,TM0.5 & SM0.7,TM0.5 \\
    \midrule
    TGMF  & 4.48$_{\pm{0.02}}$/7.46$_{\pm{0.04}}$ & 4.59$_{\pm{0.01}}$/7.60$_{\pm{0.04}}$ & 4.75$_{\pm{0.00}}$/7.90$_{\pm{0.03}}$ & 5.15$_{\pm{0.01}}$/8.44$_{\pm{0.03}}$ & 6.16$_{\pm{0.03}}$/9.17$_{\pm{0.05}}$ & 6.35$_{\pm{0.01}}$/9.55$_{\pm{0.03}}$ \\
    LSTM-GRMF & 2.73$_{\pm{0.05}}$/5.27$_{\pm{0.06}}$ & 3.25$_{\pm{0.03}}$/6.07$_{\pm{0.06}}$ & 3.68$_{\pm{0.04}}$/6.89$_{\pm{0.06}}$ & 2.76$_{\pm{0.02}}$/5.29$_{\pm{0.03}}$ & 3.39$_{\pm{0.02}}$/6.08$_{\pm{0.04}}$ & 3.84$_{\pm{0.02}}$/6.92$_{\pm{0.04}}$ \\
    FMDT-Tucker & 3.07$_{\pm{0.08}}$/5.58$_{\pm{0.12}}$  & 3.29$_{\pm{0.06}}$/5.92$_{\pm{0.08}}$  & 3.84$_{\pm{0.02}}$/6.97$_{\pm{0.08}}$  & 3.49$_{\pm{0.05}}$/5.98$_{\pm{0.10}}$  & 3.61$_{\pm{0.01}}$/6.50$_{\pm{0.05}}$  & 3.95$_{\pm{0.05}}$/7.37$_{\pm{0.08}}$  \\
    GLTL  & 5.98$_{\pm{0.28}}$/9.12$_{\pm{0.41}}$ & 6.10$_{\pm{0.21}}$/9.33$_{\pm{0.33}}$ & 6.40$_{\pm{0.15}}$/9.86$_{\pm{0.26}}$ & 6.06$_{\pm{0.11}}$/9.44$_{\pm{0.18}}$ & 6.28$_{\pm{0.19}}$ /9.80$_{\pm{0.40}}$ & 6.33$_{\pm{0.15}}$/9.98$_{\pm{0.31}}$  \\
    KPMF  & 4.43$_{\pm{0.06}}$ /7.66$_{\pm{0.08}}$  & 5.08$_{\pm{0.04}}$/8.40$_{\pm{0.10}}$  &  6.45$_{\pm{0.11}}$/9.61$_{\pm{0.19}}$ & 5.19$_{\pm{0.08}}$ /8.44$_{\pm{0.26}}$  & 6.57$_{\pm{0.10}}$/9.34$_{\pm{0.42}}$  &  6.88$_{\pm{0.12}}$/9.74$_{\pm{0.36}}$ \\
    WDG-CP  & 3.10$_{\pm{0.00}}$/5.66$_{\pm{0.03}}$ &	3.54$_{\pm{0.01}}$/6.31$_{\pm{0.02}}$ &	4.35$_{\pm{0.00}}$/7.32$_{\pm{0.02}}$ & 3.27$_{\pm{0.00}}$/5.89$_{\pm{0.02}}$	& 3.99$_{\pm{0.01}}$/6.88$_{\pm{0.04}}$ &	4.64$_{\pm{0.02}}$/7.66$_{\pm{0.02}}$	\\
    GLOSS  & 3.92$_{\pm{0.02}}$/7.00$_{\pm{0.04}}$&	4.35$_{\pm{0.03}}$/7.95$_{\pm{0.02}}$&	5.16$_{\pm{0.03}}$/9.02$_{\pm{0.02}}$&	4.93$_{\pm{0.00}}$/8.57$_{\pm{0.03}}$&	5.66$_{\pm{0.02}}$/8.75$_{\pm{0.05}}$&	6.91$_{\pm{0.04}}$/9.62$_{\pm{0.10}}$ \\
    \rowcolor{gray!15}LETC-separate  & 2.72$_{\pm{0.00}}$/5.02$_{\pm{0.03}}$&	3.16$_{\pm{0.02}}$/5.69$_{\pm{0.03}}$&3.76$_{\pm{0.00}}$/6.65$_{\pm{0.03}}$	&	3.07$_{\pm{0.01}}$/5.60$_{\pm{0.04}}$	&3.44$_{\pm{0.03}}$/6.12$_{\pm{0.03}}$	&3.79$_{\pm{0.02}}$/6.71$_{\pm{0.04}}$ \\
    % \midrule
    \rowcolor{gray!15}LETC($\tau=1$)  & 2.48$_{\pm{0.01}}$/\textbf{4.82$_{\pm{0.02}}$}  & 3.07$_{\pm{0.01}}$/5.68$_{\pm{0.04}}$  &  \textbf{3.62$_{\pm{0.02}}$}/\textbf{6.46$_{\pm{0.04}}$} & \textbf{2.58$_{\pm{0.01}}$}/\textbf{5.06$_{\pm{0.03}}$}  & \textbf{3.16$_{\pm{0.00}}$}/\textbf{5.79$_{\pm{0.03}}$}  & 3.64$_{\pm{0.00}}$/6.49$_{\pm{0.02}}$  \\
    \rowcolor{gray!15}LETC($\tau=2$)  & \textbf{2.45$_{\pm{0.01}}$}/4.84$_{\pm{0.03}}$	 & \textbf{3.06$_{\pm{0.01}}$}/\textbf{5.64$_{\pm{0.02}}$} &	3.63$_{\pm{0.01}}$/6.47$_{\pm{0.02}}$ &	2.65$_{\pm{0.01}}$/5.19$_{\pm{0.04}}$ &	3.18$_{\pm{0.00}}$/5.85$_{\pm{0.04}}$	 &\textbf{3.63$_{\pm{0.03}}$}/\textbf{6.48$_{\pm{0.03}}$} \\
    \rowcolor{gray!15}LETC($\tau=3$)  & 2.53$_{\pm{0.03}}$/5.07$_{\pm{0.05}}$	& 3.13$_{\pm{0.02}}$/5.83$_{\pm{0.03}}$ & 3.64$_{\pm{0.01}}$/6.55$_{\pm{0.02}}$ & 2.80$_{\pm{0.01}}$/5.51$_{\pm{0.02}}$ & 3.30$_{\pm{0.01}}$/6.11$_{\pm{0.02}}$	& 3.69$_{\pm{0.01}}$/6.62$_{\pm{0.03}}$\\
    \midrule
     & \multicolumn{6}{c}{Portland (MAE/RMSE)} \\
    \midrule
    Models & SM0.3,TM0.2 & SM0.5,TM0.2 & SM0.7,TM0.2 & SM0.3,TM0.5 & SM0.5,TM0.5 & SM0.7,TM0.5 \\
    \midrule
    TGMF  & 5.56$_{\pm{0.04}}$/8.52$_{\pm{0.07}}$ & 6.21$_{\pm{0.03}}$/8.90$_{\pm{0.07}}$ & 7.23$_{\pm{0.02}}$/9.53$_{\pm{0.05}}$ & 5.79$_{\pm{0.02}}$/8.76$_{\pm{0.03}}$ & 6.38$_{\pm{0.01}}$/9.10$_{\pm{0.03}}$ & 7.48$_{\pm{0.02}}$/11.42$_{\pm{0.04}}$ \\
    LSTM-GRMF & 3.95$_{\pm{0.03}}$/6.83$_{\pm{0.06}}$ & 4.55$_{\pm{0.02}}$/7.39$_{\pm{0.04}}$ & 5.57$_{\pm{0.01}}$/8.61$_{\pm{0.03}}$ & 4.10$_{\pm{0.02}}$/7.08$_{\pm{0.05}}$ & 4.74$_{\pm{0.04}}$/7.44$_{\pm{0.06}}$ & 5.50$_{\pm{0.02}}$/8.46$_{\pm{0.03}}$\\
    FMDT-Tucker & 6.54$_{\pm{0.01}}$/9.79$_{\pm{0.01}}$  & 6.65$_{\pm{0.05}}$/9.72$_{\pm{0.09}}$  & 6.84$_{\pm{0.05}}$/10.06$_{\pm{0.09}}$  & 6.59$_{\pm{0.04}}$/9.83$_{\pm{0.10}}$  & 6.66$_{\pm{0.03}}$/9.65$_{\pm{0.07}}$  & 6.85$_{\pm{0.03}}$/10.11$_{\pm{0.05}}$  \\
    GLTL  & 7.57$_{\pm{0.21}}$/10.22$_{\pm{0.31}}$  & 7.77$_{\pm{0.24}}$/10.85$_{\pm{0.50}}$  & 9.32$_{\pm{0.33}}$/12.55$_{\pm{0.48}}$  & 8.40$_{\pm{0.22}}$/11.16$_{\pm{0.41}}$  & 9.83$_{\pm{0.19}}$/13.15$_{\pm{0.37}}$  & 10.89$_{\pm{0.28}}$/14.99$_{\pm{0.46}}$  \\
    KPMF  & 5.78$_{\pm{0.04}}$/9.06$_{\pm{0.08}}$  & 6.12$_{\pm{0.04}}$/8.91$_{\pm{0.07}}$  & 7.75$_{\pm{0.03}}$/10.55$_{\pm{0.07}}$  & 4.35$_{\pm{0.05}}$/7.35$_{\pm{0.10}}$  & 6.52$_{\pm{0.04}}$/9.41$_{\pm{0.07}}$  & 6.65$_{\pm{0.04}}$/9.89$_{\pm{0.06}}$  \\
    WDG-CP  & 3.84$_{\pm{0.00}}$/6.98$_{\pm{0.01}}$	&4.54$_{\pm{0.01}}$/7.29$_{\pm{0.03}}$&	5.47$_{\pm{0.02}}$/8.33$_{\pm{0.04}}$&	4.31$_{\pm{0.00}}$/7.28$_{\pm{0.02}}$&	4.96$_{\pm{0.01}}$/7.77$_{\pm{0.02}}$&	5.76$_{\pm{0.01}}$/8.96$_{\pm{0.03}}$ \\
    GLOSS  & 4.90$_{\pm{0.03}}$/7.95$_{\pm{0.07}}$	&5.80$_{\pm{0.03}}$/8.97$_{\pm{0.06}}$&	6.82$_{\pm{0.04}}$/10.26$_{\pm{0.08}}$&	5.28$_{\pm{0.04}}$/8.37$_{\pm{0.06}}$&	5.54$_{\pm{0.03}}$/8.67$_{\pm{0.05}}$&	7.05$_{\pm{0.03}}$/10.53$_{\pm{0.06}}$ \\
    \rowcolor{gray!15}LETC-separate  & 4.08$_{\pm{0.01}}$/6.91$_{\pm{0.02}}$& 4.84$_{\pm{0.01}}$/7.41$_{\pm{0.01}}$&	5.47$_{\pm{0.00}}$/8.32$_{\pm{0.02}}$&	5.01$_{\pm{0.02}}$/8.09$_{\pm{0.01}}$&	5.37$_{\pm{0.01}}$/7.97$_{\pm{0.01}}$&	5.83$_{\pm{0.01}}$/8.67$_{\pm{0.02}}$ \\
    % \midrule
    \rowcolor{gray!15}LETC($\tau=1$)  & \textbf{3.78$_{\pm{0.02}}$}/6.74$_{\pm{0.02}}$	& \textbf{4.44$_{\pm{0.01}}$}/\textbf{7.14$_{\pm{0.02}}$} & 5.35$_{\pm{0.00}}$/8.33$_{\pm{0.02}}$ & \textbf{3.96$_{\pm{0.03}}$}/\textbf{6.87$_{\pm{0.02}}$}	& \textbf{4.67$_{\pm{0.01}}$}/\textbf{7.35$_{\pm{0.02}}$}	& \textbf{5.44$_{\pm{0.00}}$}/\textbf{8.38$_{\pm{0.02}}$} \\
    \rowcolor{gray!15}LETC($\tau=2$)  & \textbf{3.78$_{\pm{0.00}}$}/\textbf{6.70$_{\pm{0.01}}$}&	4.45$_{\pm{0.00}}$/7.21$_{\pm{0.02}}$&	\textbf{5.34$_{\pm{0.01}}$}/\textbf{8.29$_{\pm{0.03}}$}&	4.13$_{\pm{0.01}}$/7.29$_{\pm{0.02}}$	&4.70$_{\pm{0.00}}$/7.46$_{\pm{0.01}}$&	5.54$_{\pm{0.01}}$/8.57$_{\pm{0.02}}$ \\
    \rowcolor{gray!15}LETC($\tau=3$)  & 3.84$_{\pm{0.01}}$/6.92$_{\pm{0.01}}$&	4.45$_{\pm{0.00}}$/7.23$_{\pm{0.02}}$&	5.39$_{\pm{0.00}}$/8.42$_{\pm{0.02}}$&	4.33$_{\pm{0.01}}$/7.62$_{\pm{0.01}}$	&4.86$_{\pm{0.00}}$/7.73$_{\pm{0.01}}$&	5.76$_{\pm{0.01}}$/8.91$_{\pm{0.02}}$ \\
    \bottomrule
    \multicolumn{5}{l}{\scriptsize{Best results are bold marked.}}
    \end{tabular}%
  \label{kriging_results}%
\end{table}%

\begin{table}[!htb]
  \centering
  \caption{Running time on PeMS-4W data}
  \footnotesize
    \begin{tabular}{l|cccccc}
    \toprule
    Models & \multicolumn{6}{c}{PeMS-4W,running time(in second)} \\
    \midrule
          & \multicolumn{1}{l}{SM0.3,TM0.2} & \multicolumn{1}{l}{SM0.5,TM0.2} & \multicolumn{1}{l}{SM0.7,TM0.2} & \multicolumn{1}{l}{SM0.3,TM0.5} & \multicolumn{1}{l}{SM0.5,TM0.5} & \multicolumn{1}{l}{SM0.7,TM0.5} \\
    \midrule
    TGMF  & 2511.6 & 2594.5 & 2739.2 & 2727.3 & 2795.1 & 2844.1 \\
    LSTM-GRMF & 8965.7 & 8465.3 & 8382.2 & 8172.3 & 8437.1 & 8715.3 \\
    FMDT-Tucker & 2892.8 & 9311.1 & 5783.6 & 4925.3 & 10902.1 & 6353.0 \\
    GLTL  & 7408.5 & 7808.0  & 6522.9 & 9673.7 & 6652.5 & 6118.5 \\
    KPMF  & 3488.1 & 1418.4 & \textbf{824.7} & 2343.9 & 4918.2 & 3763.3 \\
    WDG-CP  & 1979.5& 2057.7 &2056.6 & 2049.3&	2045.9&	2214.5 \\
    GLOSS  & 6390.8	&6384.9	& 6350.8&	6333.8&	6372.5&	6342.1 \\
    \rowcolor{gray!15}LETC  & \textbf{990.6} & \textbf{1075.1} & 1084.3 & \textbf{1067.9} & \textbf{1152.5} & \textbf{1164.3} \\		
    \bottomrule
    \multicolumn{5}{l}{\scriptsize{Best results are bold marked.}}
    \end{tabular}%
  \label{pems_time}%
\end{table}%

% It can be seen that LETC achieves the highest accuracy in most cases on both of the two datasets. 
As can be seen that LETC achieves the best kriging results consistently in all testing scenarios. By setting different $\tau$ values, LETC can adapt to various temporal perceptual fields.
With the decrease of both spatial and temporal observation rates, the performances of all models are degenerated while LETC is supposed to be more robust. 
Compared with performing imputation and kriging separately, the holistic LETC always achieves better results. This finding demonstrates the importance of modeling global and local correlations at the same time. On the one hand, local smoothness benefits low-rank tensor completion by acting as physical constraints. On the other hand, global information existing beyond local proximity provides more referable features for neighbor nodes to avoid over-smoothing.
Therefore, LETC can achieve network-wide kriging with high accuracy, even though there exists a large proportion of incomplete observations (missing data) in measured locations.

Observing the running time in Tab. \ref{pems_time}, we can find another superiority of LETC is that it can save nearly half the time cost compared to WDG-CP, which is the most scalable method in the baselines. As discussed in Section \ref{complexity analysis}, the computational bottleneck is the updating of graph regularization terms which would introduce large matrix equations, so the room for improvement is limited. Note that KPMF adopts stochastic gradient descent method to update parameters, which is susceptible to local minima, so it could converge to a inferior solution with fewer iteration times. 
LSTM-GRMF also produces competitive performances, especially when TM missing rate is high. Owing to the elaborate LSTM model, it can learn long-term temporal dependency and more complex trends in time series from training data, to produce reasonable estimates when less temporal data is given. However, the costs are a lot of training expenses and memory consumption, which are not suitable for network-wide applications.
Moreover, both CP, Tucker and MF based models in the baselines require an accurate rank estimation in advance and this parameter has great impacts on estimation accuracy and convergence. Although Algorithm \ref{r_tsvt} also contains a rank parameter, in later paragraphs we show that the rank in rSVD has minimal impact on accuracy.
Overall, the proposed LETC achieves state-of-the-art performances in all evaluating tasks, while requires the least computational burden at the same time.

\begin{figure}[!htb]
\centering
\subfigure[PeMS, SM3TM2]{
\centering
\includegraphics[scale=0.35]{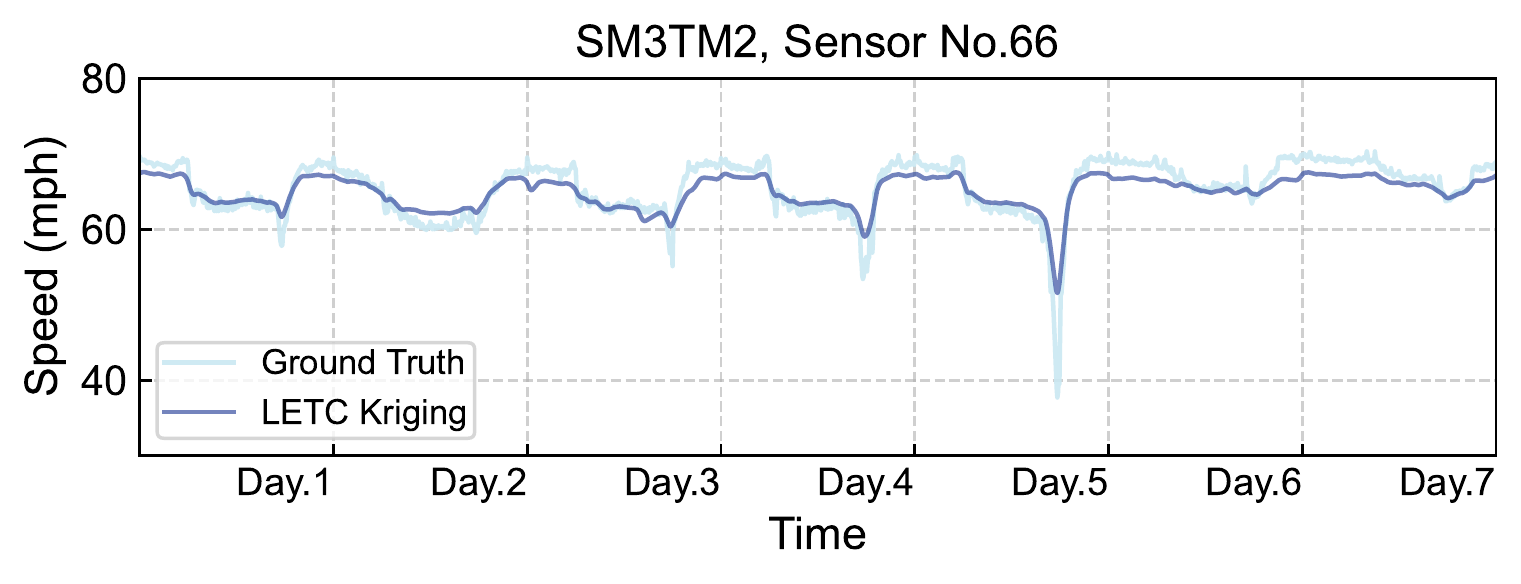}
}
\subfigure[PeMS, SM5TM2]{
\centering
\includegraphics[scale=0.35]{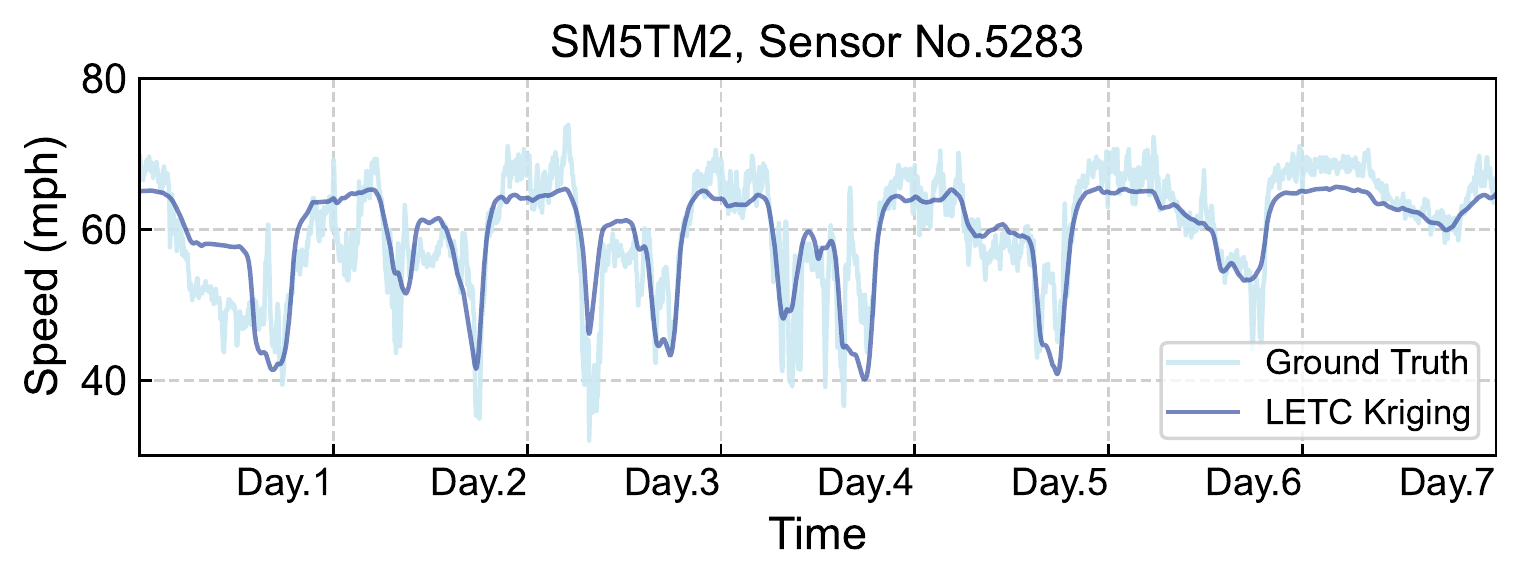}
}
\subfigure[PeMS, SM5TM5]{
\centering
\includegraphics[scale=0.35]{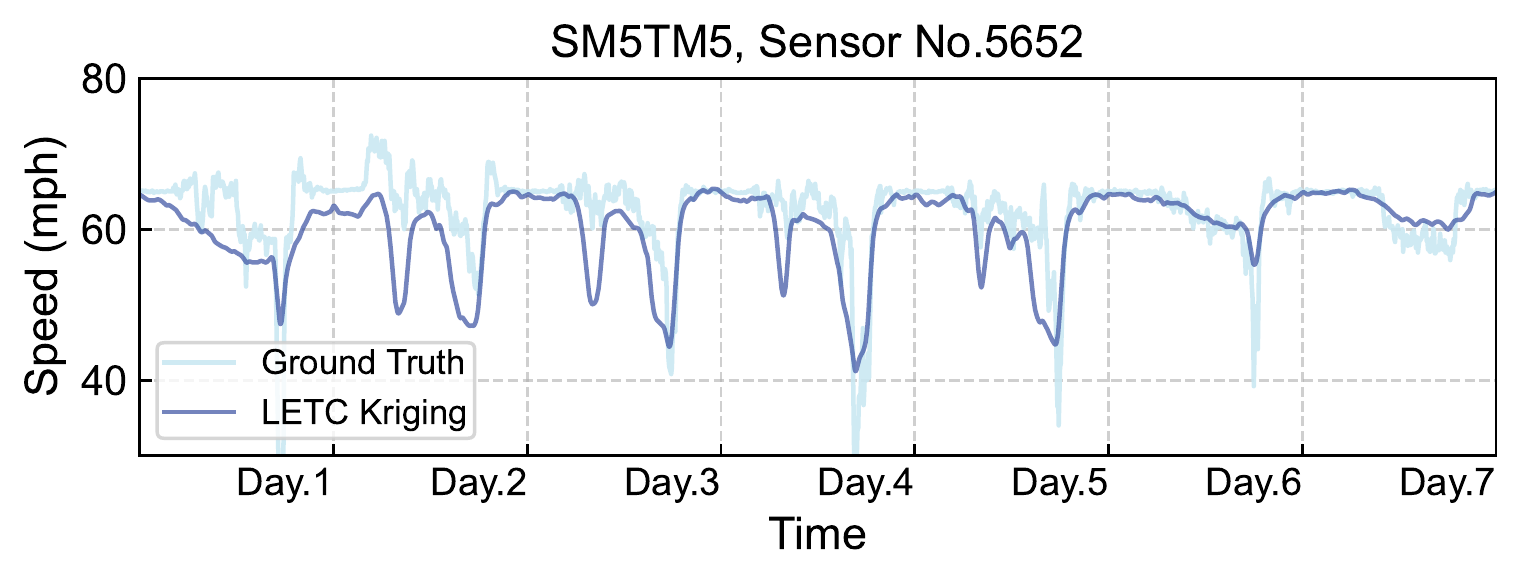}
}
\subfigure[PeMS, SM5TM5]{
\centering
\includegraphics[scale=0.35]{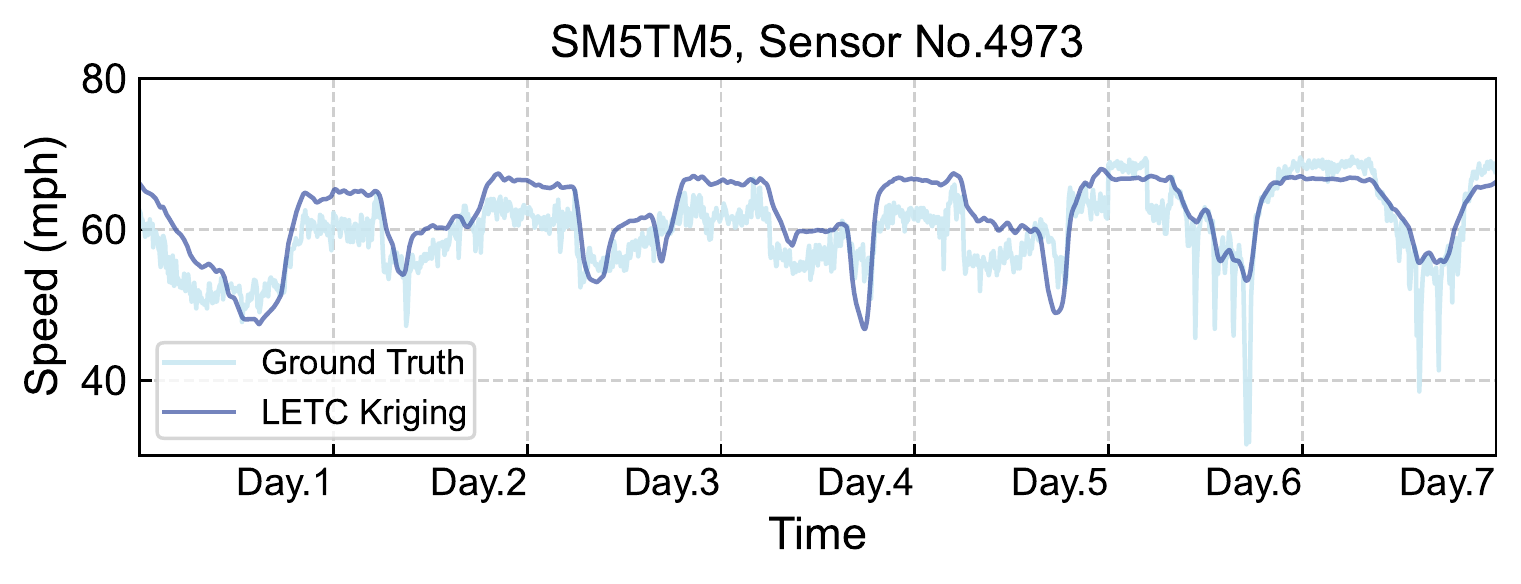}
}
\subfigure[PeMS, SM7TM2]{
\centering
\includegraphics[scale=0.35]{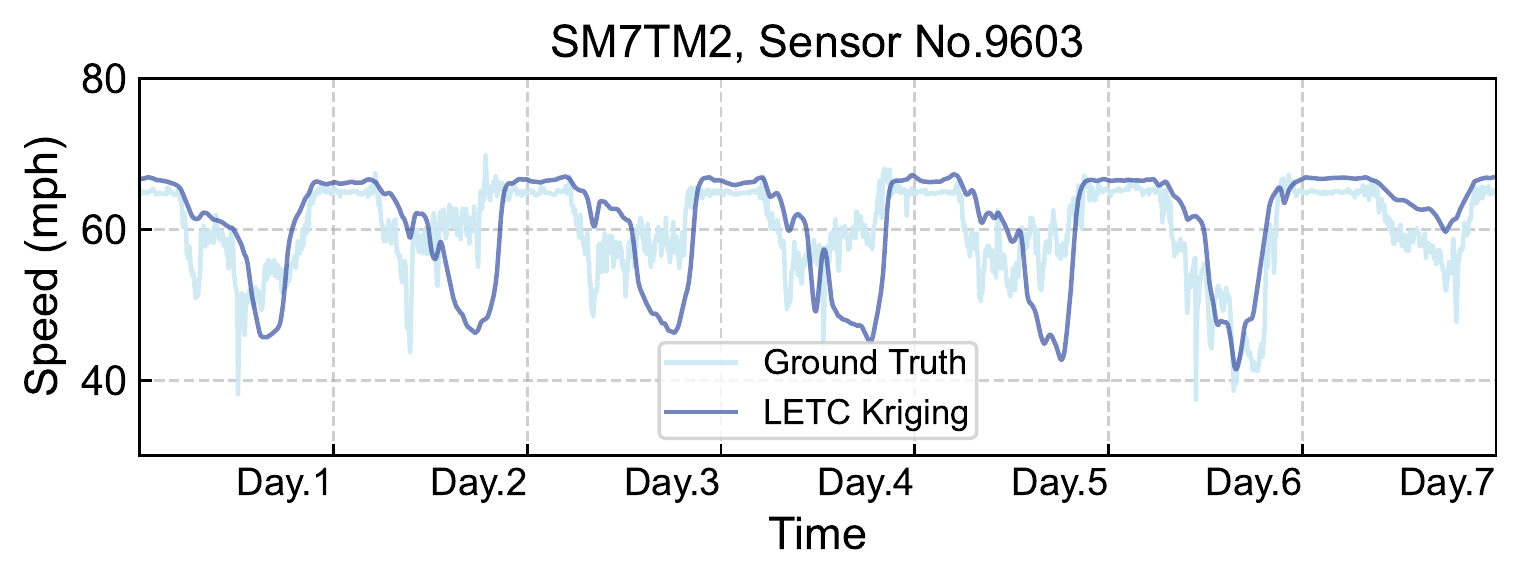}
}
\subfigure[PeMS, SM7TM5]{
\centering
\includegraphics[scale=0.35]{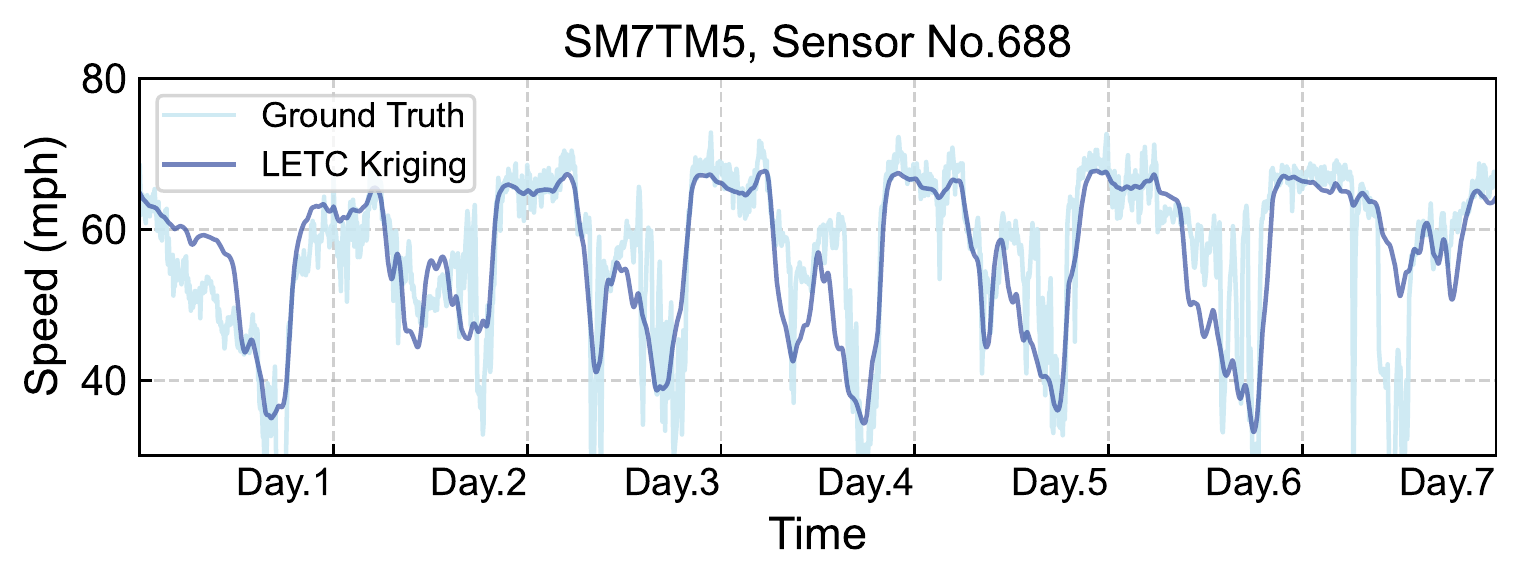}
}
\caption{Kriging results visualization for PeMS-4W data. Six randomly selected sensors with different observation conditions are shown. The ground truth values and estimation values are given for comparison.}
\label{pems_visual}
\end{figure}

\begin{figure}[!htb]
\centering
\subfigure[Portland, SM3TM2]{
\centering
\includegraphics[scale=0.35]{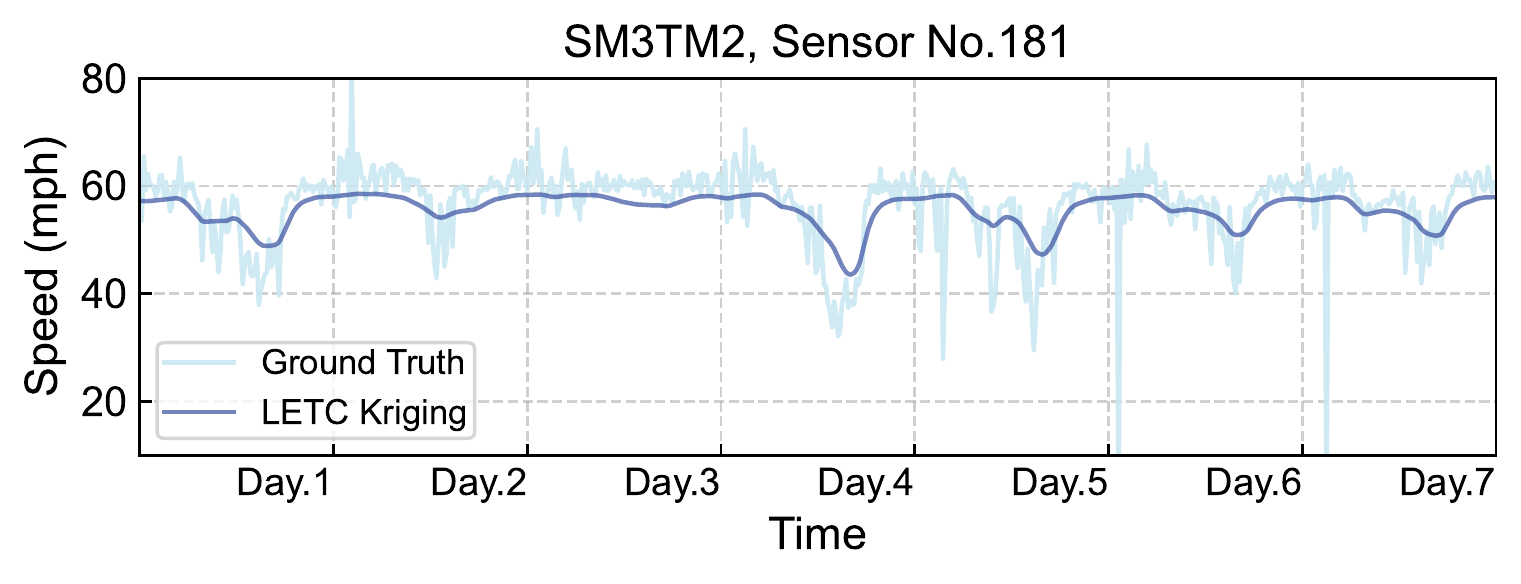}
}
\subfigure[Portland, SM5TM2]{
\centering
\includegraphics[scale=0.35]{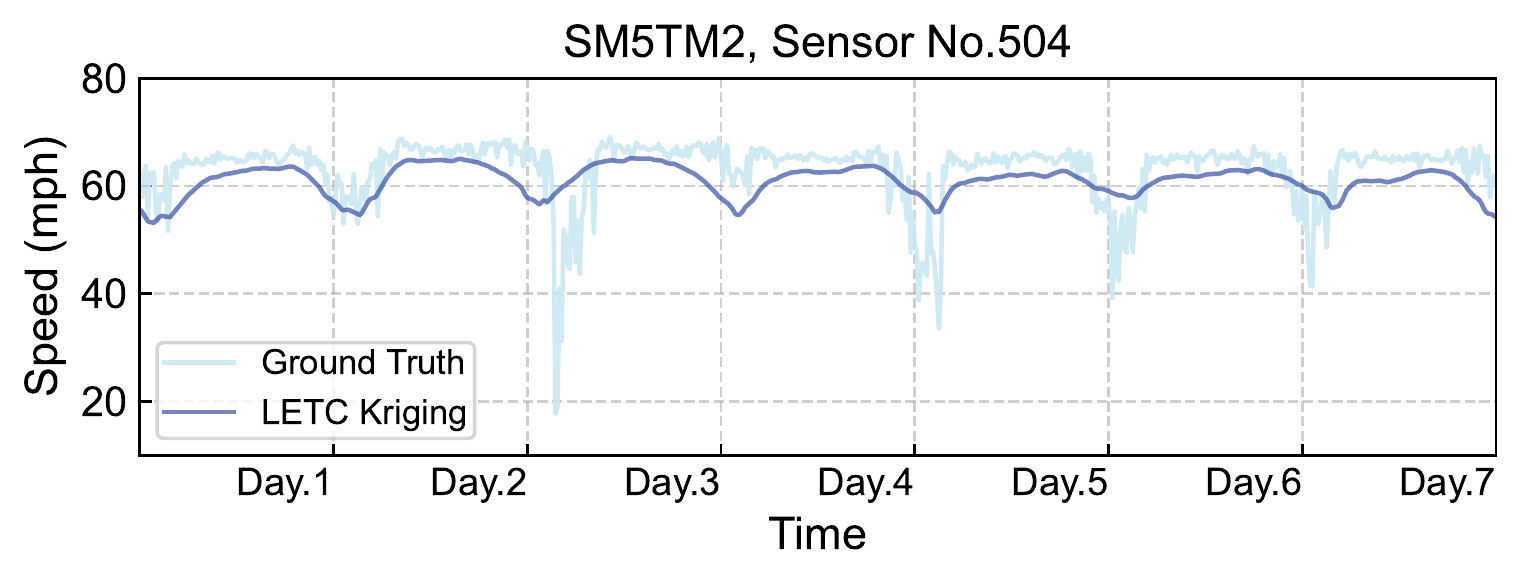}
}
\subfigure[Portland, SM5TM5]{
\centering
\includegraphics[scale=0.35]{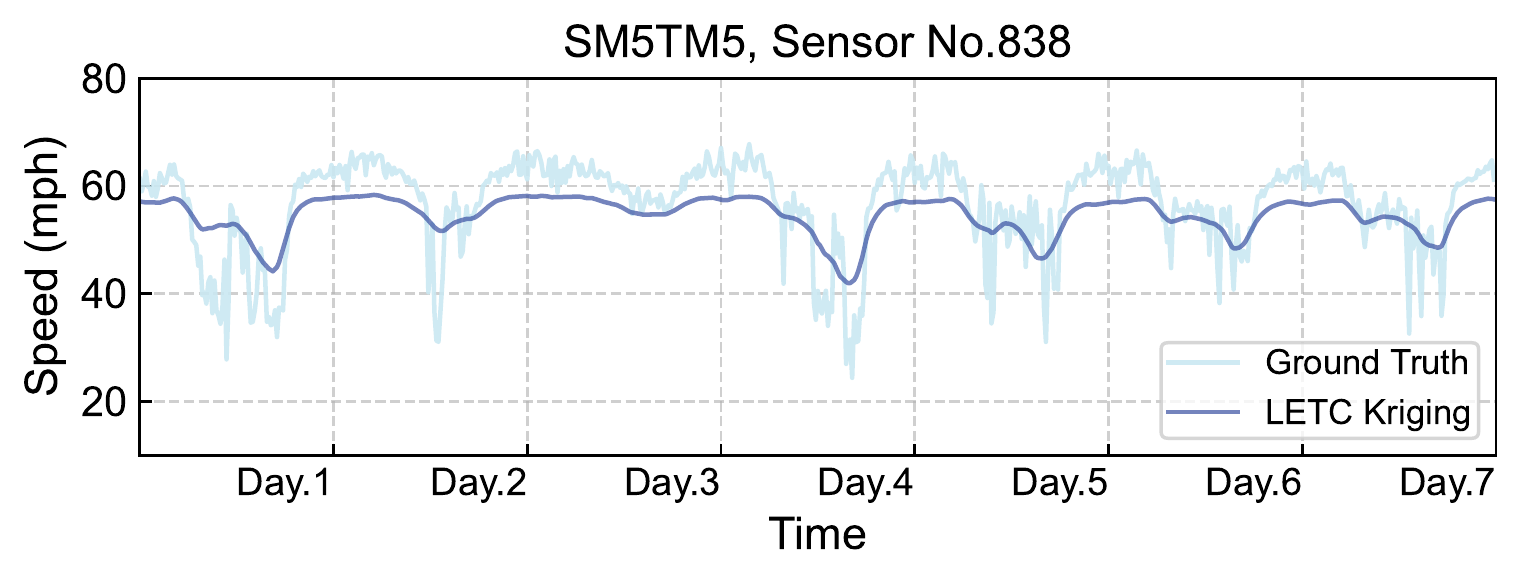}
}
\subfigure[Portland, SM7TM2]{
\centering
\includegraphics[scale=0.35]{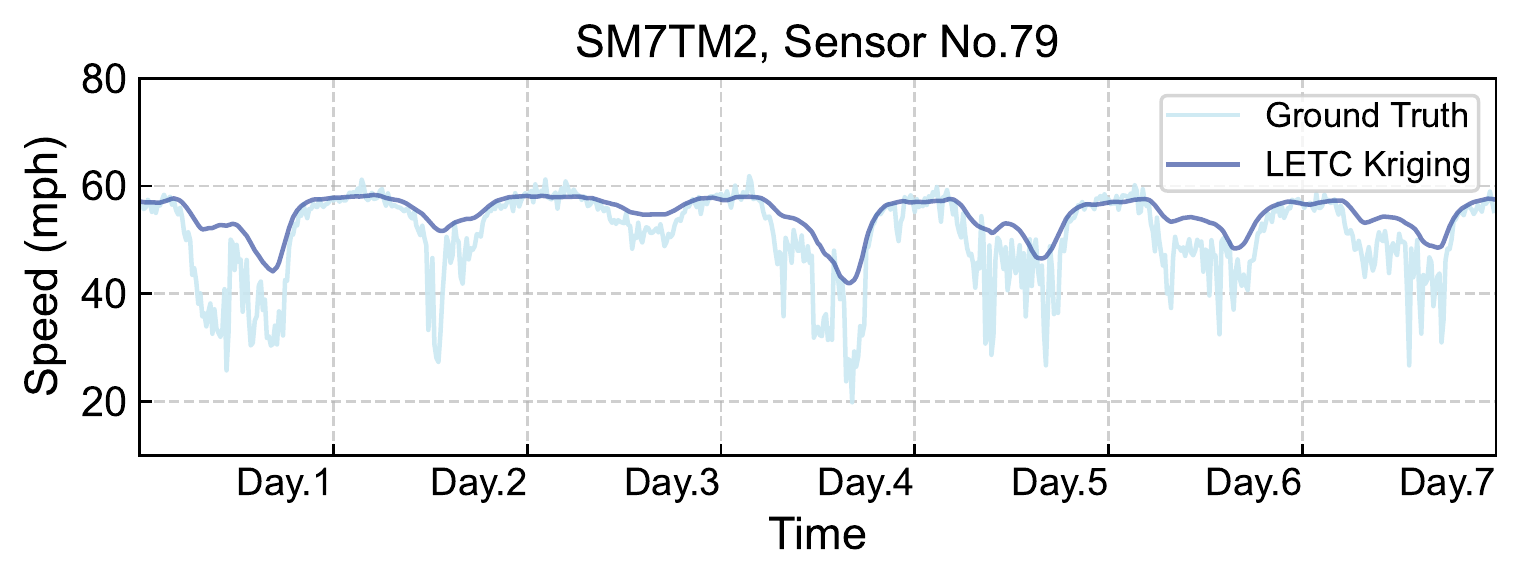}
}
\subfigure[Portland, SM7TM2]{
\centering
\includegraphics[scale=0.35]{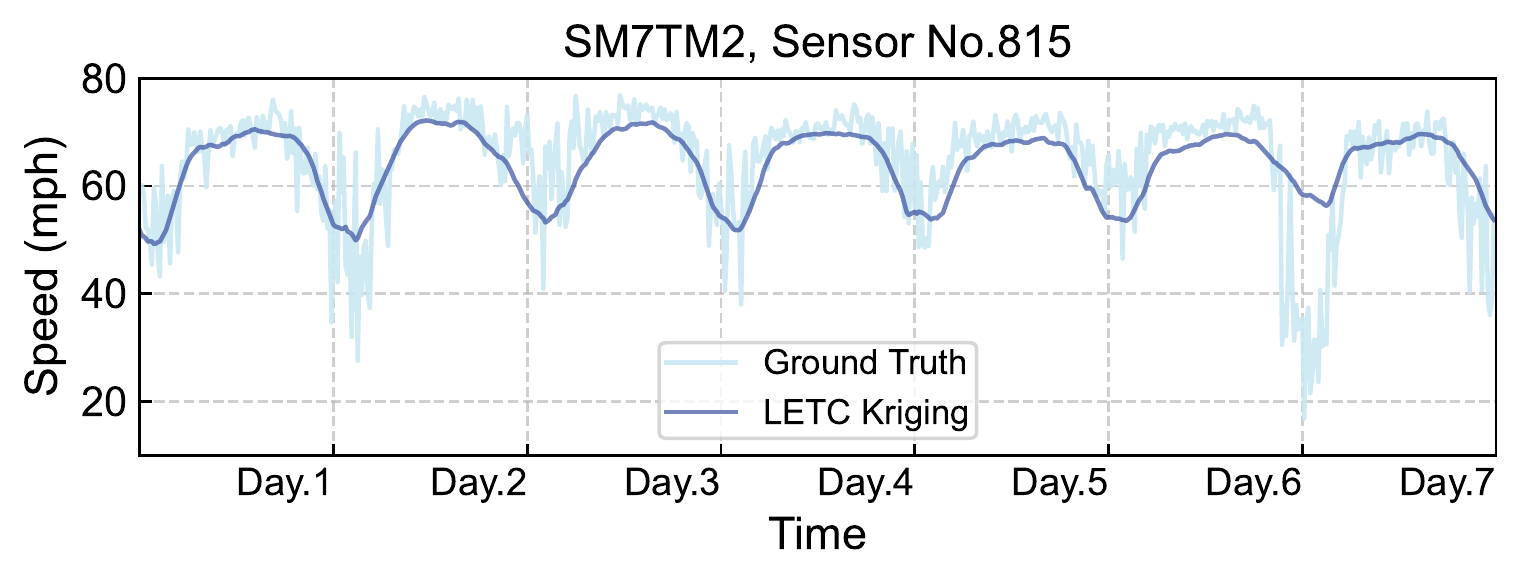}
}
\subfigure[Portland, SM7TM5]{
\centering
\includegraphics[scale=0.35]{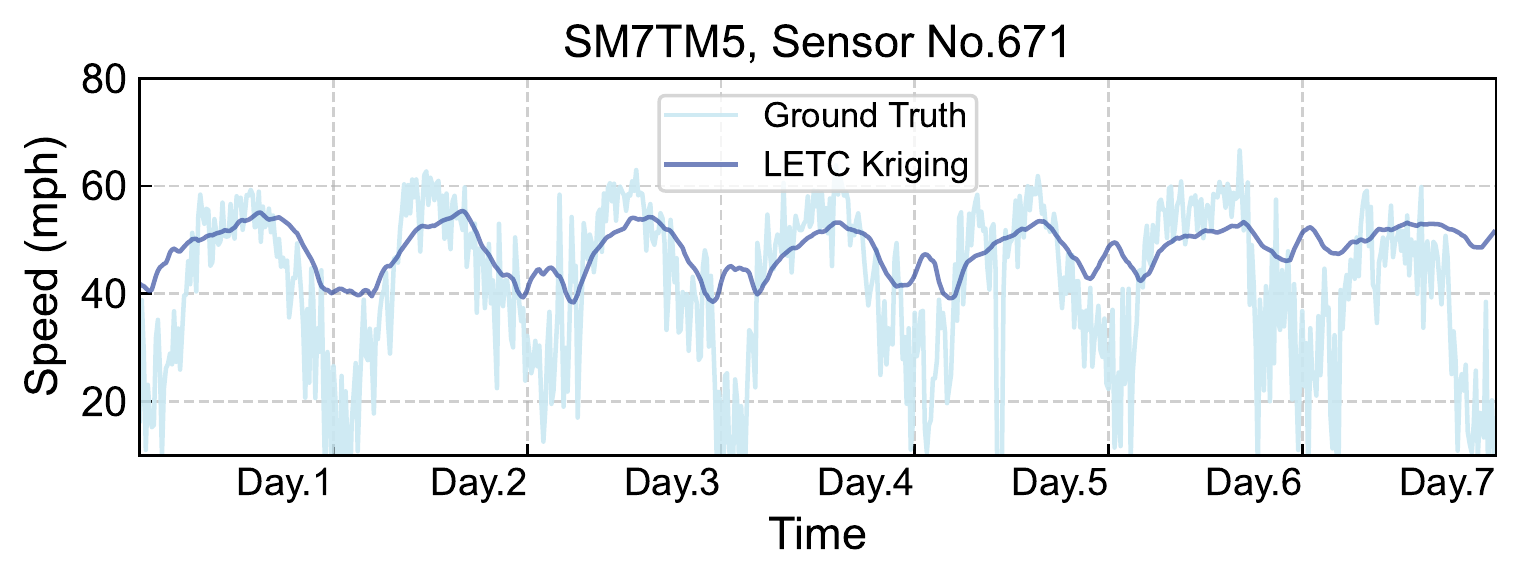}
}
\caption{Kriging results visualization for Portland data. Six randomly selected sensors with different observation conditions are shown. The ground truth values and estimation values are given for comparison.}
\label{port_visual}
\end{figure}

Furthermore, we give some kriging results on both two data sets in Fig. \ref{pems_visual} and \ref{port_visual} to intuitively show the speed estimation performances. Each figure depicts the estimation value and ground truth value for roads randomly selected from unmeasured locations. Overall, LETC estimates accurate time series for unknown locations, even with low observation rate as well as temporal loss. Another facet can be found that LETC produces relatively more smooth and denoised signals than the raw data due to the temporal regularization and low-rank condition.

\begin{figure}[!htbp]
\centering
\includegraphics[scale=0.5]{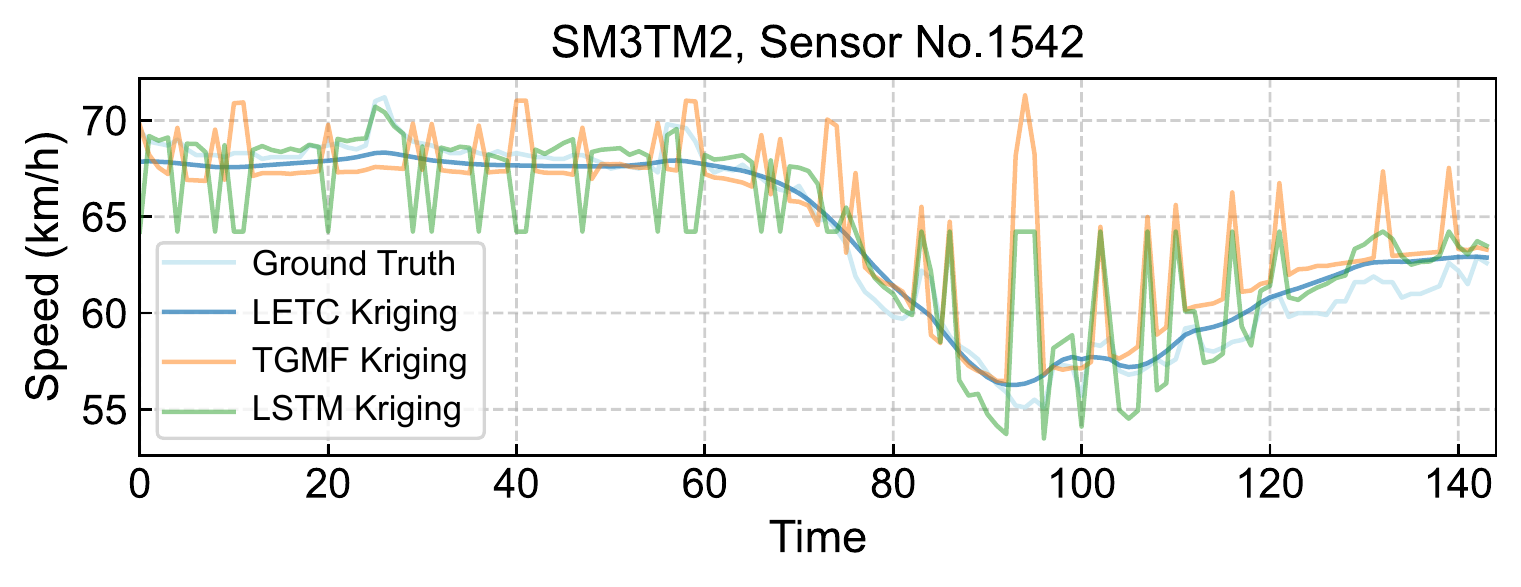}
\caption{Detailed comparison of different kriging models.}
\label{compare_example}
\end{figure}

We also give the results of half a day on the first day of PeMS data for detailed comparison in Fig. \ref{compare_example}.
As can be seen, LETC produces more smooth and stable time series than TGMF and LSTM-GRMF. With the elimination of data noises, LETC estimation is closer to the ground truth value. Because there is totally no historical data available, the LSTM model which is widely used to perform time-dimension prediction, is less effective and fluctuates a lot. 

\begin{figure}[!htb]
\centering
\subfigure[Ground truth speed values]{
\centering
\includegraphics[scale=0.14]{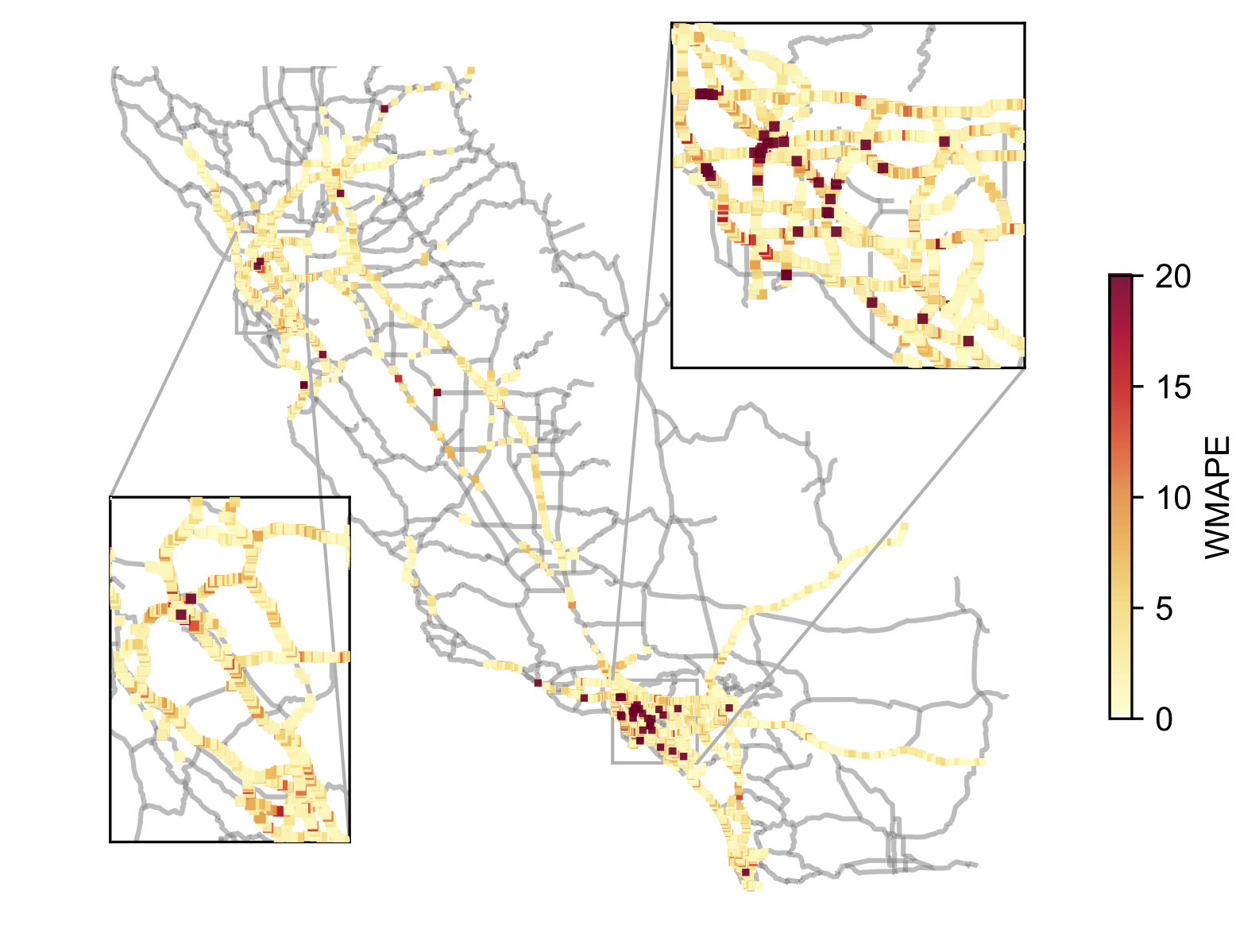}
}
\centering
\subfigure[LETC kriging speed values (Proposed method)]{
\centering
\includegraphics[scale=0.14]{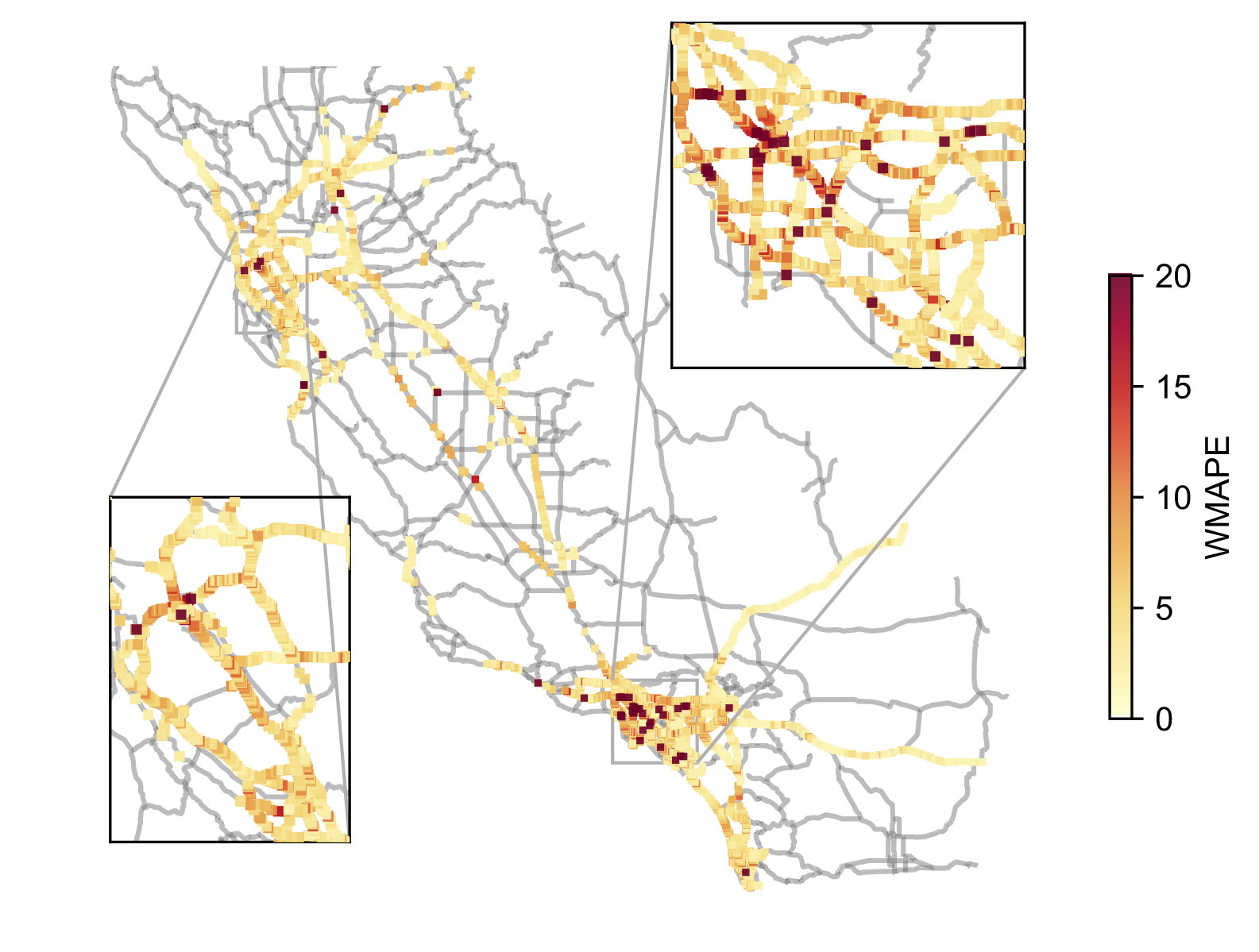}
}
\caption{WMAPE of each sensor on the PeMS highway network. (a) WMAPE of LETC. (b) WMAPE of LSTM-GRMF.}
\label{pems_network_error}
\end{figure}

The ground truth speed values, LETC and LSTM-GRMF kriging speed values as well as the estimation residuals (absolute value of the difference between the true value and the estimated value) of LETC are visualized on the whole California highway network in Fig. \ref{pems_network} (more examples are given in \nameref{Appendix B.}). Moreover, we also display the Weighted Mean Absolute Percentage Error (WMAPE) of each sensor in Fig. \ref{pems_network_error}.
\begin{figure}[!htb]
\centering
\subfigure[Ground truth speed values]{
\centering
\includegraphics[scale=0.14]{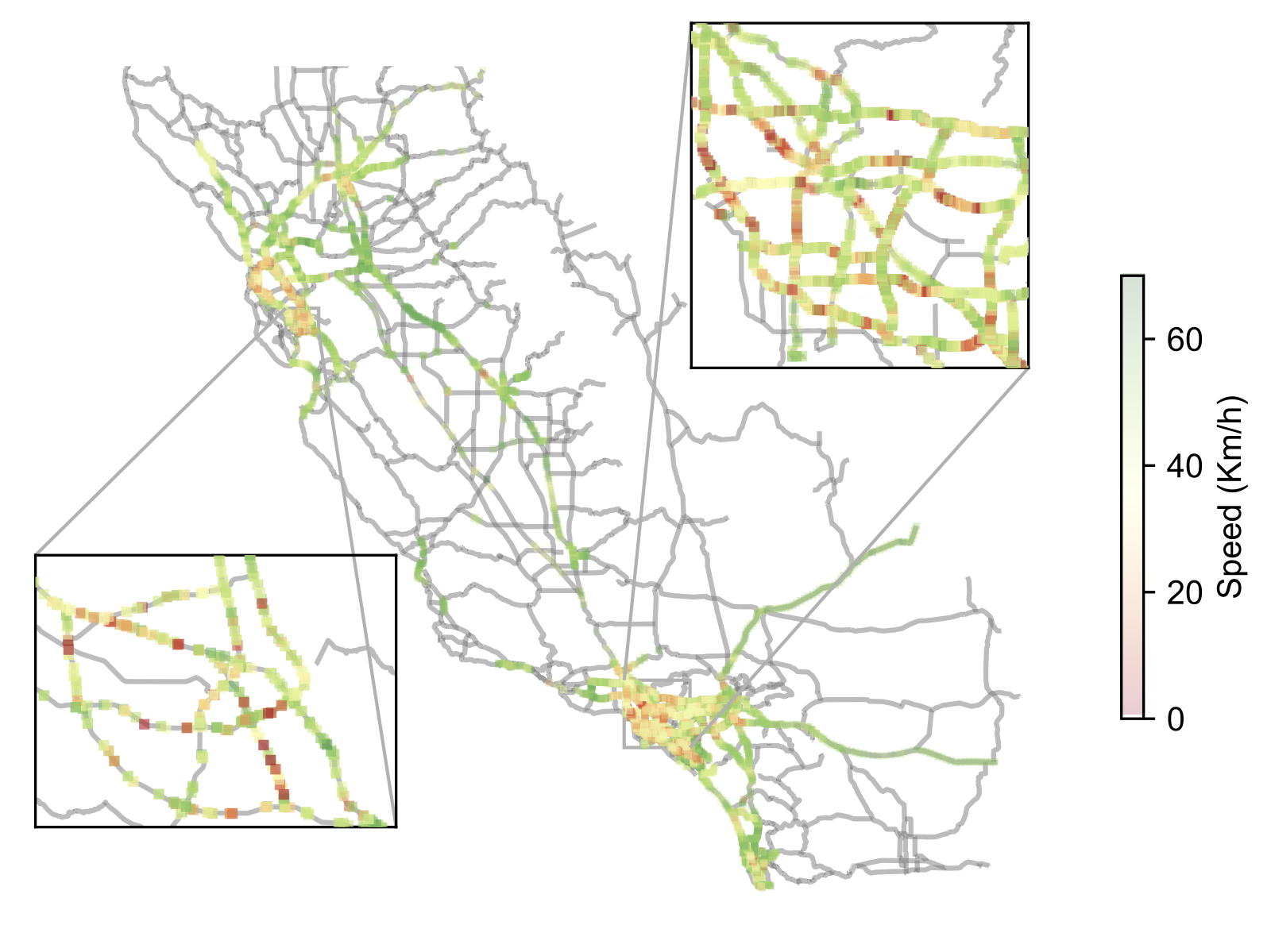}
}
\centering
\subfigure[LETC kriging speed values (Proposed method)]{
\centering
\includegraphics[scale=0.14]{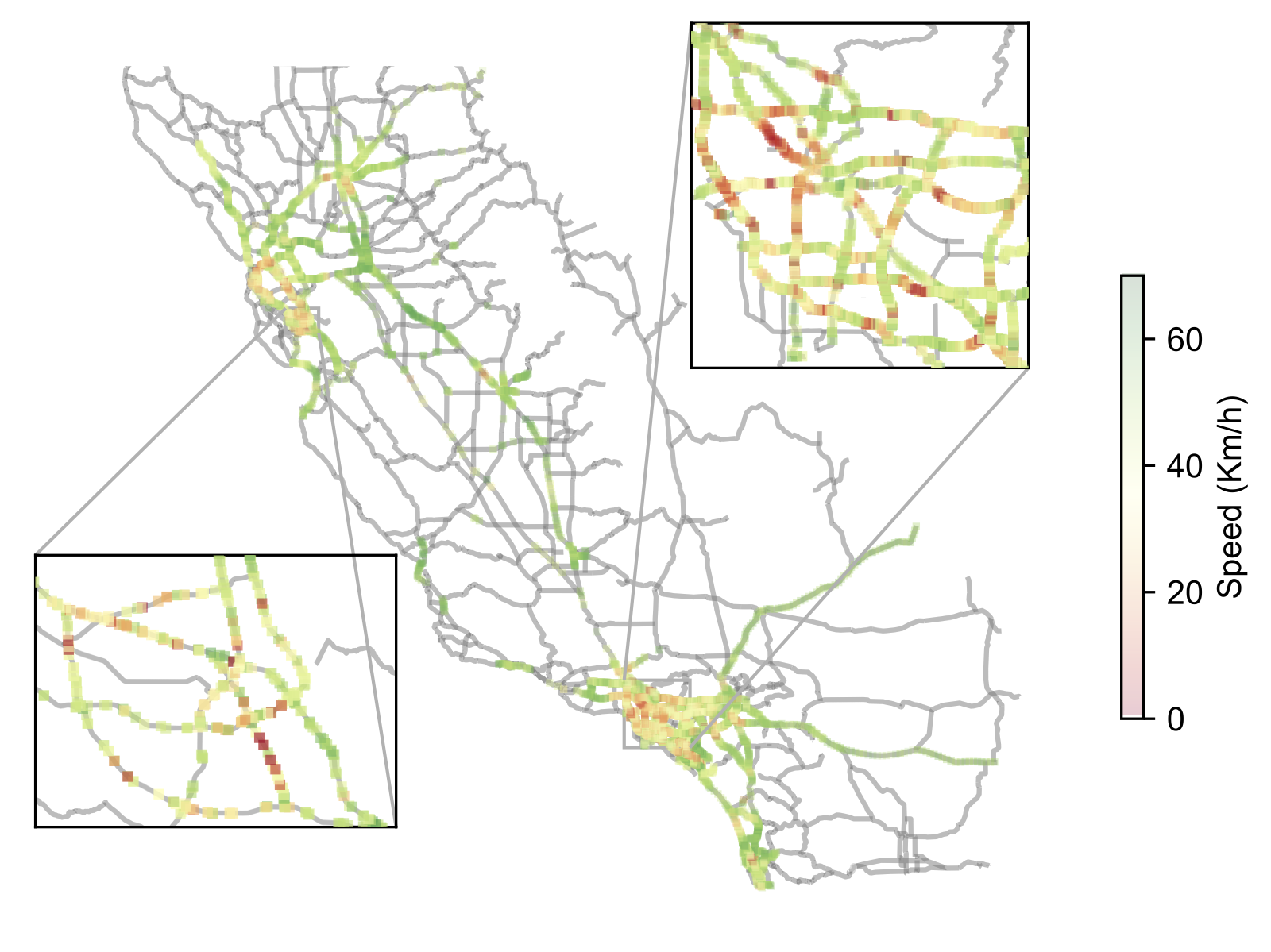}
}
\centering
\subfigure[LETC kriging residuals]{
\centering
\includegraphics[scale=0.14]{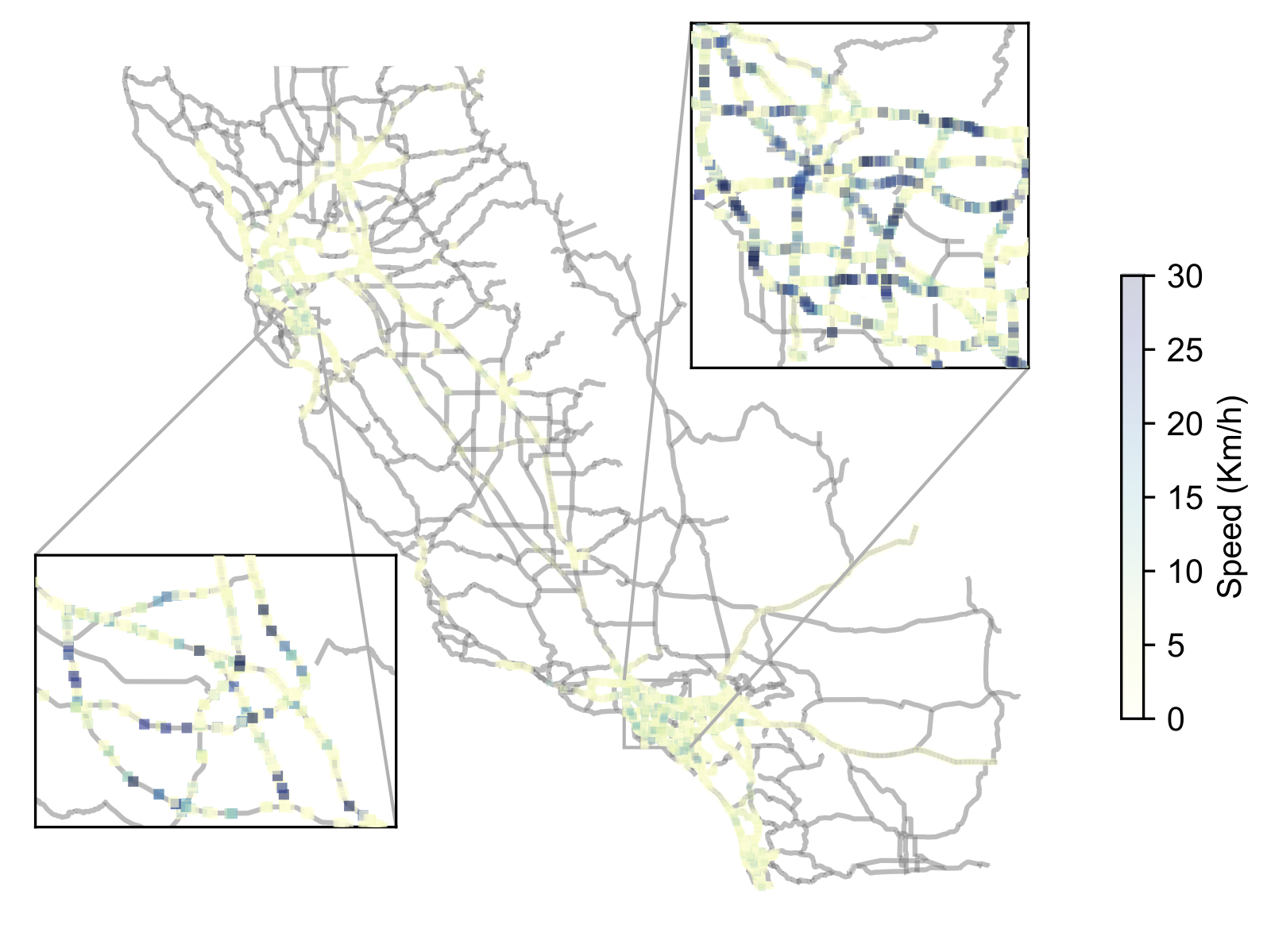}
}
\centering
\subfigure[LSTM-GRMF kriging speed values (Best benchmark)]{
\centering
\includegraphics[scale=0.14]{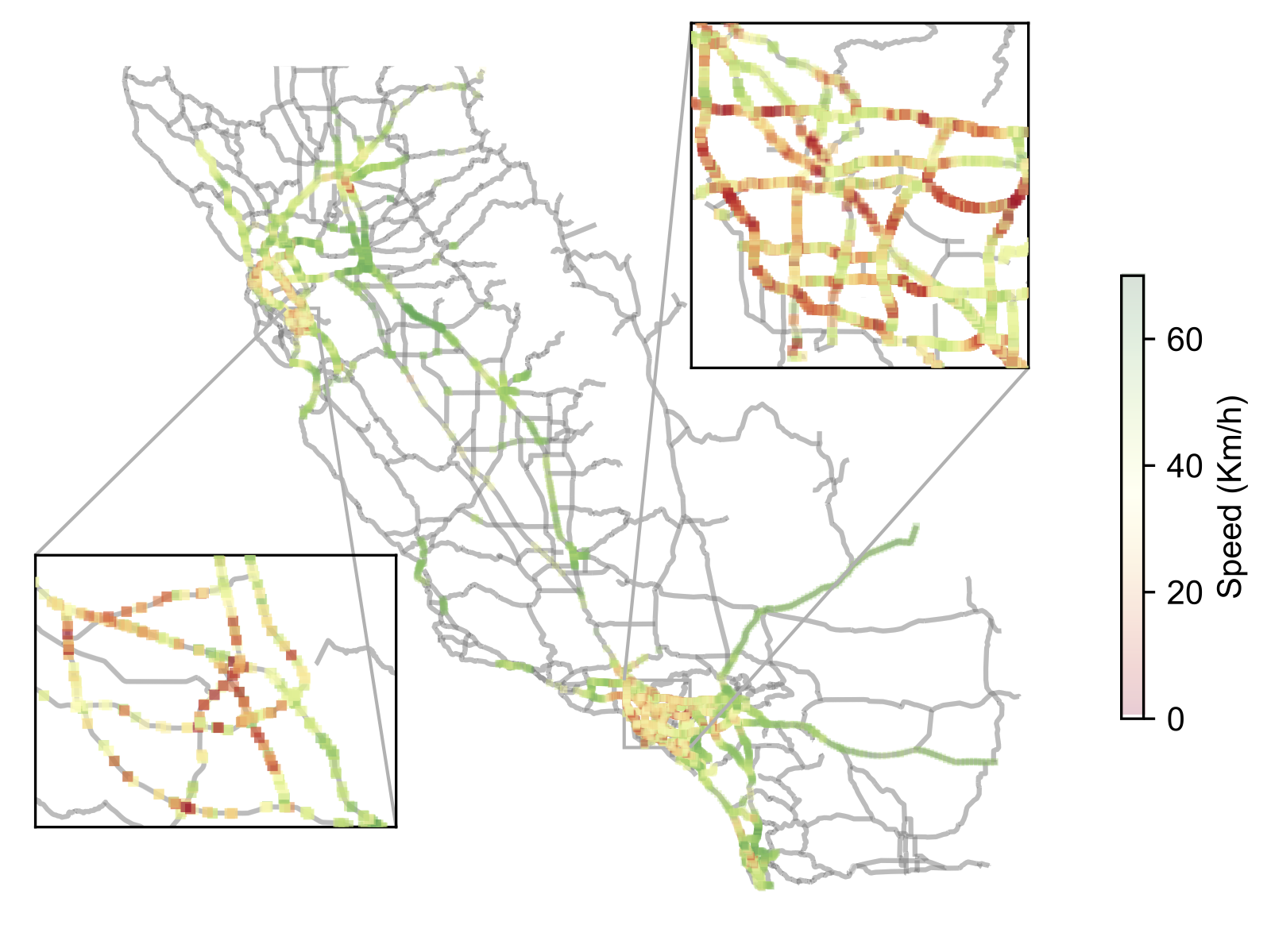}
}
\caption{LETC kriging results on the PeMS highway network. Traffic state at 7:00 in the morning is displayed as an example. (a) Ground truth values. (b) LETC estimated traffic speed values. (c) LETC kriging residuals of each sensor. (d) LSTM-GRMF estimated traffic speed values.}
\label{pems_network}
\end{figure}

As we can see, LETC can capture not only the global spatial distributions of traffic speed, but also the speed changes due to local congestion of morning peak. Compared (a), (b) with (d), LETC produces more realistic speed patterns than LSTM-GRMF, which is the best model in the baselines, especially for the local speed changes in congestion areas. Fig. \ref{pems_network} (c) also shows the LETC estimated and true speed values fit well, with more than $90\%$ of road segments whose residual values are less than $5$.
The kriging errors of each link in Fig. \ref{pems_network_error} (a) demonstrate that relatively big errors occur in the areas where there is few or even no neighbor sensors, or congestion districts (e.g., south Bay area) with drastic changes in traffic speed. As kriging is in fact to aggregate the data of neighbor sensors, the estimation difficulty increases when little referable information is available. Overall, LETC produces smaller sensor-level errors than benchmark model on average, especially for those sensors located at congestion areas.

\subsection{Algorithmic analysis}
\subsubsection{Ablation studies}
To inspect the significance of each component of the proposed model, in this part we conduct ablation studies to compare the performances of several model variations. These variations include three aspects: accelerating algorithms, tensor nuclear norm, and spatiotemporal regularization.

Ablation studies about accelerating algorithms:
\begin{itemize}
    \item \textbf{LETC-CG iter3} and \textbf{LETC-CG iter10}: To test the impact of the numbers of CG inner loops, we set 3 and 10 iterations for implementing Algorithm \ref{CG}.
    \item \textbf{LETC-nCG}: LETC without conjugate gradient. The matrix equation in Eq. \eqref{sylvester} is solved by using the Bartels-Stewart method and this algorithm is achieved by directly calling a package API function $\operatorname{scipy.linalg.solve\_sylvester(\cdot)}$.
    \item \textbf{LETC-nRSVD}: LETC without randomized SVD. The economic (compact) SVD is used instead.
\end{itemize}

Ablation studies about unitary transform and TNN:
\begin{itemize}
    \item \textbf{LETC-FFT}: We replace TGFT with traditional FFT, then the t-TNN becomes the previously developed tensor nuclear norm \citep{lu2019tensor}.
    \item \textbf{LETC-GFT}: We replace the TGFT in our model by the original GFT used in \citep{deng2021graph}. To achieve this adjustment, we first transpose the input tensor into shape of $(\text{time of day}\times\text{days}\times\text{locations})$ and calculate the graph operator in Eq. \eqref{gft_cal} by spatial Laplacian. Then, the GFT is performed on the 'location' mode.
    \item \textbf{LETC-ROM}: Perform unitary transform by constructing random orthogonal matrix \citep{lu2019low}.
    \item \textbf{LETC-SVD}: Perform unitary transform by computing the SVD of mode-3 unfolding \citep{chen2021scalable}.
    \item \textbf{LETC-SNN}: We replace t-TNN by sum of nuclear norm (SNN), which is another widely used tensor norm definition developed by \citep{liu2013tensor}.
\end{itemize}

Ablation studies about spatial and temporal dependency regularization:
\begin{itemize}
    \item \textbf{LETC-circulant}: LETC with circulant and undirected temporal Laplacian by setting $\mathbf{L}_{\Phi}'=\mathbf{L}_{\Phi}\mathbf{L}_{\Phi}^{\mathsf{T}}$ and $\mathbf{\Phi}=\mathbf{I}_T$.
    \item \textbf{LETC-symmetric}: LETC with symmetric graph Laplacian regularization, which is commonly used for spatial modeling.
\end{itemize}

\begin{table}[!htbp]
  \centering
  \caption{Ablation studies}
  \footnotesize
    \begin{tabular}{l|cc}
    \toprule
          & \multicolumn{2}{c}{Portland} \\
    \midrule
    Models & MAE/RMSE & \multicolumn{1}{c}{Running time (in second)} \\
    \midrule
    \rowcolor{gray!15}LETC(CG iter3) & 4.61/7.40 & 145.2 \\
    LETC(CG iter10) & 4.62/7.39 & 369.2 \\
    LETC-nCG & 4.62/7.41 & 4976.7 \\
    LETC-nRSVD & 4.65/7.45 & 194.5 \\
    \midrule
    LETC-circulant & 4.77/7.54 & 146.9 \\
    LETC-symmetric & 4.82/7.51 & 148.0 \\
    \midrule
    LETC-FFT & 4.71/7.50 & 133.8 \\
    LETC-GFT & 4.77/7.55 & 641.0 \\
    LETC-ROM & 4.69/7.42 & 146.3 \\
    LETC-SVD & 4.60/7.38 & 176.5 \\
    LETC-SNN & 4.72/7.50 & 345.7 \\
    \bottomrule
    \end{tabular}%
  \label{ablation_results}%
\end{table}%
We test the above model variations on Portland data and report the resulting performances and running time in Tab. \ref{ablation_results}. Among these models, LETC with CG iter3 can be viewed as the benchmark. 
First, by observing the results of LETC-nRSVD and LETC-nCG that are versions without numeric approximation algorithms, we can find that the numeric accelerating methods in this work, e.g., rSVD and CG, have no negative impact on kriging precision, while reducing the computational burden at some degree. Especially, the running time of LETC-nCG is prohibitive, which means that CG plays a key role in kriging scalability. By comparing the results of LETC(CG iter3) with LETC(CG iter10), an interesting observation is that the increase of iteration times of CG would not benefit the kriging accuracy in this condition. The underlying reason for this results is that Eq. \eqref{vectorization} is highly structure so that CG could convergence with only a few iterations.

Second, model becomes much less effective when replacing the proposed spatial and temporal regularization. Model with symmetric graph Laplacian shows the worst precision. Such results indicates the necessity of taking the moving direction of traffic flow and causality of time series into consideration.

Third, by comparing different unitary transform horizontally, it is observed that the proposed TGFT is a good balance of accuracy and efficiency. FFT features fast computation by utilizing the conjugate symmetry property, but has worse completion accuracy. GFT shows inferior performances on both precision and time cost. Since the GFT is conducted on the 'location' mode, the tensor SVT needs to be computed a large number of times, which is not cost-effective for network-wide applications. It is noteworthy that SVD based unitary transform also gives competitive results. This method obtains the transform operator by a data-driven and iterative way, which requires more data and time cost to update the transform matrix. In addition, as SVD needs to be performed on each unfolding matrix, SNN requires more computation resources to converge and shows inferior accuracy than TNN. This observation is consistent with the findings in \citep{lu2019low}. Overall, TGFT offers an explainable tool to perform TNN minimization, while at the same time keeps high accuracy and efficiency. Ones can incorporate prior knowledge (e.g., periodicity property) about spatiotemporal traffic data into tensor completion through TGFT.

\subsubsection{Influence of graph partitioning}
\label{graph_part}
One significant superiority of LETC is that it can achieve efficient kriging at network-wide and one may wonder why the scalability of kriging model is important 
from the perspective of algorithmic performance.
To demonstrate the necessity and significance of kriging on large-scale data, in this section we conduct experiments on different data size using both of the two datasets. Specifically, we adopt the same graph partitioning method as in \citep{chen2021scalable} to divide the adjacent matrices in $\{1,2,4,8,16,32,64\}$ parts for PeMS data and $\{1,2,4,8,16,32\}$ parts for Portland data. The more partitions, the less data scale of each group. In each division case, we conduct kriging on all partitions and record the accuracy and running time of the whole data.

\begin{figure}[!htb]
\centering
\subfigure[PeMS, LETC]{
\centering
\includegraphics[scale=0.5]{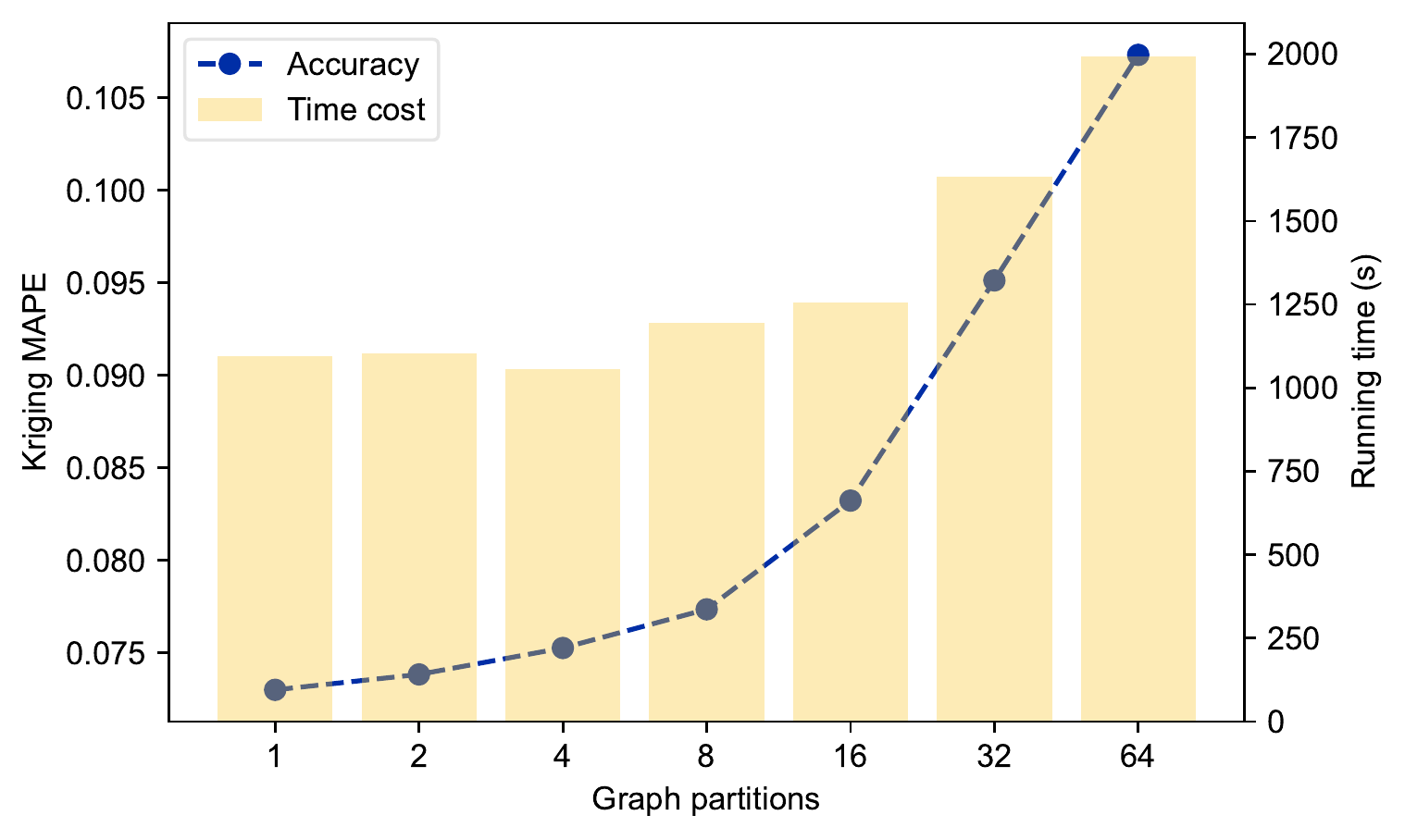}
}
\subfigure[PeMS, TGMF]{
\centering
\includegraphics[scale=0.5]{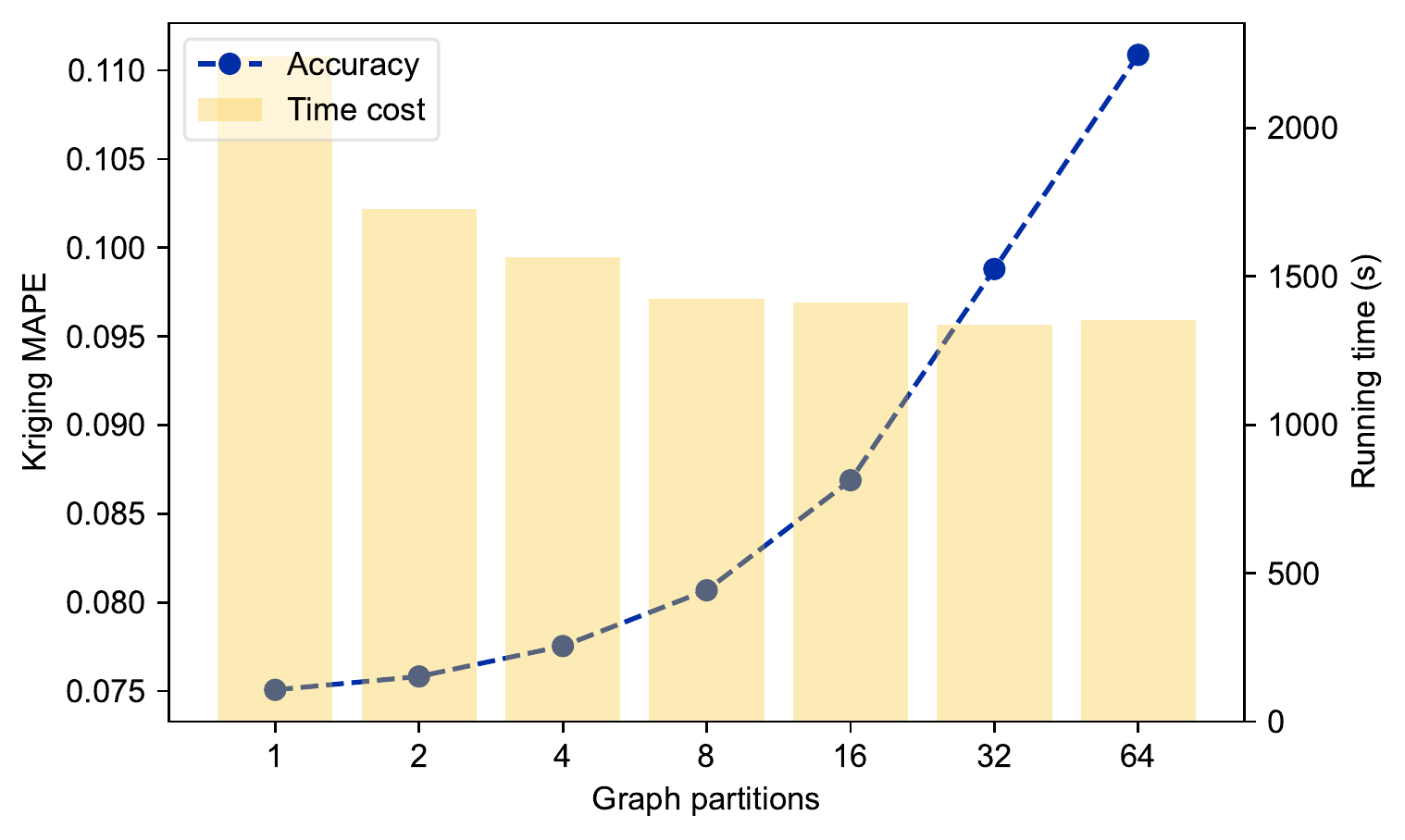}
}
\caption{Model performances under different graph partitions on PeMS-4W data. (a) The proposed LETC. (b) TGMF model. The kriging accuracy and running time are both shown in this figure.}
\label{pems_part}
\end{figure}

\begin{figure}[!htb]
\centering
\subfigure[Portland, LETC]{
\centering
\includegraphics[scale=0.5]{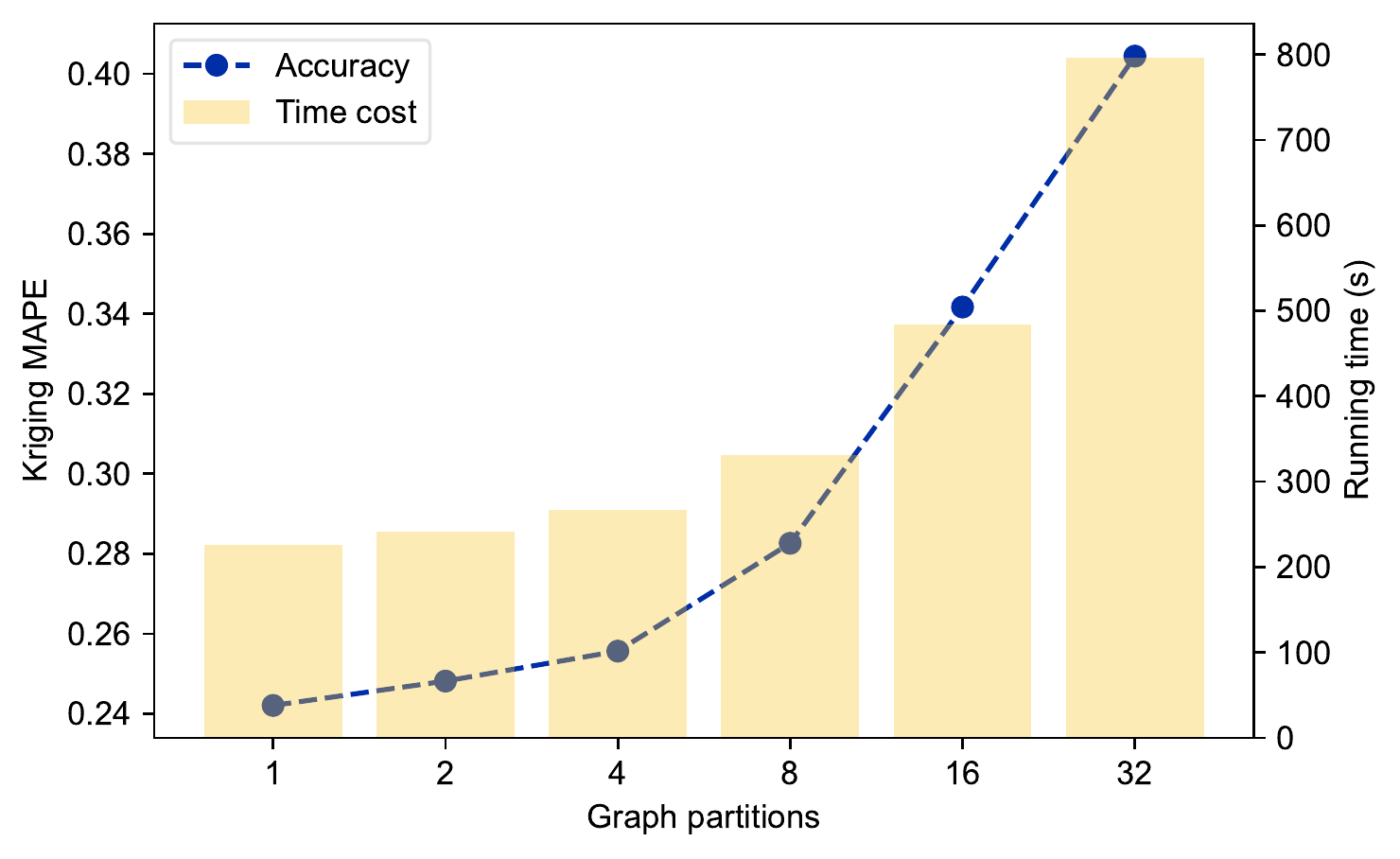}
}
\subfigure[Portland, TGMF]{
\centering
\includegraphics[scale=0.5]{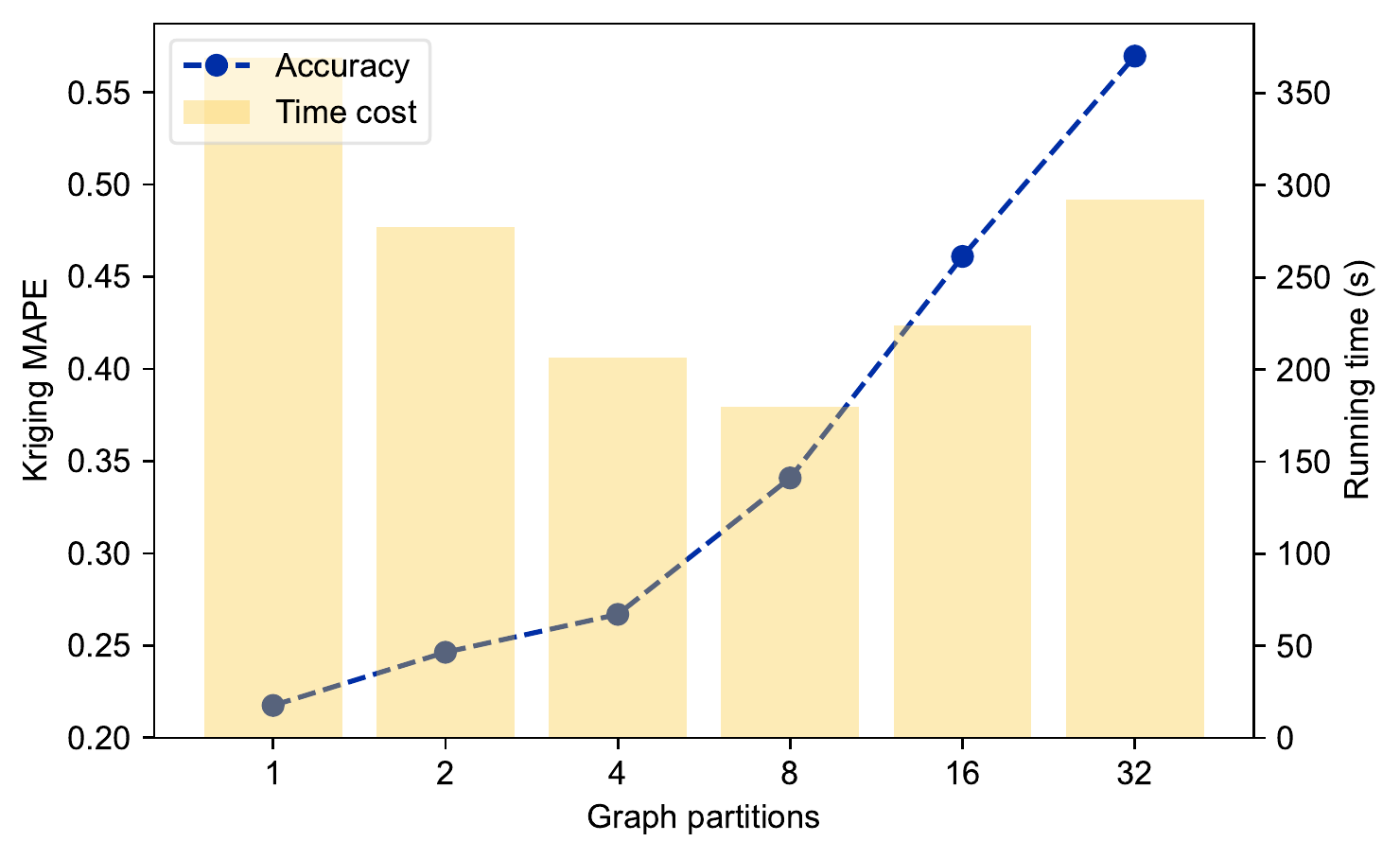}
}
\caption{Model performances under different graph partitions on Portland data. (a) The proposed LETC. (b) TGMF model. The kriging accuracy and running time are both shown in this figure.}
\label{port_part}
\end{figure}

Figs. \ref{pems_part} and \ref{port_part} show the changes of kriging performances of our LETC and another baseline method TGMF with the increase of graph partitions (decrease of data scales). As can be seen, for both of the two data, the kriging accuracy of LETC and TGMF decrease remarkably and the best performances are achieved when there is no partitioning. The reasons to explain this phenomenon could be twofold: 1) Previous study \citep{chen2021scalable} has demonstrated that low-rank tensor completion model could achieve better performances on randomly missing data imputation tasks when the data size is large. And in this work, we aim at kriging with incomplete observations and the t-TNN minimization part of our model could benefit from large data scale and learn a better low-rank pattern from a wider range. 2) The spatial graph Laplacian is at the core of kriging model, and it is directly obtained from the network topology structure. When the network is larger, more complicated and comprehensive spatial relations are incorporated in adjacent matrix and the influence of local anomaly can be weakened. Therefore, for unobserved locations there are more informative neighborhoods to provide referable observations when the network is larger.

Another important observation is that the running time of LETC also gradually increases with more partitions, while the results of TGMF is quite different. Generally speaking, LETC entails the least computation cost when there exists no partitioning while it becomes the largest for TGMF in this situation. This may stem from the reason that LETC can reduce the rank of any arbitrary tensor size in the objective function, but TGMF with a fixed rank settings can not adapt to varying input data size. 

Therefore, above results reveal that the proposed LETC that has better kriging scalability than other baseline models is very suitable for large-scale kriging and the motivations behind kriging at network-wide are consideration of both efficiency and precision.

\subsubsection{Sensitivity analysis of hyper-parameter}
\label{hpt}
It is generally acknowledged that the hyper-parameter tuning is a long-standing topic in machine learning communities. Most recently, some newly emerged Bayesian hyper-parameter optimization framework has been successfully applied to tensor/matrix factorization models to alleviate this problem \citep{lei2022bayesian,chen2022bayesian}. However, learning the model hyper-parameters on network-wide data set may be quite difficult. In this work, two main hyper-parameters of LETC that have obvious impact on kriging precision are the regularization factors $\lambda_1$ and $\lambda_2$. To discuss about the influence of $\lambda_1$ and $\lambda_2$, we examine the model performances under different parameter settings. Moreover, another potential factor that may affect the sensitivity is the size of input data, we also take it into consideration.

In LETC, the relative magnitudes of $\lambda_1$ and $\lambda_2$ can make a difference. So we fix $\lambda_1=0.1$ and then set $\rho=\lambda_2/\lambda_1$ as $\{0.01,0.05,0.1,0.5,1,5\}$. We expect to show that the influence of data scale on hyper-parameter sensitivity, so we adopt the same graph partitions on Portland data as in section \ref{graph_part} and for each partition group we conduct kriging using all six settings of $\rho$. Note that only the impact of $\lambda_2$ is discussed here as $\lambda_2$ controls the spatial regularization degree, but similar results can be observed by simply fixing $\lambda_2$ and varying $\lambda_1$.

\begin{figure}[!htbp]
  \centering
  \includegraphics[scale=0.6]{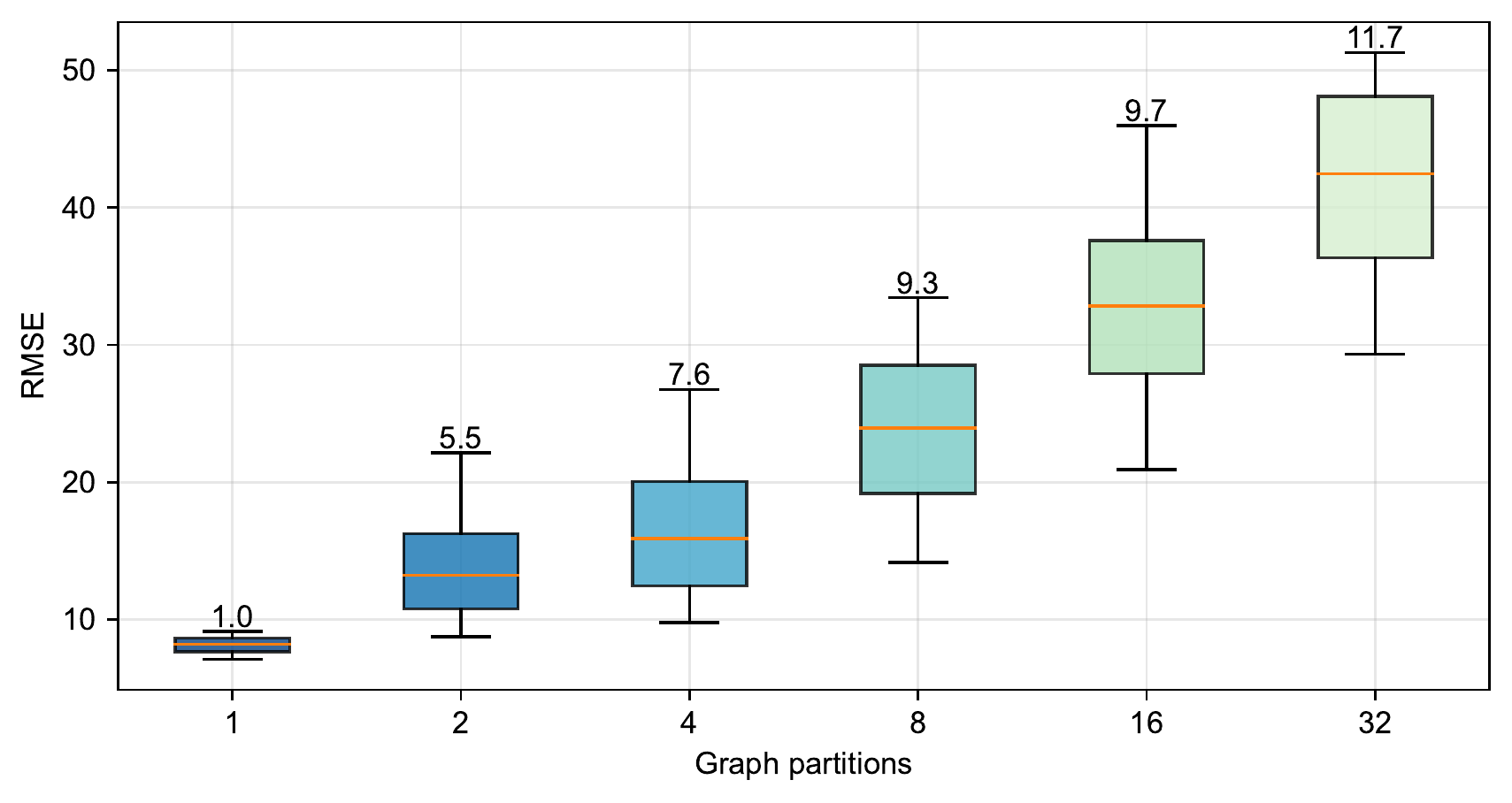}
  \caption{Sensitivity of hyper-parameters under different settings and data scales. The number on the top of each box denotes the IQR.}
  \label{boxplot}
\end{figure}

Fig. \ref{boxplot} gives the box plot for the LETC performances of all $\rho$ setting groups under 6 data scales. Of this figure, each box represents a series of kriging results using different $\rho$ settings on this number of graph partition. We also report the interquartile range (IQR) as a quantitive description, which is defined by the difference between the third and first quartile of each group.

There are two observations from Fig. \ref{boxplot}: (1) The RMSE mean value (orange line in this figure) of each group monotonously increases as the partitions increase, and this result is the same as in Section \ref{graph_part}; (2) If we define the IQR as the parameter sensitivity, it is noteworthy that the sensitivity reduces dramatically as the data scale becomes larger and the distance between the upper bound and lower bound of the box also reflects this trend.
As a consequence, although we need to specify the model hyper-parameters for LETC without a hyper-parameter learning procedure, we could avoid laborious parameter tuning process on network-wide kriging task because the influence of $\lambda_1$ and $\lambda_2$ is marginal when kriging on the whole network. 
In practice, one can obtain the best parameter settings on a small subset of target dataset using cross-validation or grid search techniques and transfer them to the whole data directly, to seek for better model performances.

More results about model parameter sensitivity are discussed in \nameref{Appendix C.}.

\section{Conclusion and future directions}\label{conclusions}

Sparse sensor coverage and data corruption inhibit availability of fine-grained and network-wide traffic speed data for transportation managements.
To cope with this issue, we propose a Laplacian enhanced low-rank tensor completion (LETC) based kriging model which features both low-rankness and multi-dimensional correlations for network-wide traffic speed kriging under limited observations. 
Three types of speed correlation: temporal continuity, temporal periodicity, and spatial proximity are carefully chosen and modeled by three forms of graph Laplacian. 
Specifically, the temporal periodicity is encoded by a new temporal graph Fourier transform (TGFT) and thus the tensor nuclear norm minimization problem can be solved in parallel by an array of daily subproblem. 
Considering the directional nature of traffic flow and traffic network, a novel diffusion graph regularization (DGR) is developed to model spatial proximity. To encourage time series continuity, we introduce a generalized temporal consistency regularization (GTCR). To further adapt for large-scale data, two efficient numeric techniques including conjugate gradient (CG) and randomized tensor singular value thresholding (r-tSVT) algorithm are applied to reduce the computational complexity. By performing experiments on two public million-level traffic speed datasets, we can draw the following conclusion: 
\begin{enumerate}
\item Our proposed LETC model reconstructs more realistic temporal and spatial patterns of traffic speed than competing models, thereby achieving state-of-the-art kriging performances even with low observation rates, while at the same time saving more than half computing time compared with baseline methods; 
\item Model ablation studies demonstrate the effectiveness of each model component as well as the accelerating techniques. Especially, the use of DGR and GTCR benefits the kriging precision significantly, and TNN minimization induced by TGFT maintains high accuracy, efficiency and better interpretability; 
\item The sensitivity of LETC hyper-parameters is marginal on large-scale data and can be tuned in an easy routine.
Moreover, this work also provides new insights into spatiotemporal traffic data kriging at network level: making fewer efforts to divide dataset could benefit both efficiency and accuracy.
\end{enumerate}

There are some directions worth an extension. First, the model performance is dependent on the calculation of adjacent matrix, where a Gaussian kernel is directly used in this work. It is promising to learn an adaptive adjacent matrix from the raw location information to improve the performance. Second, we only conduct kriging for traffic speed, as the distance based adjacent relationships are practicable for loop speed. However, for trajectory based speed or traffic volume, this assumption may not work. How to achieve accurate kriging for traffic volume is a open question.

\section*{Appendix A.}
\label{Appendix A.}
Main hyper-parameter settings for LETC and other baseline models are given as follows: the convergence threshold $\epsilon$ for all above models is set to $10^{-3}$. For LETC, we set the regularization parameter $\mu$ initializing with $10^{-3}$ and updating in each iteration with $\mu=\min\{1.5\times\mu,10^4\}$. The regularization parameters 
$\lambda_1, \lambda_2$ are selected from cross-validations with $\lambda_1=0.01$ and $\lambda_2=0.1$. 
For both PeMS and Portland data, we set $\tau={1,2,3}$ and report corresponding results. For LSTM-GRMF, TGMF and KPMF, a pre-specified rank is necessary, we set it as 60 for Portland data and 200 for PeMS data. For FMDT-Tucker, we set the delay parameter $\tau=[128,256,1]$ for Portland and $[64,32,1]$ for PeMS. The initial rank parameter is set to 10. As for GLTL, the orthogonal projection strategy is adopted in these experiments. For both of the two data, we compute the spatial adjacent matrix by Eq. \eqref{Gaussian}, and prepare the symmetric one for baseline models by selecting the maximum values of its transpose and itself. For PeMS data, the distance matrix is based on travel distance, while coordinate position distance is adopted for Portland data. For GLOSS, we consider graph Laplacian on mode 1 and mode 2 unfoldings. And we set the Lagrange penalty parameter $\beta_1=\beta_2=\beta_3=\beta=10^{-3}, \psi_1=\psi_2=\psi_3=\frac{1}{3}$. As for WDG-CP, the rank is set to 20 for Portland data and 60 for PeMS data. The graph regularization parameter is selected the same as TGMF, and the sparsity parameter is chosen as 1000 according to their work.

\section*{Appendix B.}
\label{Appendix B.}
Supplementary figures of kriging results on PeMS networks, different time slots from Fig. \ref{pems_network} are given as follows.

\begin{figure}[!htb]
\centering
\subfigure[Ground truth speed values]{
\centering
\includegraphics[scale=0.09]{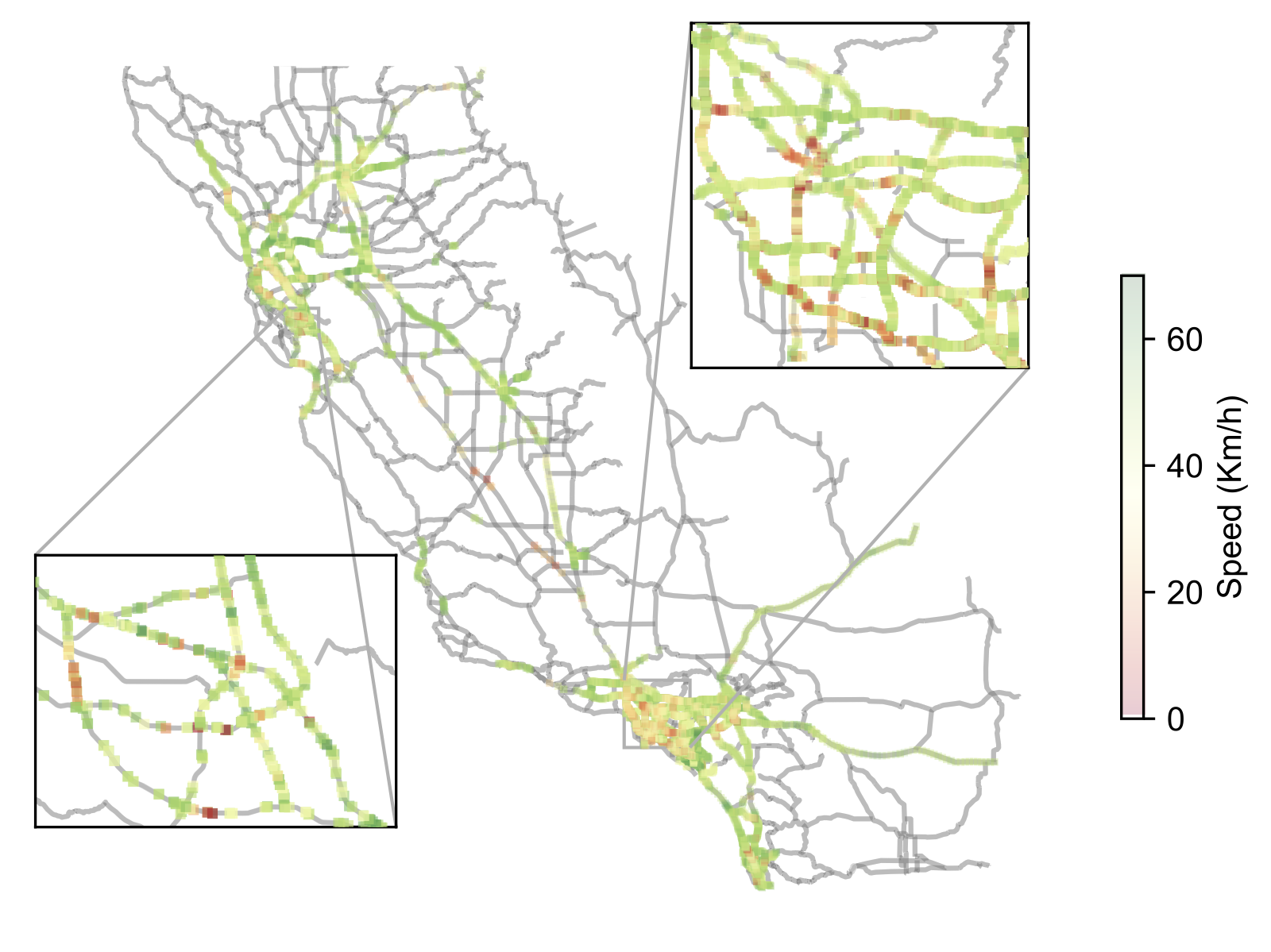}
}
\centering
\subfigure[LETC kriging speed values (Proposed method)]{
\centering
\includegraphics[scale=0.09]{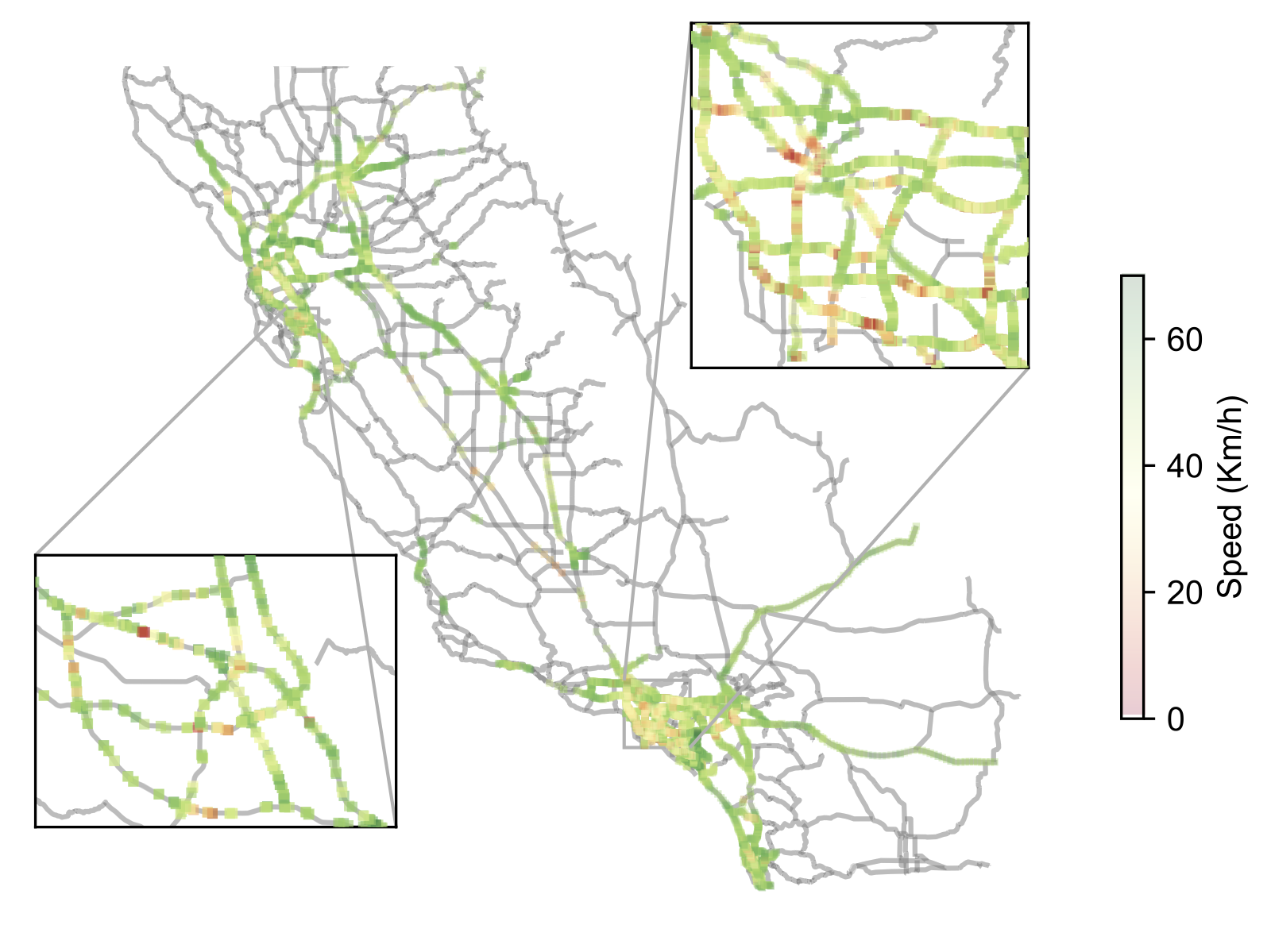}
}
\centering
\subfigure[LSTM-GRMF kriging speed values (Best benchmark)]{
\centering
\includegraphics[scale=0.09]{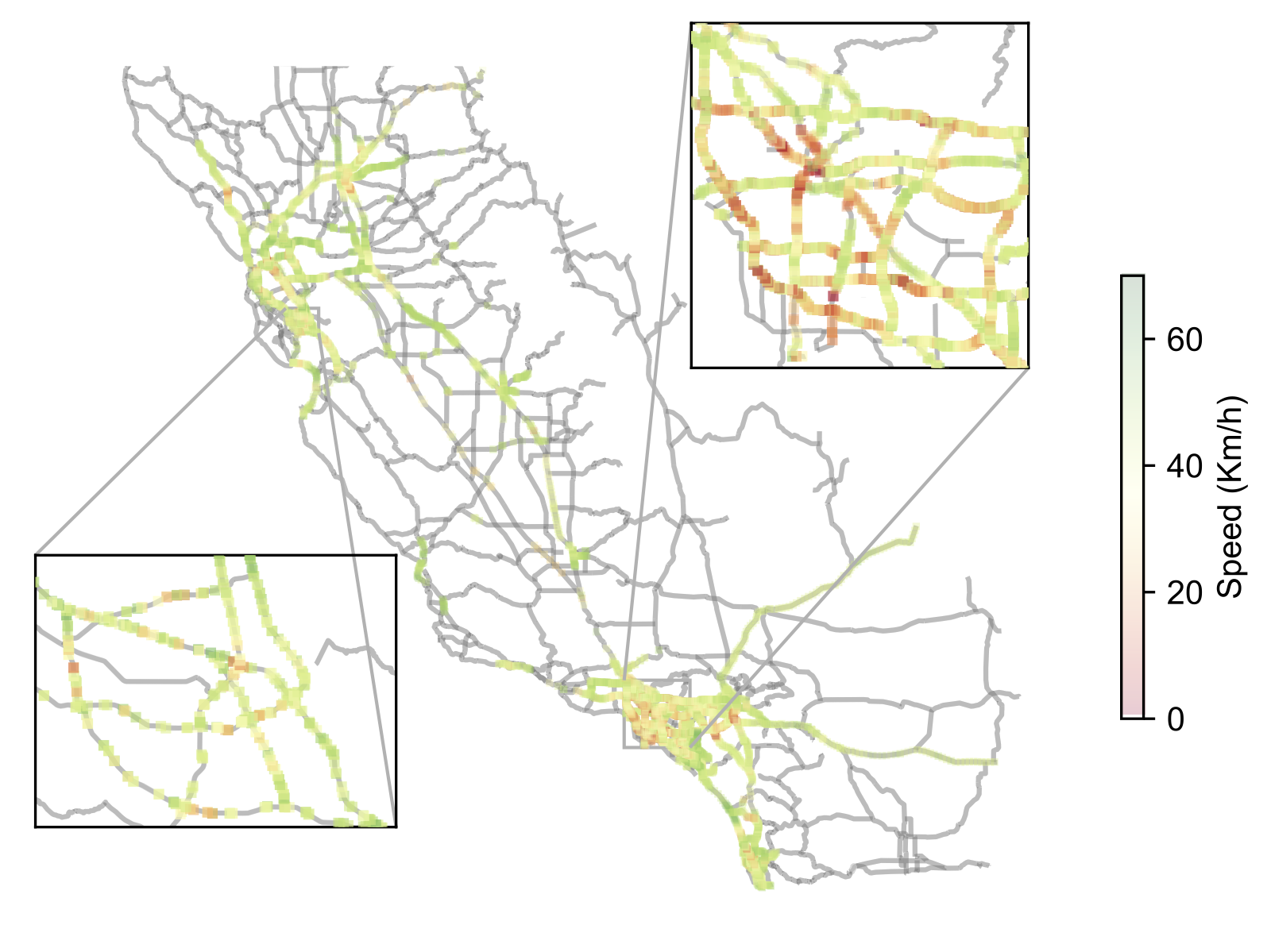}
}
\centering
\subfigure[Ground truth speed values]{
\centering
\includegraphics[scale=0.09]{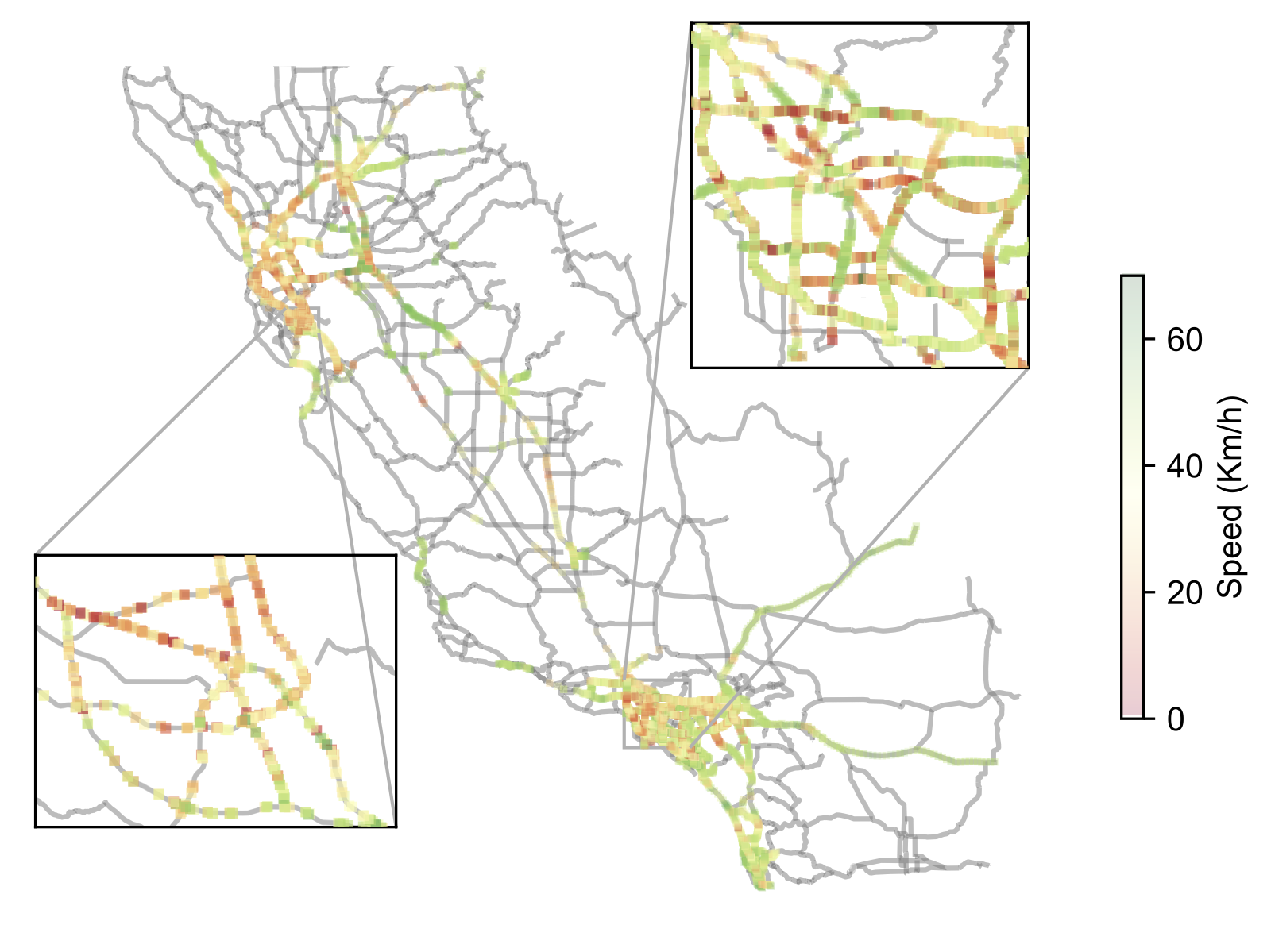}
}
\centering
\subfigure[LETC kriging speed values (Proposed method)]{
\centering
\includegraphics[scale=0.09]{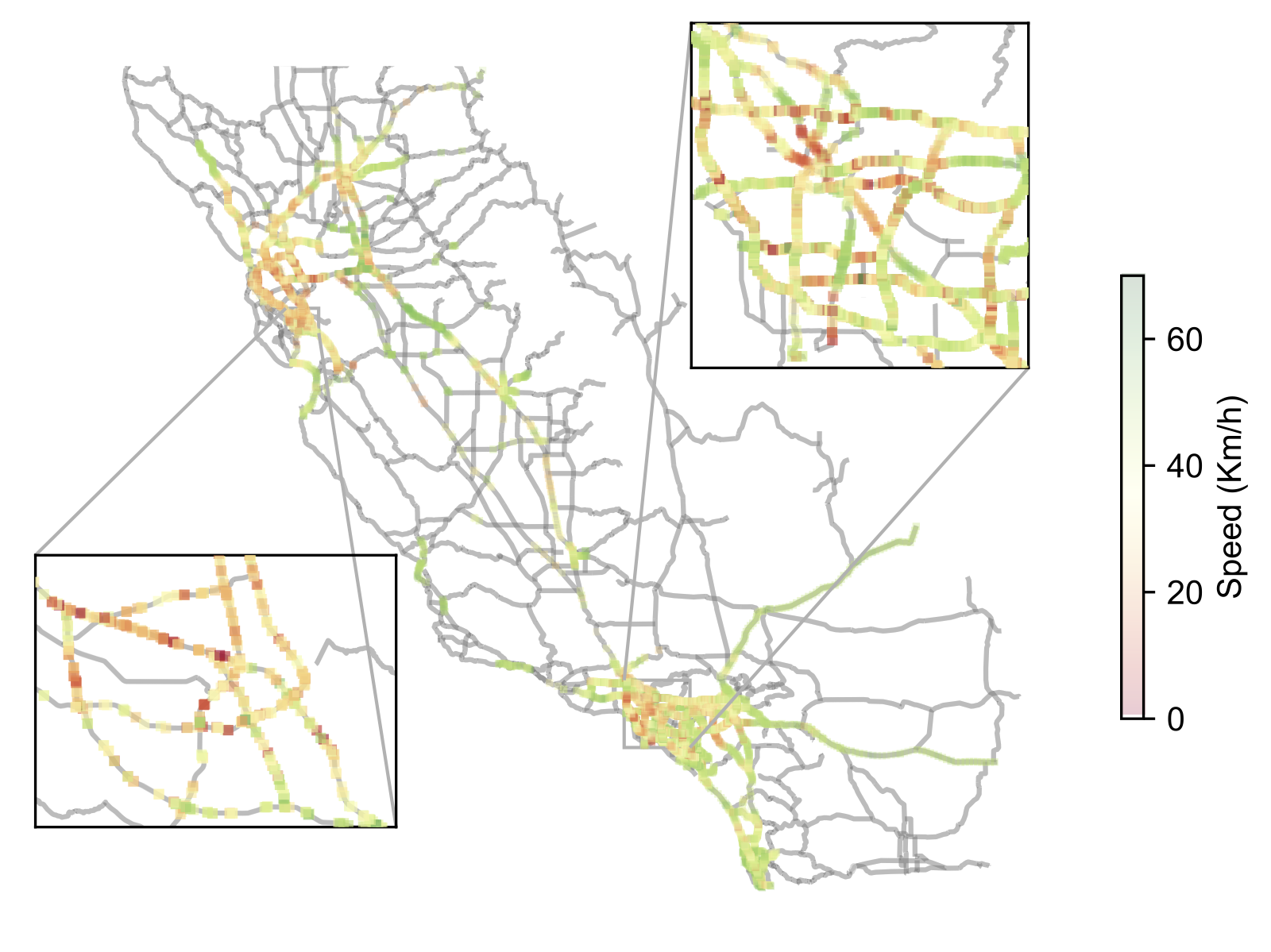}
}
\centering
\subfigure[LSTM-GRMF kriging speed values (Best benchmark)]{
\centering
\includegraphics[scale=0.09]{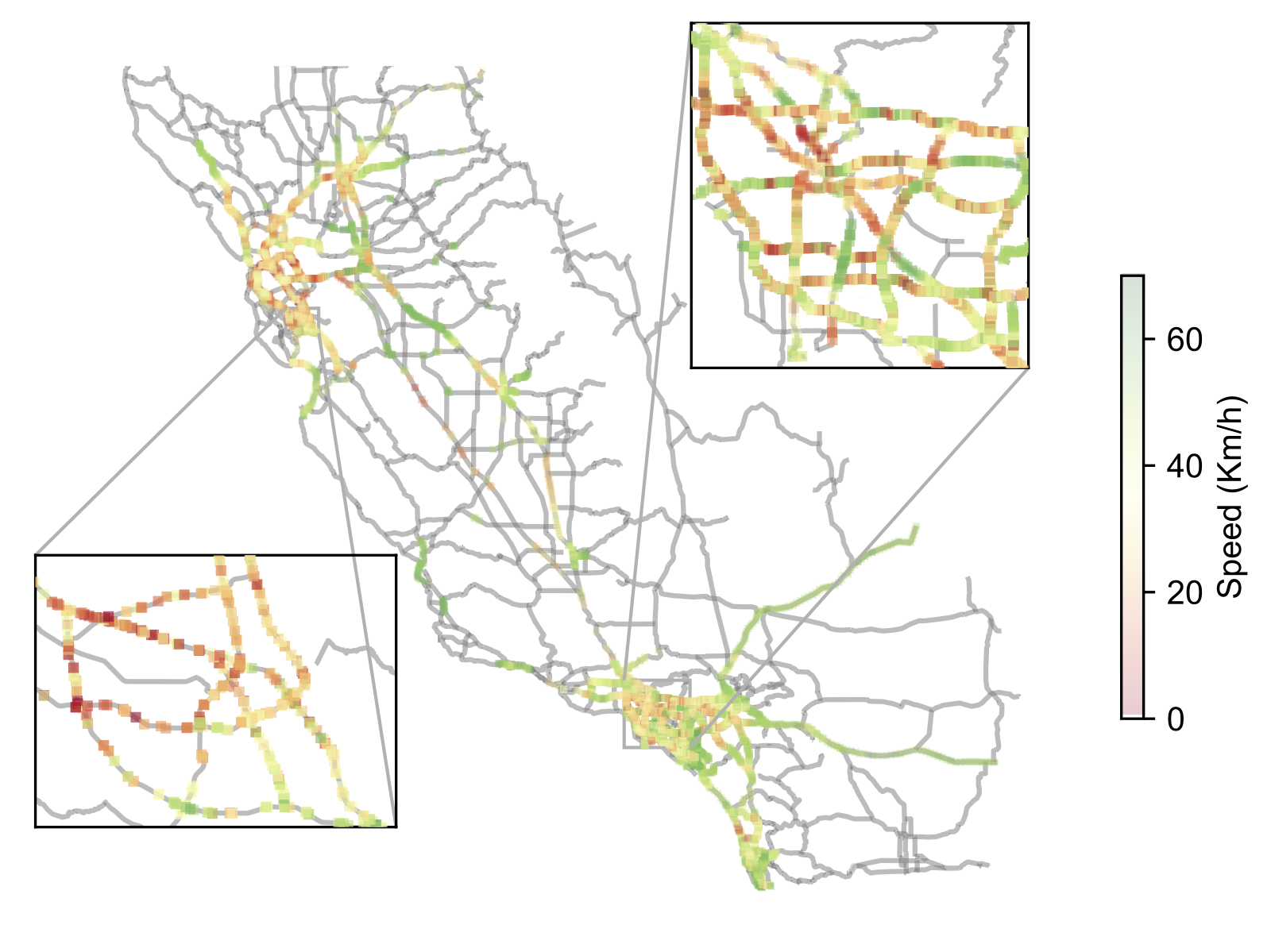}
}
\caption{Supplementary figures of kriging results on PeMS networks. Results on two different time slots are plotted. (a)-(c): 14:30 P.M., (d)-(f): 17:30 P.M.}
\label{pems_network_sup}
\end{figure}

% \minew{Experiment results on PeMS data with different diffusion kernels.}

\section*{Appendix C.}
\label{Appendix C.}
We also conduct sensitivity studies on another two hyper-parameters, i.e., the Gaussian kernel parameter $\sigma$ in Eq. \eqref{Gaussian}, and the rank parameter $k$ for rSVD in Algorithm \ref{r_tsvt}.

\begin{figure}[!htbp]
  \centering
  \includegraphics[scale=0.5]{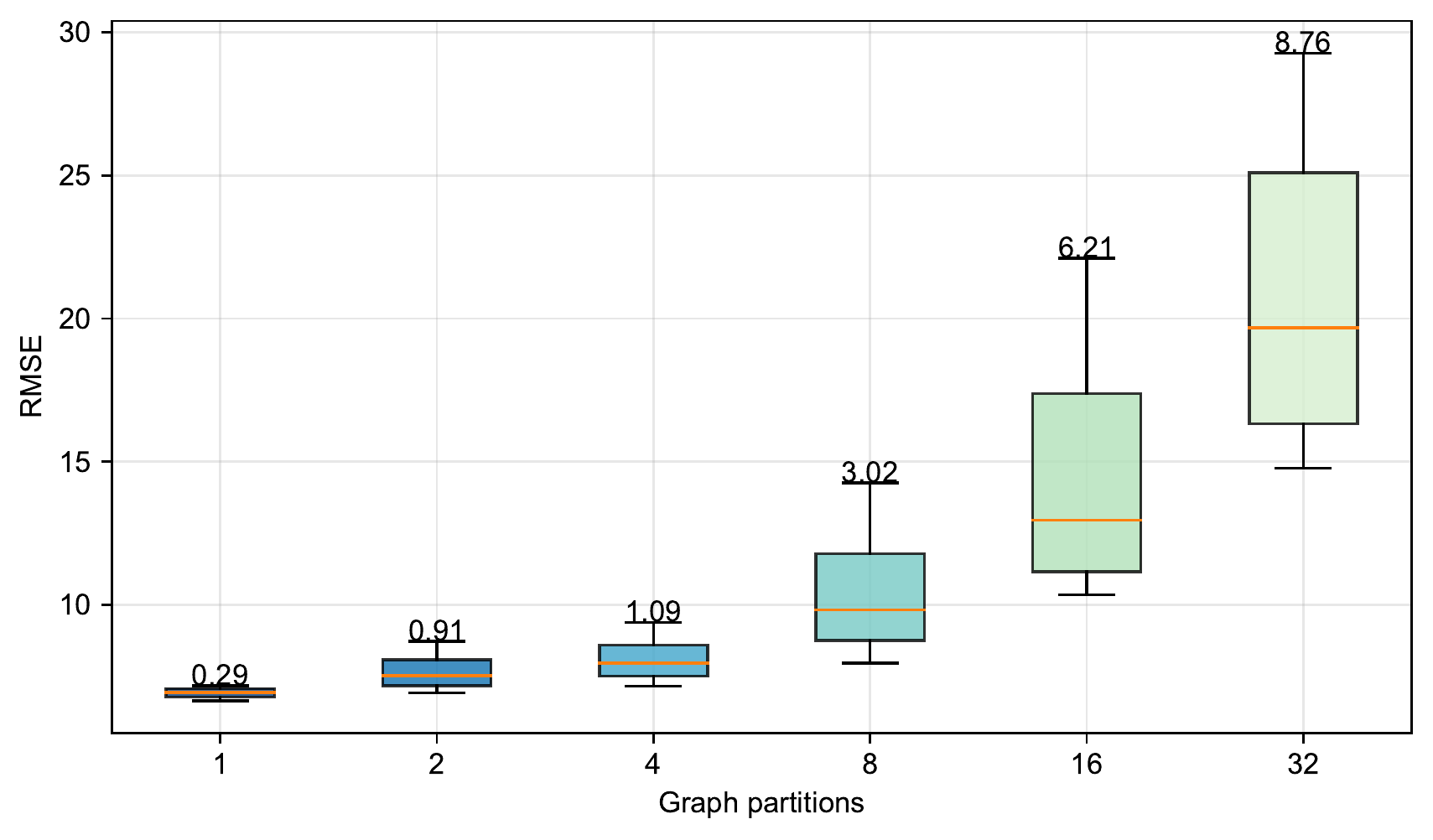}
  \caption{Sensitivity of kernel function under different settings and data scales. The number on the top of each box denotes the IQR.}
  \label{kernel_sensitivity}
\end{figure}

In general, the Gaussian distance function is given as:
\begin{equation*}
\label{Gaussian_delta}
a^s_{ij}=\operatorname{exp}\left(-\left(\frac{\operatorname{dist}(v_i,v_j)}{\delta\sigma}\right)^2\right),
\end{equation*}
where $\delta$ is a scale parameter needs to be tuned. Similar to the practice in section \ref{hpt}, we set $\delta$ as $\{0.5,0.75,1.0,1.5,2.0,2.5\}$ and test LETC performances under different graph partitions. Each box in Fig. \ref{kernel_sensitivity} reports the results of all parameter settings under certain graph partition.  

As we can see, model shows varying performances with different $\delta$ values, which is consistent with the findings in the literature \citep{lei2022bayesian,chen2022bayesian}. Nevertheless, by reducing the number of graph partition, the range of fluctuation denoted by IQR decreases drastically, showing similar trends to $\lambda$ in section \ref{hpt}. As a consequence, it is not necessary to conduct hyper-parameter learning on such large-scale dataset, as the benefits could be less significant.

\begin{figure}[!htbp]
  \centering
  \includegraphics[scale=0.6]{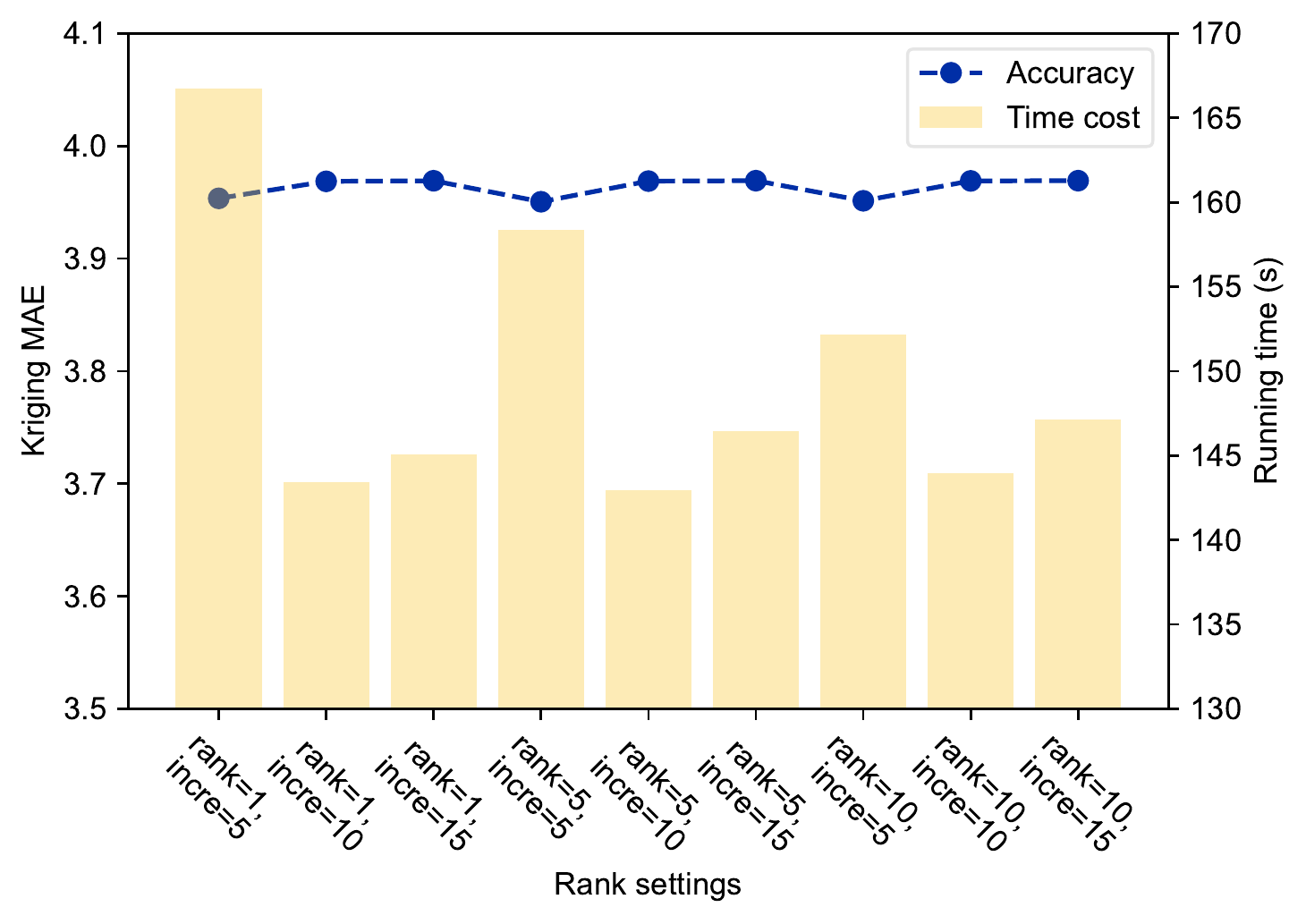}
  \caption{Model performances under different rSVD rank settings.}
  \label{rank_sensitivity}
\end{figure}

Rank value $k$ in rSVD is supposed to control the convergence speed. In our implementation, we set a small initial value at first, then increase it per iteration. When the whole LETC algorithm converges, $k$ is usually convergent. Therefore, this value mainly affects the convergence speed, has minor impacts on solving accuracy. Fig. \ref{rank_sensitivity} shows LETC performances with different rank initial and increment values. From these results, it is preferable to set a small value initially, and then increase it with large step. 

\section*{Appendix D.}
\label{Appendix D.}
In this section we give detailed derivation of closed-form solution to problem \eqref{dirichlet}. To ease the presentation, we consider the fully-observed case and assume the graph to be symmetric (similar conclusions can be draw for other settings), and the objective can be rearranged as follows:
    \begin{equation*}
    \Bar{\mathbf{X}}=\arg \min_{\Bar{\mathbf{X}}} \Vert\Bar{\mathbf{X}}-\mathbf{X}\Vert_{\mathbf{D}}^2 + \operatorname{Tr}(\Bar{\mathbf{X}}^{\mathsf{T}}\mathbf{L}\Bar{\mathbf{X}}).
\label{dirichlet1}
\end{equation*}
By setting the gradient of $\Bar{\mathbf{X}}$ to zero, we have:
\begin{equation*}
\begin{aligned}
    &(\mathbf{D}+\mathbf{D}^{\mathsf{T}})(\Bar{\mathbf{X}}-\mathbf{X}) + (\mathbf{L}+\mathbf{L}^{\mathsf{T}})\Bar{\mathbf{X}}=0,\\
    \Rightarrow &(\mathbf{I}+\mathbf{D}^{-1}\mathbf{L})\Bar{\mathbf{X}}=\mathbf{X},\\
    \Rightarrow &\Bar{\mathbf{X}}=(\mathbf{I}+\widetilde{\mathbf{L}})^{-1}\mathbf{X}.
\end{aligned}
\end{equation*}

\section*{Acknowledgement}
This research was sponsored by the National Natural Science Foundation of China (52125208), and the Science and Technology Commission of Shanghai Municipality (No. 22dz1203200).

% \newpage
\footnotesize
\bibliographystyle{elsarticle-harv}
\bibliography{main}

% \begin{thebibliography}{33}
% \expandafter\ifx\csname natexlab\endcsname\relax\def\natexlab#1{#1}\fi
% \expandafter\ifx\csname url\endcsname\relax
%   \def\url#1{\texttt{#1}}\fi
% \expandafter\ifx\csname urlprefix\endcsname\relax\def\urlprefix{URL }\fi

% \bibitem[{Asif et~al.(2016)Asif, Mitrovic, Dauwels, and
%   Jaillet}]{asif2016matrix}
% Asif, M.~T., Mitrovic, N., Dauwels, J., Jaillet, P., 2016. Matrix and tensor
%   based methods for missing data estimation in large traffic networks. IEEE
%   Transactions on intelligent transportation systems 17~(7), 1816--1825.
% \bibitem[{Bolte et~al.(2014)Bolte, Sabach, and Teboulle}]{bolte2014proximal}
% Bolte, J., Sabach, S., Teboulle, M., 2014. Proximal alternating linearized
%   minimization for nonconvex and nonsmooth problems. Mathematical Programming
%   146~(1), 459--494.
% \end{thebibliography}

\end{document}